\documentclass{movens_report}

\usepackage{amsmath,amsfonts,bm}

\usepackage{hyperref}
\usepackage{url}
\usepackage{xspace}
\usepackage{booktabs}
\usepackage{listings}
\usepackage{makecell}
\usepackage{graphicx}
\usepackage{pifont}
\usepackage{multirow}
\ifPDFTeX
  \newcommand{\movensfont}{\ttfamily}
\else
  \newfontfamily\movensfont{lmmonolt10}[
    Extension=.otf,
    UprightFont=*-regular,
    BoldFont=*-bold,
    ItalicFont=*-oblique,
    BoldItalicFont=*-boldoblique
  ]
\fi

\definecolor{MovensBrandColor}{HTML}{AA4A44}
\DeclareRobustCommand{\movensbrand}[1]{\mbox{{\movensfont\bfseries\textcolor{MovensBrandColor}{#1}}}}
\DeclareRobustCommand{\movenszero}{\texorpdfstring{\movensbrand{Movens-Zero}}{Movens-Zero}\xspace
}
\DeclareRobustCommand{\movensdata}{\texorpdfstring{\mbox{Movens-5M}}{Movens-5M}\xspace
}
\newcommand{\movensintent}{\texorpdfstring{\mbox{Movens-Intent}}{Movens-Intent}\xspace}
\newcommand{\mmrtc}{MM-RTC\xspace}
\newcommand{\tick}{\textcolor{green}{\ding{51}}}
\newcommand{\cross}{\textcolor{red}{\ding{55}}}

\title{\texorpdfstring{\fontsize{20}{24}\selectfont
  Movens-Zero: Scaling Humanoid Motion Foundation Models with Large-Scale Human Videos}{Movens-Zero: Scaling Humanoid Motion Foundation Models with Large-Scale Human Videos}}

\author{Wendong Bu\textsuperscript{1,3,4,*},
  Tinghong Ye\textsuperscript{1,*},
  Qizhou Wang\textsuperscript{2,3,*},
  Jiacheng Li\textsuperscript{5},
  Zhiqi Ge\textsuperscript{1},
  Yuze Lin\textsuperscript{1},
  Zhongqi Yue\textsuperscript{6},\\
  Yi Su\textsuperscript{2},
  Jingyu Wang\textsuperscript{1},
  Kai Shen\textsuperscript{1},
  Jingdong Liu\textsuperscript{1},
  Chengyuan Deng\textsuperscript{1},
  Xiaoxuan Chen\textsuperscript{1},
  Xin Tan\textsuperscript{7},\\
  Jiaming Ji\textsuperscript{8},
  Wenqiao Zhang\textsuperscript{1},
  Chenghong Jiang\textsuperscript{4},
  Tuoyu Li\textsuperscript{1},
  Siliang Tang\textsuperscript{1,3},
  Jun Xiao\textsuperscript{1},\\
  Tat-Seng Chua\textsuperscript{9},
  Yueting Zhuang\textsuperscript{1},
  Xingxing Wang\textsuperscript{2},
  Juncheng Li\textsuperscript{1,3,4,\ding{41}}}

\zjuaffiliation{\textsuperscript{1}\,Zhejiang University\quad
  \textsuperscript{2}\,Unitree Robotics\quad
  \textsuperscript{3}\,Joint Laboratory of Embodied Intelligence, ZJU \& Unitree Robotics\quad\\
  \textsuperscript{4}\,Bit Oasis\quad
  \textsuperscript{5}\,HiThink Research\quad
  \textsuperscript{6}\,Microsoft Research\quad
  \textsuperscript{7}\,East China Normal University\quad\\
  \textsuperscript{8}\,Peking University\quad
  \textsuperscript{9}\,National University of Singapore
}

\zjuauthornote{\textsuperscript{*}Equal Contribution.\qquad \textsuperscript{\ding{41}}Corresponding Author.}

\zjureporttype{Technical Report}
\zjurunningtitle{\textbf{Movens-Zero:} Scaling Humanoid Motion Foundation Models with Large-Scale Human Videos}
\zjuprojectpage{https://movens-zero.github.io}
\date{October 2026}

\begin{document}

\maketitle

\begin{abstract}
Scaling humanoid motion foundation models is constrained by the scarcity of diverse, robot-executable data and by the difficulty of jointly learning from heterogeneous multimodal inputs while supporting real-time closed-loop control.
We present \textbf{\movensdata}, constructed from raw Internet videos through clip generation, quality filtering, semantic annotation, motion recovery, and physical refinement with humanoid retargeting and execution verification.
It contains 15 million human-motion sequences and 5 million physically executable humanoid-motion sequences spanning over 10,000 hours, aligned with text, RGB video, speech, and music.
Building on this data, we introduce \textbf{\movenszero}, a scalable understanding--generation--execution architecture that supports joint learning across modalities and real-time closed-loop execution.
Task-mixed flow-matching training jointly optimizes a single action expert across seven generation tasks, enabling direct motion generation from text, speech, music, and video. Multimodal Real-Time Chunking connects overlapping chunks under continuously updated instructions and robot states for continuous closed-loop execution.
We further introduce \textbf{\movensintent}, a four-track benchmark with 8,192 expert-verified condition--motion pairs across text, speech, music, and video.
Experiments show consistent gains across all four modalities when scaling execution-verified data from 1\% to 100\% and action-expert capacity from 64M to 911M parameters, while demonstrating that data quality and cross-modal supervision improve generation and execution.
\end{abstract}

\begin{figure*}[!h]
\centering
\includegraphics[width=\linewidth]{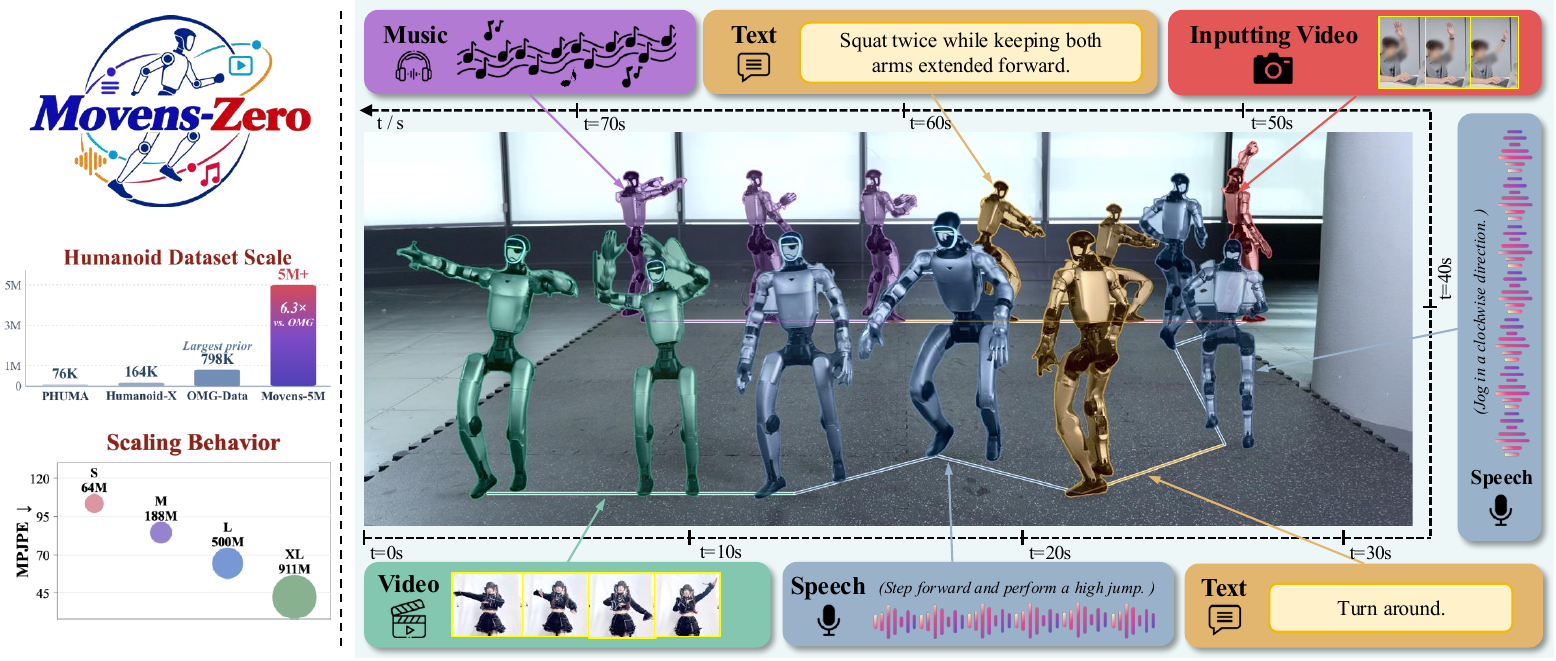}
\caption{Overview of \movenszero, a scalable understanding--generation--execution architecture built on \movensdata. It translates multimodal human intent directly into physically executable humanoid behaviors in real time.}
\label{fig:overview}
\end{figure*}

\section{Introduction}

\begin{table*}[t]
\centering
\small
\caption{Comparison with representative human and humanoid motion datasets.
``Types'' counts published dataset-level motion categories.
``Scenes'' counts coarse environments.
``Constraints'' and ``Policies'' count physical-constraint and control-policy families used in curation and execution verification, with real-world collection counted as one constraint setting.
``Agg.'' denotes aggregation.
``--'' marks unavailable and inapplicable entries.}
\label{tab:dataset_comparison}
\resizebox{\textwidth}{!}{
\begin{tabular}{lcrrrr|cc|cc|cccc|l}
\toprule
\multirow{2}{*}{\textbf{Dataset}} &
\multirow{2}{*}{\textbf{Open}} &
\multicolumn{2}{c}{\textbf{\# Sequences}} &
\multirow{2}{*}{\textbf{Hours}} &
\multirow{2}{*}{\textbf{Frames}} &
\multicolumn{2}{c}{\textbf{Diversity}} &
\multicolumn{2}{c}{\textbf{Quality checks}} &
\multicolumn{4}{c}{\textbf{Aligned modalities}} &
\multirow{2}{*}{\textbf{Source}} \\[-3pt]
\cmidrule(lr){3-4}\cmidrule(lr){7-8}\cmidrule(lr){9-10}\cmidrule(lr){11-14}
& & \textbf{Human} & \textbf{Humanoid} & & & \textbf{Types} & \textbf{Scenes} & \textbf{Constraints} & \textbf{Policies} & \textbf{Text} & \textbf{Music} & \textbf{Speech} & \textbf{RGB} & \\
\midrule
\multicolumn{15}{l}{\textit{Human motion datasets}} \\
AMASS~\citep{mahmood2019amass}             & \tick  & 11.3K  & -- & 40.0h    & $\approx$4.32M   & -- & 1 & \cross & \cross & \cross & \cross & \cross & \cross & MoCap agg. \\
LAFAN1~\citep{harvey2020robust}            & \tick  & 77     & -- & 4.6h     & 496.7K  & 15 & 1 & \cross & \cross & \cross & \cross & \cross & \cross & MoCap \\
KIT-ML~\citep{plappert2016kit}             & \tick  & 3.9K   & -- & 11.2h    & 4.03M   & -- & 1 & \cross & \cross & \tick  & \cross & \cross & \cross & MoCap agg. \\
HumanML3D~\citep{guo2022humanml3d}         & \tick  & 14.6K  & -- & 28.6h    & 2.06M   & -- & 1 & \cross & \cross & \tick  & \cross & \cross & \cross & MoCap agg. \\
BABEL~\citep{punnakkal2021babel}           & \tick  & 13.2K  & -- & 43.5h    & 4.70M   & 260 & 1 & \cross & \cross & \tick  & \cross & \cross & \cross & MoCap agg. \\
SnapMoGen~\citep{guo2025snapmogen}         & \tick  & 20K    & -- & 44.0h    & 4.7M    & -- & 1 & \cross & \cross & \tick  & \cross & \cross & \cross & MoCap \\
AIST++~\citep{li2021aistpp}                & \tick  & 1.4K   & -- & 5.2h     & 1.1M    & 10 & 1 & \cross & \cross & \cross & \tick  & \cross & \tick  & Multiview video \\
BONES-RP01~\citep{shi2026gpc}              & \cross & 343K   & -- & $>$680h  & $>$293.8M & -- & 1 & \cross & \cross & \tick  & \cross & \cross & \tick  & MoCap \\
Motion-X++~\citep{zhang2025motionxpp}      & \tick  & 120.5K & -- & 180.9h   & 19.5M     & -- & 2 & \cross & \cross & \tick  & \tick  & \cross & \tick  & Video + agg. \\
MotionLib~\citep{wang2025motionlib}         & \cross & 1.21M  & -- & 1456.4h  & $\sim$137M & -- & -- & 1 & 1 & \tick  & \cross & \cross & \tick  & Video + agg. \\
MotionMillion~\citep{fan2025motionmillion} & \tick  & 2.0M   & -- & $>$2000h & $>$216M   & -- & 2 & \cross & \cross & \tick  & \cross & \cross & \cross & Web video + agg. \\
OpenT2M~\citep{cao2026opent2m}             & \tick  & 1.0M   & -- & $>$2800h & $>$302.4M & -- & -- & \cross & 1 & \tick  & \cross & \cross & \cross & Data agg. \\
HuMo100M~\citep{cao2025motionctrl}          & \cross & 5.7M   & -- & 8508.3h  & 918.9M     & -- & -- & \cross & 1 & \tick  & \cross & \cross & \tick  & Web video + agg. \\
MotionHub~\citep{ling2024motionllama}      & \tick  & 1.11M  & -- & 1528h    & 164.98M    & -- & -- & \cross & \cross & \tick  & \tick  & \tick  & \cross & Data agg. \\
OmniHuMo~\citep{li2026anymo}               & \tick  & 3.2M   & -- & 5048h    & 545.2M     & -- & 2 & 3 & \cross & \tick  & \tick  & \tick  & \tick  & Web video \\
\midrule
\multicolumn{15}{l}{\textit{Humanoid motion datasets}} \\
OmniRetarget~\citep{yang2025omniretarget}  & \tick  & --     & 2.1K   & 4.0h        & $\approx$432K   & --  & 3 & 4 & 1 & \cross & \cross & \cross & \cross & MoCap + agg. \\
BONES-SEED~\citep{bones2026seed}            & \tick  & 142.2K & 142.2K & 288.3h      & 124.6M & 20  & 1 & \cross & \cross & \tick  & \cross & \cross & \cross & MoCap \\
Humanoid-X~\citep{mao2024learning}          & \tick  & 163.8K & 163.8K & 240.3h      & 20.7M  & 8   & -- & \cross & 1 & \tick  & \cross & \cross & \tick  & Video + agg. \\
Humanoid-Union~\citep{wei2025humanoidunion} & \cross & --     & 170K   & $>$260h & $>$28.1M & --  & -- & \cross & 1 & \tick  & \cross & \cross & \cross & Video + agg. \\
PHUMA~\citep{lee2025phuma}                  & \tick  & --     & 76.0K  & 73.0h       & 7.88M  & 11  & 2 & 3 & 3 & \cross & \cross & \cross & \cross & Video + MoCap \\
OMG-Data~\citep{huang2026omg}               & \tick  & --     & 798.2K & 1122.0h     & 121.2M & --  & -- & 3 & 1 & \tick  & \tick  & \cross & \cross & Data agg. \\
\midrule
\textbf{\movensdata (Ours)} & \tick & \textbf{15M} & \textbf{5M} & \textbf{$>$10,000h} & \textbf{$>$1.08B} & \textbf{297} & \textbf{16} & \textbf{6} & \textbf{4} & \tick & \tick & \tick & \tick & \textbf{Web video} \\
\bottomrule
\end{tabular}
}
\end{table*}

Scaling training data and model capacity has substantially improved the capabilities and generalization of language and vision models~\citep{brown2020language,geminiteam2023gemini,openai2024sora}.
However, training data for humanoid robots are difficult to acquire directly from the Internet, unlike the vast web text corpora available for language modeling.
But the Internet contains countless human videos capturing actions and scenes, offering a potential data source~\citep{mao2024learning,wu2026sugar}.
This raises a natural question: \textbf{can we scale learning through human videos from the Internet to build generalist humanoid robots?}
Generalist humanoid robots would use vision and hearing to perceive their surroundings and understand human instructions in real time, then autonomously generate and execute humanoid motion in a closed loop, just like humans.
However, such robots have yet to be realized.
Based on experience in language and vision, we argue that this is because humanoid motion foundation models have not scaled up to the same extent as general-purpose foundation models in language and vision, limiting emergent capabilities such as multimodal instruction following.

Why haven't humanoid motion foundation models scaled up?
Two closely related challenges constrain further scaling.
\textbf{First, scaling data construction: how can human videos from the Internet be converted into supervision suitable for large-scale robot learning?}
Internet videos vary widely in quality, and issues such as camera shake and occlusion can introduce errors into motion recovery~\citep{shen2024gvhmr,wang2025prompthmr}.
Even when human motion is accurately recovered, the human--robot embodiment gap can introduce further errors during motion retargeting~\citep{xu2025parc,araujoRetargetingMattersGeneral2025}.
Moreover, even if the resulting robot motion is accurate, it may not satisfy physical constraints, so its executability still needs to be verified.
Building generalist humanoid robots also requires multimodal human instructions accurately aligned with robot motion to support large-scale training for multimodal instruction following.
\textbf{Second, scaling model training: how can large-scale training improve multimodal instruction following while supporting real-time closed-loop control?}
Text, audio, and video differ in format and temporal structure.
A shared motion-prediction architecture is needed to handle these heterogeneous inputs.
Building on this architecture, a common learning objective and joint training strategy are needed to learn from data across modalities.
Scaling model capacity also increases inference costs, making real-time closed-loop control more challenging.

\begin{figure*}[t]
\centering
\includegraphics[width=\linewidth]{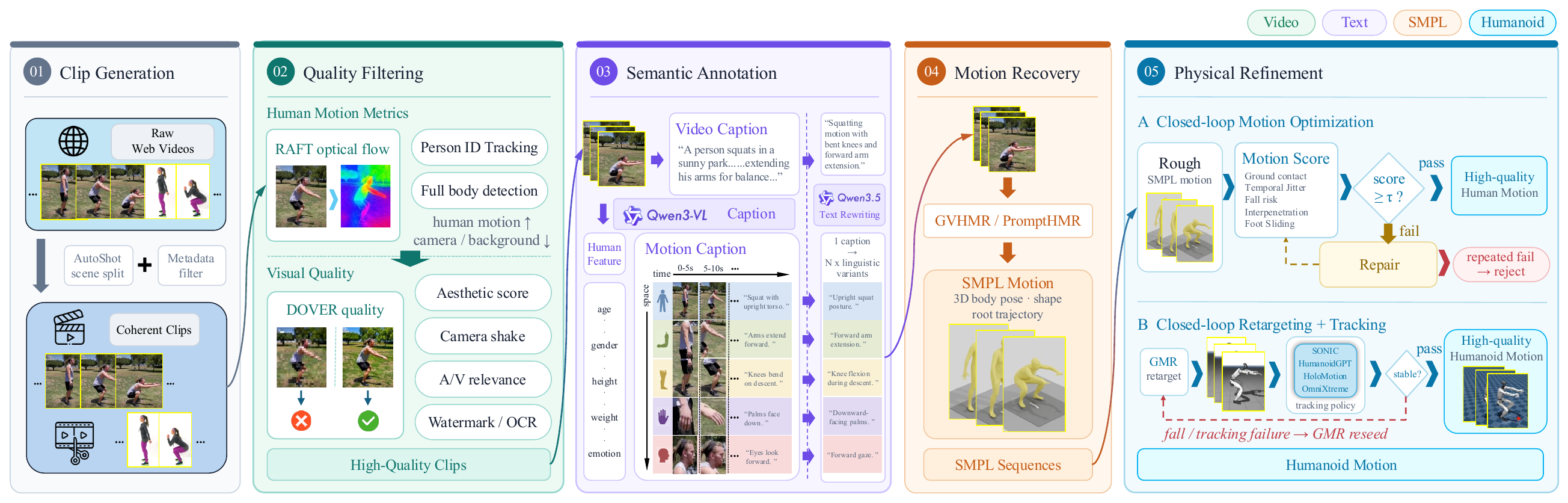}
\caption{Overview of the \movensdata data construction pipeline. Raw web videos are segmented and filtered, semantically annotated, and reconstructed as SMPL-family estimates canonicalized to SMPL-X. Closed-loop physical refinement, humanoid retargeting, and tracking verification produce high-quality, executable humanoid motions.}
\label{fig:pipeline}
\end{figure*}

To construct supervision suitable for large-scale robot learning from human videos, we develop a multi-stage data engine that integrates fine-grained quality assessment, iterative motion refinement, and execution-based verification.
The pipeline shown in Figure~\ref{fig:pipeline} performs filtering, semantic annotation, motion recovery, physical refinement, cross-embodiment retargeting, and execution-based verification.
Through this pipeline, we construct \textbf{\movensdata} entirely from raw videos collected across diverse Internet sources, rather than simply aggregating existing small-scale motion datasets.
The resulting dataset contains more than \textbf{10,000 hours} and \textbf{5 million physically executable humanoid motion sequences}, paired with corresponding human motions and aligned text, RGB video, speech, and music.
As summarized in Table~\ref{tab:dataset_comparison}, with \textbf{297 motion types}, \textbf{16 coarse scenes}, \textbf{6 physical-constraint families}, and verification under \textbf{4 tracking policies}, \movensdata provides the volume, diversity, multimodal supervision, and executable quality needed to study data scaling for multimodal instruction following.
Details of data filtering, motion quality assessment, retargeting and execution verification are provided in Appendix~\ref{app:dataset_details}.
To support consistent evaluation of multimodal instruction following in humanoid robots, we further introduce \textbf{\movensintent}, a four-track benchmark comprising 8,192 expert-verified  multimodal instructions across text, speech, music, and video~(\textit{e.g.}, executing spoken commands, dancing to music, and video-based teleoperation).

Building on \movensdata, we introduce \textbf{\movenszero}, an understanding--generation--execution architecture that unifies heterogeneous multimodal inputs for scalable instruction-following training while supporting real-time closed-loop control (Figure~\ref{fig:model}).
Its pretrained omni-modal backbone converts text, video, and audio into a shared token sequence that preserves temporal order and cross-modal context.
A shared flow-based action expert generates future motion conditioned jointly on this context, the latest observed robot state, and frame-wise motion progress.
At each generation step, Multimodal Real-Time Chunking (MM-RTC) receives the latest multimodal instruction and observed robot state and generates an overlapping motion chunk with the previous prediction's unexecuted tail kept fixed, enabling continuous execution by a high-frequency tracking controller.
We jointly train the action expert in a task-mixed flow-matching stage across seven generation tasks.
During data and model scaling, the unified motion representation and training objective support joint learning from heterogeneous multimodal inputs, while MM-RTC translates multimodal instructions into motion chunks at a low generation frequency to support real-time closed-loop execution.
These designs enable multimodal instruction following to benefit from data and model scaling while remaining compatible with real-time closed-loop control.
The action expert scale configurations, task-mixture ratios, and complete training hyperparameters are provided in Appendices~\ref{app:method_details} and \ref{app:training_details}.

Our experiments reveal three findings.
\textbf{First, \movenszero jointly maintains condition alignment, motion fidelity, and robot executability across input modalities}:
on the source-disjoint \movensintent benchmark, it achieves a text-conditioned R@3 of 62.57\% (Table~\ref{tab:text_results}), reduces video-conditioned W-MPJPE$_{100}$ to 49.1\,mm, 16.2\% lower than the strongest baseline (Table~\ref{tab:video_results}), reduces music-conditioned FID-k to 58.43, 1.7\% lower than the strongest baseline (Table~\ref{tab:music_results}), and achieves a speech-conditioned R@3 of 74.30\%, exceeding the strongest baseline by 11.90 percentage points (Table~\ref{tab:speech_results}).
\textbf{Second, increasing both data coverage and action-expert capacity yields consistent downstream gains}:
with the model size fixed, scaling \movensdata from 1\% to 100\% improves Text R@3 from 12.72\% to 62.57\% (Table~\ref{tab:data_scaling}). On the full dataset, scaling the action expert from 64.2M to 910.5M parameters improves Text R@3 from 10.30\% to 62.57\% (Table~\ref{tab:model_scaling}).
\textbf{Third, data quality and cross-modal supervision improve generation and execution at matched scale}:
with sample counts and training budgets matched, full data curation raises Text R@3 by 13.85 percentage points and Tracking Score from 0.41 to 0.72. Removing video--motion or audio--motion alignment lowers Text R@3 by 0.83 and 2.55 percentage points, respectively, while removing text--motion alignment increases Video MPJPE from 40.90 to 56.84\,mm (Table~\ref{tab:ablations}).
Together, these findings establish a scalable path from web-scale multimodal human demonstrations to multimodal instruction following through humanoid motion generation and execution.

\section{Related Work}

\subsection{Humanoid Motion Generation and Control}

A prevalent approach to semantic humanoid control follows a generator--tracker paradigm: a high-level model generates reference motions, which are executed by a low-level tracking policy.
OMG supports language, audio, and human-motion conditions; TextOp generates short-horizon motions from streaming text; and UniAct converts multimodal commands into motion tokens for whole-body tracking~\citep{huang2026omg,xie2026textop,jiang2025uniact}.
Other methods directly couple semantic commands with robot actions or control-oriented motion representations~\citep{wang2025sentinel,li2025language}.
\movenszero retains the interpretable generator--tracker structure while learning its shared motion prior from diverse, execution-verified data; the latest robot state conditions online generation before low-level tracking.

\subsection{Scaling Motion Generation and Control}

Scaling has been explored separately in kinematic motion generation and humanoid control.
ScaMo studies explicit scaling laws for autoregressive human-motion generation, while large motion models and datasets demonstrate continued gains from increasing data and model capacity~\citep{lu2025scamo,wang2025motionlib,fan2025motionmillion,rempe2026kimodo,li2026anymo}.
For physical control, SONIC and Humanoid-GPT investigate scaling motion data, policy capacity, and simulation compute, with related systems further expanding training corpora and controller architectures~\citep{luo2025sonic,qi2026humanoid,chen2026holomotion,zeng2026scalingbfm}.
In contrast, \movensdata enables us to study whether scaling multimodal, robot-executable motion data can jointly improve semantic generation and physical execution.

\section{Movens-5M}
\label{sec:omnihm}

Internet videos provide a scalable source of diverse human behaviors, but raw web footage cannot be used directly as humanoid training data.
In-the-wild videos contain camera motion, shot transitions, occlusion, and uncertain scale, gravity, and coordinate frames, all of which make world-grounded motion recovery inherently noisy.
More fundamentally, humans and humanoid robots are heterogeneous embodiments with different body proportions, joint structures, ranges of motion, and actuation constraints.
Consequently, a motion that appears plausible in a human representation may still contain physical artifacts, transfer poorly to a humanoid, or become unstable under closed-loop tracking.
\movensdata addresses these challenges with the scalable data engine in Figure~\ref{fig:pipeline}, progressively converting raw Internet videos into multimodally aligned and physically executable humanoid motion data.
This section first introduces the scalable data construction pipeline (Section~\ref{sec:omnihm_pipeline}), then describes data augmentation for expanding aligned supervision (Section~\ref{sec:data_augmentation}), summarizes dataset scale and diversity (Section~\ref{sec:dataset_composition}), and finally validates motion quality and humanoid executability (Section~\ref{sec:data_quality_validation}).

\subsection{Scalable Data Construction Pipeline}
\label{sec:omnihm_pipeline}

As shown in Figure~\ref{fig:pipeline}, \movensdata converts raw Internet videos into multimodally aligned and physically executable humanoid trajectories through five stages: clip generation, quality filtering, semantic annotation, motion recovery, and physical refinement.
The stages are ordered from inexpensive media checks to costly motion reconstruction and simulation, allowing unsuitable samples to be removed early without narrowing the collection to a small set of actions or scenes.
Throughout the pipeline, a stable clip identity links the source video, temporal offsets, annotations, recovered human motion, retargeted humanoid reference, and execution record, making the final dataset both scalable to assemble and straightforward to audit.

\begingroup
\titleformat{\subsubsection}
  {\fontsize{11pt}{13.2pt}\selectfont\sffamily\bfseries\color{ZJUBlue}}
  {\thesubsubsection}{0.65em}{}

\paragraph{Clip generation.}
This stage transforms untrimmed source videos into valid, coherent clips while preserving the provenance needed for leakage-free dataset splits.
It first validates basic media integrity and then partitions long recordings at shot boundaries so that each retained clip captures a temporally coherent event.
Source-level splitting, near-duplicate removal, and stable clip identities make these inputs scalable to process without contaminating evaluation sets.
\textit{\textbf{(a)~Source collection and media validation.}}
We crawl videos from diverse Internet sources and first reject malformed files or samples that fail basic duration, resolution, and frame-rate requirements.
The packaged clip is checked for a readable video stream, positive duration, and consistent frame-rate metadata; when audio is present, its availability and temporal alignment are recorded explicitly rather than silently treating a missing track as valid supervision.
\textit{\textbf{(b)~Shot-aware segmentation and split integrity.}}
Since a long video often mixes edits and unrelated activities, AutoShot~\citep{zhu2023autoshot} detects shot boundaries and divides it into temporally coherent clips before expensive multimodal inference.
Training and evaluation splits are formed at the original-video level before temporal slicing, and near-duplicate clips are removed across splits.
Every descendant clip retains its source identifier and frame offset, preventing adjacent segments or re-uploaded footage from leaking nearly identical motions across splits.

\paragraph{Quality filtering.}
This stage combines visual-quality screening with human-motion diagnostics so that polished but motion-poor footage is not confused with useful behavior.
Perceptual and semantic filters remove degraded or irrelevant footage, while person-aware temporal statistics distinguish articulated movement from edits, camera motion, and playback artifacts.
Requiring clips to pass both branches yields reconstruction-ready inputs without systematically discarding valid fast actions.
\textit{\textbf{(a)~Visual quality.}}
Raw clips are screened along complementary perceptual, semantic, and temporal dimensions.
DOVER~\citep{wu2023dover} and auxiliary diagnostics identify severe aesthetic or technical degradation at low cost.
Qwen3-VL~\citep{bai2025qwen3vl} then evaluates whether a clip depicts a real, sufficiently visible person whose body motion is central to the footage, excluding animation, gameplay, static interfaces, and unlocalized crowds while retaining diverse live-action activities.
We aggregate the model's normalized yes/no likelihood and keep clips with human-relevance score $s_{\mathrm{human}}\geq0.90$.
A separate temporal-rate classifier labels each clip as slow, normal, or fast and retains the slow and normal classes.
Here, temporal rate measures playback and encoding validity---for example, whether frames appear unnaturally accelerated---rather than the physical speed or difficulty of the depicted action.
\textit{\textbf{(b)~Human-motion metrics.}}
Visual quality alone cannot distinguish genuine articulated motion from a hard cut or a moving camera.
We therefore estimate dense optical flow with RAFT~\citep{teed2020raft} and use person masks to separate motion within the human region from motion in the background.
The signals are normalized by frame rate and tested with robust local statistics, including spike strength, temporal isolation, and person-mask overlap.
This human/background decomposition detects cuts, subject disappearance or teleportation, and abrupt background changes without automatically rejecting a valid fast limb movement.
Clips are retained only when none of the robust discontinuity tests fires; the detailed branch definitions and thresholds are reported in Appendix~\ref{app:motion_quality}.

\paragraph{Semantic annotation.}
This stage converts each retained clip into complementary video-level, motion-level, and multimodal conditions grounded on a shared timeline.
It combines factual video descriptions, localized body-motion captions, structured motion and scene labels, and modality-specific audio and text conditions.
The resulting representation preserves sample identity and temporal provenance so that every downstream training window can retrieve the condition aligned with its motion.
\textit{\textbf{(a)~Video caption and human features.}}
For every retained clip, a vision-language model jointly produces a factual two-to-four-sentence video caption and structured human features from the same visual evidence.
The caption summarizes the visible subjects, scene, interactions, salient objects, and overall activity, while excluding inferred intent or events that cannot be verified from the frames.
The human record captures the number of visible people, stable within-clip subject identities, body visibility, and other visually observable attributes, so that multi-person footage retains actor-specific context rather than collapsing all behavior into one description.
Both outputs preserve the source identity and clip-level temporal span, providing global context for the localized motion timeline described next and a consistent conditioning record for downstream training windows.
\textit{\textbf{(b)~Motion caption.}}
The timeline describes both whole-body behavior and body-part motion, including the upper limbs, lower limbs, hands, and face, and identifies objects that are visibly manipulated by a hand or foot.
An annotation validator enforces chronological, contiguous, and non-overlapping coverage, merges unreliable micro-segments when appropriate, and requests regeneration when coverage or duration checks fail.
Closed-set motion and scene labels complement the free-form captions with a consistent hierarchy for sampling and evaluation.
\textit{\textbf{(c)~Multimodal synchronization and text rewriting.}}
Audio is processed on the same timeline to distinguish speech, music, and other task-relevant acoustic content.
Task-specific prompts separate linguistic intent from rhythmic conditioning, while audio--visual consistency is stored as a diagnostic rather than used as a restrictive hard gate.
Meaning-preserving text rewriting produces linguistically varied captions without changing the underlying action or its temporal extent.
Text, audio, RGB observations, recovered body parameters, and later execution records all preserve timestamps and provenance, so a training window can retrieve the condition that actually corresponds to its motion instead of relying on a coarse video-level label.

\paragraph{Motion recovery.}
This stage reconstructs each visible person and standardizes heterogeneous estimator outputs for downstream physical refinement.
Complementary single-person and multi-person estimators recover global human motion, after which their outputs are canonicalized into a shared SMPL-X representation.
Body-local articulation checks then remove sequences dominated by camera or root motion before expensive retargeting.
\textit{\textbf{(a)~Single- and multi-person recovery.}}
Single-person clips are processed with GVHMR~\citep{shen2024gvhmr} to recover gravity-aligned global trajectories, whereas multi-person clips are processed with PromptHMR~\citep{wang2025prompthmr} to recover individuals in crowded, occluded, and interaction-heavy scenes.
\textit{\textbf{(b)~Canonicalization and articulated-motion validation.}}
Their complementary outputs are canonicalized into a common SMPL-X representation~\citep{8953319} that preserves root trajectory, body pose, shape, coordinate transformations, and frame timestamps.
After removing global root translation and orientation, we measure body-local motion at the distal joints and normalize it by body scale.
Sequences with articulation score $A_{\mathrm{limb}}<0.08$ are removed, preventing camera or root motion from being mistaken for meaningful limb movement while retaining actions with limited global displacement.

\paragraph{Physical refinement.}
As depicted in Figure~\ref{fig:pipeline}, physical refinement comprises two closed loops that progressively turn recovered human motion into an executable humanoid trajectory.
The first detects local artifacts and repairs recoverable SMPL-X segments, while the second retargets the refined motion and verifies it through closed-loop humanoid tracking.
This separation preserves challenging behaviors when possible but rejects candidates that remain physically implausible or unexecutable.
\textit{\textbf{(a)~Human-motion assessment and repair.}}
Recovered SMPL-X sequences are divided into four-second chunks with a 0.5-second overlap and evaluated for five complementary artifact families: implausible ground contact, temporal jitter, fall risk, self-interpenetration, and foot sliding, following the physically grounded assessment principles of PHUMA~\citep{lee2025phuma}.
The overlap exposes defects near chunk boundaries that an independent-window test could miss.
When a failed interval has reliable context, Kimodo~\citep{rempe2026kimodo} conditionally regenerates only that region from trusted boundary and interior keyframes.
We test multiple repair seeds, stitch each proposal into the original sequence, and reapply the same checks to the target and its neighboring intervals.
This local repair loop preserves difficult but recoverable motions; sequences with persistent failures are rejected rather than being concealed by global smoothing.
\textit{\textbf{(b)~Physics-aware retargeting and execution verification.}}
The refined human motion is transformed into the robot coordinate frame and initialized with General Motion Retargeting (GMR)~\citep{araujoRetargetingMattersGeneral2025} using morphology-aware scaling and sparse human-to-robot position and orientation correspondences.
We then optimize root translation and actuated joints jointly over the sequence instead of solving every frame independently.
The sequence objective balances motion fidelity and temporal consistency with conservative joint position and velocity limits, ground-contact preservation, and foot-skating penalties.
This stage bridges differences in body proportions, joint structures, coordinate conventions, and contact geometry while remaining efficient enough for dataset-scale processing.
Kinematic plausibility does not guarantee that a humanoid controller can execute the resulting reference.
We therefore replay each retargeted candidate in simulation and evaluate it with a controller-independent score that combines average and tail joint-tracking errors.
We retain each trajectory only if its selected tracking-policy rollout satisfies $q\geq0.20$.
For full-bank validation, up to four independently trained tracking policies---SONIC, Humanoid-GPT, HoloMotion, and OmniXtreme~\citep{luo2025sonic,qi2026humanoid,chen2026holomotion,wang2026omnixtreme}---may be evaluated, and one successful rollout among the evaluated policies is sufficient.
A fall or large tracking failure can trigger another retargeting seed before final rejection.
The retained example stores the selected rollout together with the available policy scores and verification statistics, so dataset scale reflects controller-executable motions rather than visually plausible reconstructions alone.

\endgroup

\subsection{Data Augmentation}
\label{sec:data_augmentation}

\begin{figure*}[t]
  \centering
  \includegraphics[width=\textwidth]{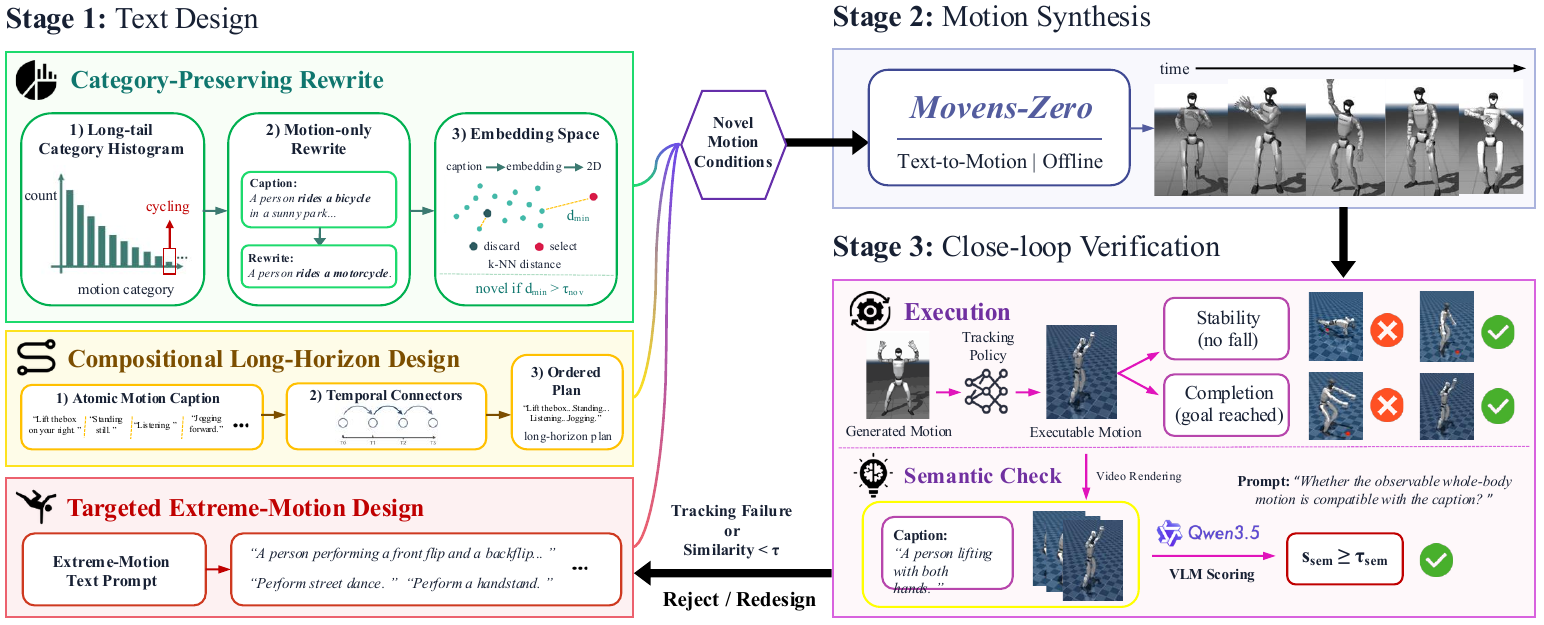}
  \caption{Overview of our data-augmentation pipeline. Text design expands motion conditions through category-preserving rewriting, long-
  horizon composition, and targeted extreme-motion design; \movenszero synthesizes the corresponding robot trajectories; and closed-loop
  tracking with semantic-consistency verification retains executable caption--motion pairs.}
  \label{fig:augmentation}
\end{figure*}

Even after web-scale curation, the motion distribution remains biased toward frequent, short, and routine behaviors, leaving rare fine-grained types, long-horizon compositions, and highly dynamic motions under-represented.
As illustrated in Figure~\ref{fig:augmentation}, we augment \movensdata through a three-stage generate-and-verify pipeline: text design expands category, temporal, and difficulty coverage; motion generation realizes these conditions; and closed-loop verification retains only executable caption--motion pairs whose tracked motions remain semantically consistent with their conditioning captions.

\paragraph{Text design.}
Text design broadens \movensdata along three complementary axes while keeping every condition concrete and suitable for motion synthesis.
Category-preserving rewriting improves coverage of under-represented fine-grained behaviors, compositional design extends supervision to long-horizon action sequences, and targeted extreme-motion design introduces dynamic and contact-rich cases that are rarely observed online.
Together, these strategies expand semantic, temporal, and difficulty coverage before motion synthesis and closed-loop verification.
\textit{\textbf{(a)~Category-preserving motion rewriting.}}
We count the support of every fine-grained motion type and prioritize under-represented categories.
For each selected sample, Qwen3.5-27B distills the raw video caption into an action-centric description, removing appearance, scene, and cinematographic attributes while preserving the executable human action.
Conditioned on the exact fine-grained type as a semantic boundary, it then changes the concrete motion instance---such as its technique, trajectory, action-critical object, or body mechanics---rather than merely paraphrasing the source.
To retain genuinely novel descriptions, we jointly encode the original caption set $\mathcal{C}$ and rewritten set $\widetilde{\mathcal{C}}$ with Qwen2.5-Omni-3B~\citep{xu2025qwen25omni}, and map their text embeddings into a shared two-dimensional space using t-SNE~\citep{van2008visualizing}.
For a rewritten caption $\tilde{c}$, its novelty is
\begin{equation}
    d_{\mathrm{nov}}(\tilde{c})
    = \min_{c\in\mathcal{C}}
      \left\|\pi(\tilde{c})-\pi(c)\right\|_2,
    \label{eq:caption_novelty}
\end{equation}
where $\pi$ denotes the joint projection.
We calibrate the threshold from nearest-neighbor distances among original captions and retain candidates above it after removing near-copies and non-motion descriptions.
\textit{\textbf{(b)~Compositional long-horizon design.}}
To move beyond isolated atomic clips, we organize three to six sub-actions into temporally ordered plans and compose their segment captions into multi-stage instructions.
Enumerating such plans produces complex long-horizon supervision that is rarely observed end-to-end in raw videos.
\textit{\textbf{(c)~Targeted extreme-motion design.}}
To cover high-dynamic and contact-rich behaviors that are scarce online~\citep{wang2026omnixtreme}, we construct targeted descriptions involving rapid support transitions, inverted poses, high-speed rotations, extreme balance, and large-range whole-body maneuvers.
This strategy expands the difficulty axis beyond the class-frequency and temporal coverage of the first two.

\paragraph{Motion generation.}
The designed captions and action plans condition the text-to-motion mode of \movenszero to synthesize robot-centric trajectories.
Atomic and extreme conditions produce individual motions; compositional plans are generated segment by segment while carrying the terminal robot state, state history, and motion prefix across boundaries, and are concatenated without resetting the humanoid between sub-actions.

\paragraph{Closed-loop verification.}
A synthesized trajectory may be kinematically plausible yet become unstable or drift from its intended semantics during closed-loop tracking.
We roll out each candidate with the motion-tracking policies and reject unstable executions under the same tracking criterion as the main curation pipeline.
For each successful rollout, we directly present the rendered video $v$ together with its conditioning caption $\tilde{c}$ to a vision-language model.
Because robot-only renderings omit manipulated objects and scene context, the comparison prompt asks whether the observable whole-body motion is compatible with the caption, without penalizing object references, appearance, background, or camera attributes that cannot be recovered from the rendering.
Let $z_{\mathrm{yes}}$ and $z_{\mathrm{no}}$ denote the output logits for the constrained responses ``Yes'' and ``No,'' respectively. We define the semantic-consistency score as
\begin{equation}
    s_{\mathrm{sem}}(v,\tilde{c})
    = \frac{\exp(z_{\mathrm{yes}})}
    {\exp(z_{\mathrm{yes}})+\exp(z_{\mathrm{no}})},
    \label{eq:semantic_consistency}
\end{equation}
and retain only candidates that pass the tracking criterion and satisfy $s_{\mathrm{sem}}\geq\tau_{\mathrm{sem}}$.

\begin{figure*}[t]
    \centering
    \includegraphics[width=\textwidth]{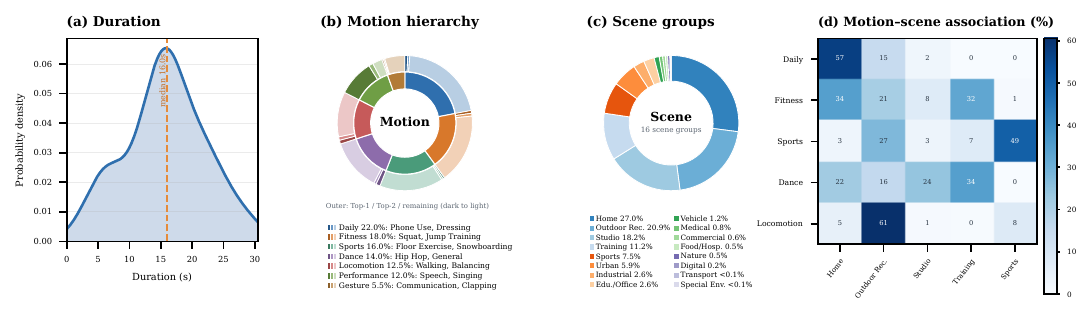}
    \caption{Statistics of \movensdata. (a) Empirical sequence-duration density. (b) Hierarchical motion composition: the inner ring shows seven coarse classes, and the outer ring shows their two largest fine-grained types plus the remainder. (c) Composition of the 16 scene groups. (d) Motion--scene association among the five largest retained motion classes and scene groups; each cell gives the percentage of a motion class occurring in the corresponding scene group.}
    \label{fig:dataset_statistics}
\end{figure*}

\subsection{Dataset Statistics}
\label{sec:dataset_composition}

We characterize \movensdata at three complementary levels: overall scale, semantic and scene composition, and motion-space distribution.
Rather than directly aggregating existing motion repositories, \movensdata is constructed from raw Internet videos and preserves temporal correspondences among video, text, speech, music, human motion, and humanoid motion within a unified pipeline.
As shown in Table~\ref{tab:dataset_comparison}, the final dataset contains 15M human-motion sequences and 5M verified humanoid-motion sequences, totaling more than 10,000 hours and 1.08B frames, and covering 297 motion types and 16 coarse-grained scenes.
The 15M/5M figures in Table~\ref{tab:dataset_comparison} count samples produced by the complete pipeline at the human-motion and humanoid-motion levels.
We summarize the temporal, categorical, and contextual distributions in Figure~\ref{fig:dataset_statistics}.

\begin{figure*}[t]
    \centering
    \includegraphics[width=\textwidth]{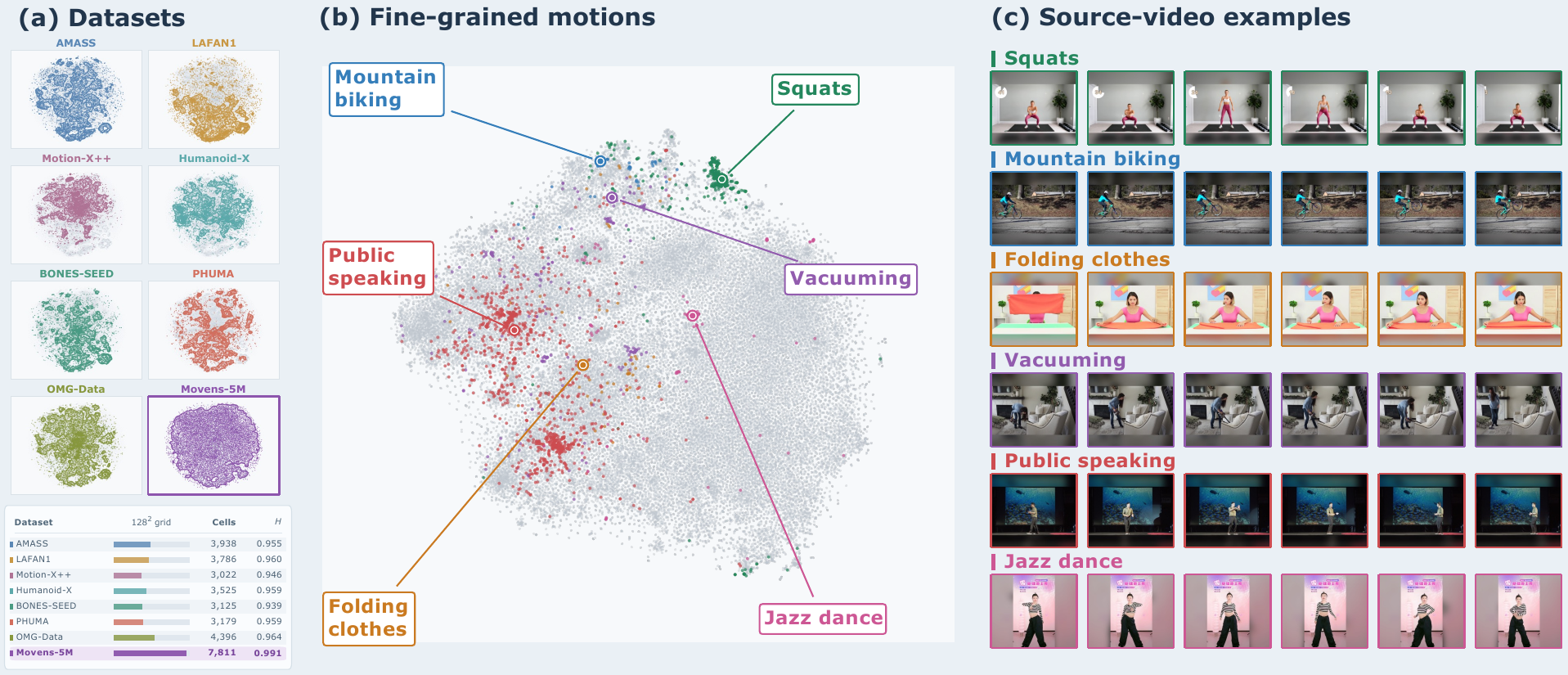}
    \caption{Motion-space visualization and examples. (a) Shared two-second G1 motion-space projection for eight datasets. Each colored view contains 10,000 windows; the seven comparators are random samples, while the Movens-5M display subset is grid-prioritized on frozen coordinates. (b) Fine-grained motion types in Movens-5M. (c) Corresponding source videos. See Appendix~\ref{app:dataset_examples} for more detailed dataset examples.}
    \label{fig:dataset_motion_tsne}
\end{figure*}

\paragraph{Motion-space visualization.}
Figure~\ref{fig:dataset_motion_tsne} complements the categorical statistics with t-SNE views~\citep{vandermaaten2008visualizing} at two scales.
Panel (a) compares AMASS~\citep{mahmood2019amass}, LAFAN1~\citep{harvey2020robust}, Motion-X++~\citep{zhang2025motionxpp}, Humanoid-X~\citep{mao2024learning}, BONES-SEED~\citep{bones2026seed}, PHUMA~\citep{lee2025phuma}, OMG-Data~\citep{huang2026omg}, and \movensdata in one shared G1 motion space.
The seven comparator datasets each contribute 10,000 seed-42 two-second windows from locally available eligible motion data.
Panels (b,c) visualize sampled motions and corresponding source-video examples.
We select six concrete motion types for their task relevance and visual clarity, considering within-class compactness while retaining every member of each highlighted category, including peripheral points.
The cross-dataset descriptor preserves the 61-frame temporal sequence of all 29 actuated G1 joints using per-window mean-centered joint angles and joint velocities; global root pose, dataset identity, text, and video appearance do not enter the descriptor or final projection objective.
The common-G1 support visualization and the complete-clip category map are distinct projections.

\subsection{Data Quality}
\label{sec:data_quality_validation}

Scale and semantic diversity do not guarantee that data are suitable for humanoid training. Motions recovered from Internet videos may appear plausible in the image while containing artifacts such as floating, temporal jitter, body interpenetration, and foot sliding. Even when the human motion itself is valid, differences between humans and robots in body proportions, joint ranges, and contact geometry may make the retargeted motion impossible for a closed-loop controller to track stably~\citep{lee2025phuma}.
Rather than summarizing quality with a single visual score, we therefore sequentially validate the physical and temporal validity of recovered human motion and the controller-level executability of retargeted humanoid trajectories.
The first level determines whether an input motion is worth transferring, while the second determines whether the transferred reference trajectory truly belongs to the learnable and executable robot behavior space.

\paragraph{Human motion quality.}
Following physically grounded motion-filtering principles~\citep{lee2025phuma}, we divide each recovered SMPL-X sequence into four-second chunks with a 0.5-second overlap between adjacent chunks, reducing the chance that artifacts near segment boundaries are missed by independent windows.
Each chunk undergoes five complementary checks. Ground contact identifies floating motions that lack reliable foot support for extended periods. Root jerk and positional high-frequency ratios detect unnatural high-frequency jitter in global and local joint trajectories. Pelvis height and the distance between the body projection and the support base characterize falling or imbalance risk. The maximum penetration of colliding triangle pairs and accumulated within-frame penetration capture deep localized and shallow widespread body self-intersections, respectively. A physical foot-contact score detects abnormal simultaneous sliding of both feet along the ground during body acceleration.
Together, these metrics constrain contact, temporal continuity, balance, geometric validity, and support consistency, preventing one averaged score from obscuring a severe artifact of another type.

Filtering is not a one-pass deletion operation.
When a failed chunk has at least one reliable neighboring context, we use Kimodo to conditionally regenerate the local interval while preserving trusted boundary and interior keyframes. The system searches multiple random seeds, stitches each candidate back into the original sequence, and reapplies the same checks to the target interval and its required neighborhood.
Only candidates that pass re-evaluation are written back to the dataset. Motions that repeatedly fail or lack reliable context are removed, ensuring that local repair does not conceal artifacts through global smoothing.
Complete definitions, thresholds, and repair constraints for the five metric families are provided in Appendix~\ref{app:motion_quality}, with specific thresholds summarized in Table~\ref{tab:filtering_thresholds}.

\paragraph{Humanoid executability.}
After passing the human-motion quality checks, each motion is first initialized kinematically through morphology-aware scaling and human--robot keypoint correspondences. We then jointly optimize motion fidelity, temporal consistency, joint position and velocity limits, ground contact, and foot sliding over the entire sequence.
This stage converts human motion semantics into a reference trajectory compatible with the robot embodiment and hardware constraints, but kinematic constraints alone cannot establish closed-loop executability.
Each candidate trajectory is replayed in simulation and retained only if its selected tracking-policy rollout satisfies $q\geq0.20$.
Full policy-bank validation can further evaluate four independently trained controllers---SONIC, Humanoid-GPT, HoloMotion, and OmniXtreme---and accepts a candidate when at least one evaluated rollout satisfies the same criterion. A fall or large tracking error may trigger a new retargeting seed before final rejection.
The common criterion makes rollout records from different controllers directly comparable and avoids mistaking controller-specific bias for the quality of the data itself.
The objectives, post-processing, and per-policy execution-verification protocol for physics-aware motion retargeting are provided in Appendix~\ref{app:retargeting_verification}.

\paragraph{Quantitative comparison.}
Table~\ref{tab:dataset_motion_quality} compares the motion quality of \movensdata with MoCap-based and video-based references, using SONIC for the execution measurements.
The table compares sampled G1 reference motions and their SONIC rollouts under a common evaluation pipeline. Per-metric sample counts are reported in the coverage table.
Fall and joint-limit rates include every completed rollout, including failures; they should be read together with the effective execution counts in Table~\ref{tab:dataset_quality_coverage}.
The tracking measurements complement the reference-motion criteria; the low Non-Floating score identifies foot-ground contact as a remaining limitation.
Evaluation settings, metric definitions, and dataset-specific qualifications are detailed in Appendix~\ref{app:dataset_quality_comparison}.

\begin{table*}[!htbp]
\centering
\scriptsize
\setlength{\tabcolsep}{1.6pt}
\renewcommand{\arraystretch}{1.15}
\caption{Complete-sequence dataset motion quality and G1 execution across MoCap-based and video-based datasets. Scale columns report full datasets or releases. Bold marks within-group best results. Higher MSNR means less high-frequency residual relative to the smoothed motion signal, but it should be interpreted jointly with jerk because motion bandwidth and over-smoothing also affect it. \textemdash\ denotes unavailable entries. Evaluation settings and dataset-specific qualifications are detailed in Appendix~\ref{app:dataset_quality_comparison}.}
\label{tab:dataset_motion_quality}
\resizebox{\textwidth}{!}{\begin{tabular}{@{}lrrr|*{8}{c}|*{7}{c}@{}}
\toprule
\multirow{3}{*}{\textbf{Dataset}} &
\multicolumn{3}{c|}{\textbf{Dataset Scale}} &
\multicolumn{8}{c|}{\textbf{Reference Motion Quality}} &
\multicolumn{7}{c}{\textbf{Closed-loop Rollout Quality}} \\
\cmidrule(lr){2-4}\cmidrule(lr){5-12}\cmidrule(lr){13-19}
& \multirow{2}{*}{\textbf{Hours}} & \multirow{2}{*}{\textbf{Frames}} &
\multirow{2}{*}{\shortstack{\textbf{\# Humanoid}\\\textbf{Sequences}}} &
\multicolumn{2}{c}{Temporal Quality} & \multicolumn{5}{c}{Foot--Support Consistency} &
\multicolumn{1}{c|}{Kinematic} & \multicolumn{5}{c}{Tracking Fidelity} &
\multicolumn{2}{c}{Execution Validity} \\
\cmidrule(lr){5-6}\cmidrule(lr){7-11}\cmidrule(lr){12-12}\cmidrule(lr){13-17}\cmidrule(lr){18-19}
& & & & \textbf{Jerk$\downarrow$} & \textbf{MSNR$\uparrow$} &
\textbf{C-Slide$\downarrow$} & \textbf{PFC$\downarrow$} &
\shortstack{\textbf{Non-}\\\textbf{Skating$\uparrow$}} &
\shortstack{\textbf{Non-}\\\textbf{Floating$\uparrow$}} &
\shortstack{\textbf{Non-}\\\textbf{Pen.$\uparrow$}} &
\shortstack{\textbf{Joint}\\\textbf{Feas.$\uparrow$}} &
\textbf{MPJPE$\downarrow$} & \textbf{g-MPJPE$\downarrow$} &
\textbf{$E_{\mathrm{vel}}\downarrow$} & \textbf{$E_{\mathrm{acc}}\downarrow$} &
\shortstack{\textbf{Tracking}\\\textbf{Score$\uparrow$}} &
\textbf{Fall$\downarrow$} & \textbf{J-Limit$\downarrow$} \\
\midrule
\multicolumn{19}{@{}l}{\textit{MoCap-based datasets}} \\
AMASS~\citep{mahmood2019amass} & 40.0 & $\approx$4.32M & \textemdash & 121.55 & 14.01 & 0.345 & 0.916 & 71.17 & 72.97 & 75.82 & 75.61 & 41.59 & 695.83 & 0.326 & 4.260 & 0.292 & 7.10 & 19.40 \\
LAFAN1~\citep{harvey2020robust} & 4.6 & 496.7K & \textemdash & 194.32 & 11.65 & 0.356 & 0.798 & 46.63 & 23.88 & 97.55 & 41.27 & 153.28 & 3028.86 & 0.592 & 7.713 & 0.084 & 63.64 & 98.70 \\
OmniRetarget~\citep{yang2025omniretarget} & 4.0 & $\approx$432K & 2.1K & 204.61 & 13.11 & 0.221 & \textbf{0.250} & 80.69 & 66.27 & \textbf{100.00} & \textbf{100.00} & 59.23 & 443.92 & 0.322 & 5.866 & 0.197 & 13.70 & 32.30 \\
BONES-SEED~\citep{bones2026seed} & 288.3 & 124.6M & 142.2K & 63.68 & 16.87 & \textbf{0.198} & 0.370 & \textbf{88.38} & 86.38 & 95.84 & 74.17 & 30.90 & \textbf{170.27} & \textbf{0.182} & \textbf{2.910} & 0.330 & 5.60 & \textbf{12.80} \\
PHUMA~\citep{lee2025phuma} & 73.0 & 7.88M & 76.0K & \textbf{31.27} & \textbf{27.31} & 0.247 & 0.414 & 67.79 & \textbf{92.05} & 84.29 & 99.97 & \textbf{27.52} & 198.66 & 0.259 & 3.181 & \textbf{0.372} & \textbf{1.10} & 14.60 \\
\midrule
\multicolumn{19}{@{}l}{\textit{Video-based datasets}} \\
Motion-X++~\citep{zhang2025motionxpp} & 180.9 & 19.5M & \textemdash & 230.24 & 10.27 & 0.220 & 1.220 & 58.19 & 64.21 & 78.10 & 72.83 & 96.12 & 710.35 & 0.450 & 8.331 & 0.207 & 31.30 & 41.60 \\
Humanoid-X~\citep{mao2024learning} & 240.3 & 20.7M & 163.8K & 1961.18 & 9.49 & 1.591 & 35.718 & 4.78 & 50.19 & 86.09 & 53.33 & 129.77 & 895.26 & 1.521 & 49.231 & 0.160 & 46.70 & 62.30 \\
OMG-Data~\citep{huang2026omg} & 1122.0 & 121.2M & 798.2K & 91.74 & 13.32 & 0.221 & 0.189 & 73.73 & 62.87 & 84.90 & 68.46 & 36.05 & 219.33 & 0.255 & 4.278 & 0.307 & 3.10 & 19.90 \\
\textbf{\movensdata~(Ours)} & \textbf{$>$10,000} & \textbf{$>$1.08B} & \textbf{5M} & \textbf{71.69} & \textbf{23.99} & \textbf{0.203} & \textbf{0.068} & \textbf{75.56} & \textbf{90.05} & \textbf{100.00} & \textbf{100.00} & \textbf{28.50} & \textbf{170.71} & \textbf{0.214} & \textbf{3.154} & \textbf{0.325} & \textbf{0.00} & \textbf{5.71} \\
\bottomrule
\end{tabular}}
\end{table*}

Human-motion screening and humanoid-execution verification assess whether the recovered human motion is physically credible and whether the robot can faithfully execute the corresponding behavior.
Passing only the first level would mix embodiment-incompatible or untrackable trajectories into the training set, while passing only the second could admit reference motions that an individual controller can barely track despite reconstruction artifacts. Applying both in sequence ensures that the retained motions simultaneously satisfy human-motion quality, robot constraints, and closed-loop execution requirements.
The final record stores synchronized human motion, the humanoid reference trajectory, the selected simulation rollout, available policy scores, and verification statistics, making every retained sample traceable to concrete quality evidence.
The hard filtering conditions and record-completeness requirements used by the complete data-assembly pipeline are listed in Table~\ref{tab:active_data_gates} of Appendix~\ref{app:dataset_details}.

\section{\movensintent Benchmark}
\label{sec:omnihm_bench}

To evaluate whether a model can translate intent expressed through heterogeneous modalities into executable humanoid motion under a consistent robot representation, we construct \textbf{\movensintent} from carefully curated examples.
The benchmark contains four tracks covering text, speech, music, and video.
All tracks use the Unitree G1 target embodiment, while each retains the natural form and temporal extent of its condition.
The following paragraphs motivate this design, describe how benchmark examples are selected and verified, and summarize the evaluation protocol for each track.

\subsection{Motivation}
A generalist humanoid motion model should understand human intent across text, audio, and video and translate it into corresponding whole-body motion.
Human intent may be conveyed through descriptive language, spoken commands, musical rhythm, or visual demonstrations, yet existing evaluations typically study these interfaces in isolation using different datasets, body representations, temporal horizons, and protocols.
This fragmentation makes it difficult to distinguish a model's general intent-to-motion capability from performance tied to a particular evaluation stack.
We therefore introduce \textbf{\movensintent}, a controlled four-track benchmark that evaluates how models translate intent expressed through text, speech, music, and video into corresponding Unitree G1 whole-body motion.
Each track contains 2,048 condition--motion pairs, giving 8,192 examples in total.
All tracks share the same target embodiment and trajectory convention, while each modality retains its natural condition format, temporal extent, and task-appropriate metrics.
Within a track, every method receives the same conditions and is scored over the same valid reference durations.

\begin{figure*}[!ht]
    \centering
    \includegraphics[width=\textwidth]{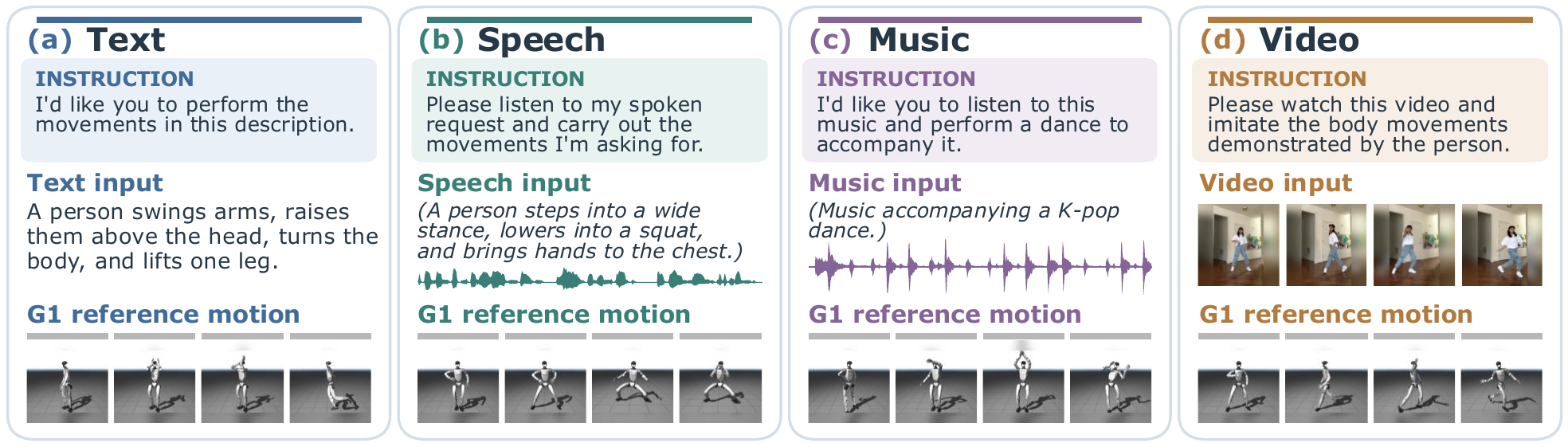}
    \caption{Examples from the \movensintent benchmark, illustrating human intent expressed through (a) text, (b) speech, (c) music, and (d) video. Each panel pairs a user request and its input condition with four frames from the corresponding G1 reference motion. The speech transcript makes the spoken content readable, and the waveforms show the complete audio recordings.}
    \label{fig:intent_benchmark_overview}
\end{figure*}

Figure~\ref{fig:intent_benchmark_overview} illustrates four ways a user can communicate a desired behavior to a humanoid.
A user may describe a movement sequence in writing, issue a spoken request, ask the robot to dance to music, or show a video of the behavior to imitate.
Across tracks, the model must interpret the intended behavior from the corresponding condition and translate it into Unitree G1 whole-body motion.
The paired G1 references provide motion targets for evaluation; Appendix~\ref{app:intent_examples} presents additional cases and more detailed reference sequences.

\subsection{Benchmark Construction}

The benchmark consists of expert-selected condition--motion pairs with no overlap with \movensdata.
Selection operates on stable source identities and combines condition quality, motion validity, execution quality, and category-aware sampling.
Every candidate must contain a complete and temporally aligned condition--motion pair, readable source media, complete metadata, and a finite, non-static Unitree G1 trajectory that has passed physical refinement and tracking verification.
Track-appropriate quality signals rank the surviving candidates, while source- and category-aware caps prevent near-duplicate clips and frequent behaviors from dominating the benchmark.
We retain complete valid sequences rather than forcing every modality into an arbitrary common crop.

For text, we select complete single-person motions with high visual--language consistency and rewrite their captions from the full video description and atomic motion timeline.
Each caption is constrained to one concrete present-tense sentence describing visible body motion; scene, appearance, speech, and narrative details are removed, and anatomical left--right statements are checked against the corresponding G1 joint trajectories.
Speech uses a separate source pool under the same motion and caption criteria, and the selected transcripts are rendered with 16 balanced English Qwen3-TTS voices before waveform duration, loudness, clipping, and silence checks.
Music candidates must preserve a complete aligned audio--motion pair and pass single-person, audio, motion, technical, overlay, watermark, and camera-stability checks; BeatAlign-based ranking and category balancing retain well-synchronized examples without narrowing the track to a few dominant dance types.
Video candidates must preserve complete RGB--motion alignment and pass media-integrity, tracking, vision--language, audio--visual, and articulated-motion checks before category-aware selection.

Automated screening defines only the candidate pool: all 8,192 final condition--motion pairs across the four tracks undergo expert inspection before inclusion.
Expert inspection covers condition correctness, motion semantics, temporal alignment, media integrity, reconstruction and retargeting artifacts, contact and foot sliding, joint-limit violations, and the rendered tracking outcome.
Any example with ambiguous intent, incorrect synchronization, visibly implausible motion, or incomplete artifacts is rejected, and every replacement must pass the same automated screening and expert inspection procedure.

Finally, all references are exported at 30\,Hz as
$\mathbf{q}\in\mathbb{R}^{36}$, comprising root position, root quaternion in $wxyz$ order, and 29 G1 joint angles in world $Z$-up coordinates.
Variable-length batches carry an explicit validity mask so padded frames never enter a metric.
Per-track manifests record source identities, temporal spans, quality measurements, expert-review status, selection scores, and media or target hashes, making every released benchmark example auditable.

\subsection{Evaluation Protocol}
Text evaluation follows the OMG embedding evaluator~\citep{huang2026omg}.
We report FID between generated and reference motion embeddings, paired text--motion Matching distance, R@1/2/3 in deterministic batches of 32, and Diversity over fixed-seed pairs of generated embeddings.
These quantities are compared only under an identical cohort and protocol; sample ordering affects retrieval batches but not FID.

Audio evaluation covers the music- and speech-conditioned tracks under protocols appropriate to their distinct acoustic semantics.
For music, we evaluate each complete valid sequence using music-to-motion BeatAlign together with G1-adapted FID-k, FID-g, and PFC~\citep{huang2026omg,tseng2023edge}.
The G1 feature variants are comparable only when the cohort, marker definition, and implementation are held fixed.
Speech is maintained as a separate command-conditioned track: music--dance metrics are not applied to spoken conditions, and a dedicated speech-semantic score is outside the current release.

Video evaluation converts predictions and references through G1 forward kinematics to a shared marker representation and reports world-space WA-MPJPE$_{100}$, W-MPJPE$_{100}$, root-trajectory error, Jitter, and G1 Foot-Sliding following the GVHMR/GENMO protocol~\citep{shen2024gvhmr,li2025genmo}.
The two windowed errors use non-overlapping 100-frame windows, with a shorter residual segment evaluated as one window.
For text and music, C-Slide and body Jerk additionally measure target-space physical quality.
Execution metrics use the canonical HoloMotion single-policy rollout~\citep{huang2026omg}, reporting root-relative and global tracking errors together with velocity, acceleration, fall, and joint-limit statistics.
Detailed metric definitions and settings are provided in Appendix~\ref{app:evaluation_protocols}.

\section{Method}
\label{sec:method}

Building on \movensdata, we introduce \movenszero, a scalable understanding--generation--execution architecture that translates multimodal human intent directly into physically executable humanoid behaviors in real time.
We first define the shared motion representation and formulate overlapping, state-conditioned chunk generation (Section~\ref{sec:problem_formulation}).
We then instantiate this formulation with an understanding--generation--execution architecture and \mmrtc, which couple omni-modal context encoding, flow-based motion generation, tracking, and continuous replanning (Section~\ref{sec:hierarchical_architecture}).
Finally, we introduce a task-mixed training strategy that jointly optimizes a single action expert over a data-proportional mixture of seven tasks using a unified conditional flow-matching objective and clean-prefix training (Section~\ref{sec:training_strategy}).

\begingroup
\let\section\subsection
\let\subsection\subsubsection
\titleformat{\subsubsection}
  {\fontsize{11pt}{13.2pt}\selectfont\sffamily\bfseries\color{ZJUBlue}}
  {\thesubsubsection}{0.65em}{}

\section{Problem Formulation}
\label{sec:problem_formulation}

We formulate omni-modal humanoid control as the repeated generation of short-horizon reference-motion chunks from the current multimodal condition and latest robot state.
We first define the robot-centric representation shared by generated motion and observed state, and then formalize state-conditioned overlapping chunk generation with a retained prefix and repeated state feedback.
Together, these definitions establish the common interface used by the model architecture and training strategy that follow.

\paragraph{Robot-centric motion representation.}
\label{sec:motion_representation}
To provide a common interface between generation and low-level tracking, we express both predicted motion and observed state in a shared 40-dimensional robot-centric representation.
This subsection defines its joint, root-motion, orientation, height, and foot-contact channels, and then describes the coordinate convention, normalization, and state tokenization used by the action expert and tracking controller.
The action expert predicts a 40-dimensional motion token at each frame.
For a humanoid with 29 actuated joints, we write
\begin{equation}
    \mathbf{x}_j=\big[
        \mathbf{q}_j,
        \Delta\mathbf{r}^{\mathrm{anc},xy}_j,
        \boldsymbol{\rho}^{\mathrm{anc}}_j,
        h_j,
        c_j^{\mathrm{L}},c_j^{\mathrm{R}}
    \big]\in\mathbb{R}^{40}.
    \label{eq:robot_motion_representation}
\end{equation}
Here, $\mathbf{q}_j\in\mathbb{R}^{29}$ contains absolute joint angles, $\Delta\mathbf{r}^{\mathrm{anc},xy}_j\in\mathbb{R}^{2}$ is the planar inter-frame root displacement in a gravity-aligned chunk anchor, and $h_j$ is the absolute root height.
At generation step $n$, the anchor rotation $\mathbf{A}_n$ is obtained from the observed root heading and held fixed within the chunk.
The six orientation channels $\boldsymbol{\rho}^{\mathrm{anc}}_j\in\mathbb{R}^{6}$ concatenate the first two columns of $\mathbf{A}_n^{\top}\mathbf{R}_j$, where $\mathbf{R}_j$ is the body-to-world root rotation.
They encode the root orientation relative to the anchor rather than an inter-frame rotation increment.
The final two channels are binary left- and right-foot contact indicators estimated from leg-joint changes and vertical root displacement.
Root height remains in world coordinates, while planar displacement and orientation share the chunk anchor.
Actions and states use the same per-channel affine scaling, with the 1st- and 99th-percentile training-set bounds mapped to $[-1,1]$.
Normalized training targets and state inputs are clipped to this interval.
At inference, the observed state is clipped after the same scaling.
In the current model, the most recent state is projected as one proprioceptive token.
The channel-wise motion semantics, coordinate conventions, and training normalization ranges are detailed in Appendices~\ref{app:method_details} and \ref{app:training_details}.

\paragraph{Closed-loop generation formulation.}
\label{sec:closed_loop_generation}
Using the shared representation, we formalize online control as repeated prediction of overlapping motion chunks conditioned on the current multimodal input and latest robot state.
We first define the horizon, committed segment, retained clean prefix, and progress encoding, and then specify the conditional generation mapping and explain how partial commitment closes the feedback loop.
At generation step $n$, $\mathbf{u}_n$ denotes the multimodal conditioning input and $\mathbf{s}_n$ denotes the latest robot state.
Let $H$ denote the prediction horizon, $H_{\mathrm{c}}$ the number of frames committed for execution at each step, $K=H-H_{\mathrm{c}}$ the overlap length, and $d$ the motion-token dimension.
For all steps after the first, $\mathbf{P}_n\in\mathbb{R}^{K\times d}$ contains the unexecuted tail of the previous prediction and serves as a clean motion prefix.
The model generates a normalized reference-motion chunk as follows:
\begin{equation}
    \hat{\mathbf{X}}_n
    =G_{\theta}\!\left(
        E_{\psi}(\mathbf{u}_n),
        \mathbf{s}_n,
        \mathbf{P}_n,
        \mathbf{g}_n
    \right)
    \in\mathbb{R}^{H\times d},
    \label{eq:closed_loop_generation}
\end{equation}
where $E_{\psi}$ is a shared omni-modal encoder, $G_{\theta}$ is the action expert, and $\mathbf{g}_n$ denotes a frame-wise progress encoding that indicates each predicted frame's position within the motion sequence.
Here, $d=40$. Each frame in $\hat{\mathbf{X}}_n$ and $\mathbf{P}_n$, as well as the state $\mathbf{s}_n$, follows the normalized robot-centric layout of $\mathbf{x}_j$ defined in Equation~\ref{eq:robot_motion_representation}.
Only the first $H_{\mathrm{c}}$ frames are released to the tracking controller before the system observes a new state and generates the next chunk; adjacent predictions therefore overlap by $K$ frames.

\section{Model Architecture}
\label{sec:hierarchical_architecture}

Having defined the closed-loop problem, we instantiate \movenszero as a scalable understanding--generation--execution architecture that translates multimodal human intent directly into physically executable humanoid behaviors in real time through the two coupled components shown in Figure~\ref{fig:model}.
The understanding--generation--execution hierarchy converts multimodal inputs into context tokens, predicts robot-centric reference motion with a flow-based action expert, and executes the result with a high-frequency tracking controller.
\mmrtc connects this hierarchy over time by retaining the unexecuted motion prefix while refreshing the multimodal condition and robot state at every generation step.

\begin{figure*}[t]
\centering
\includegraphics[width=\linewidth]{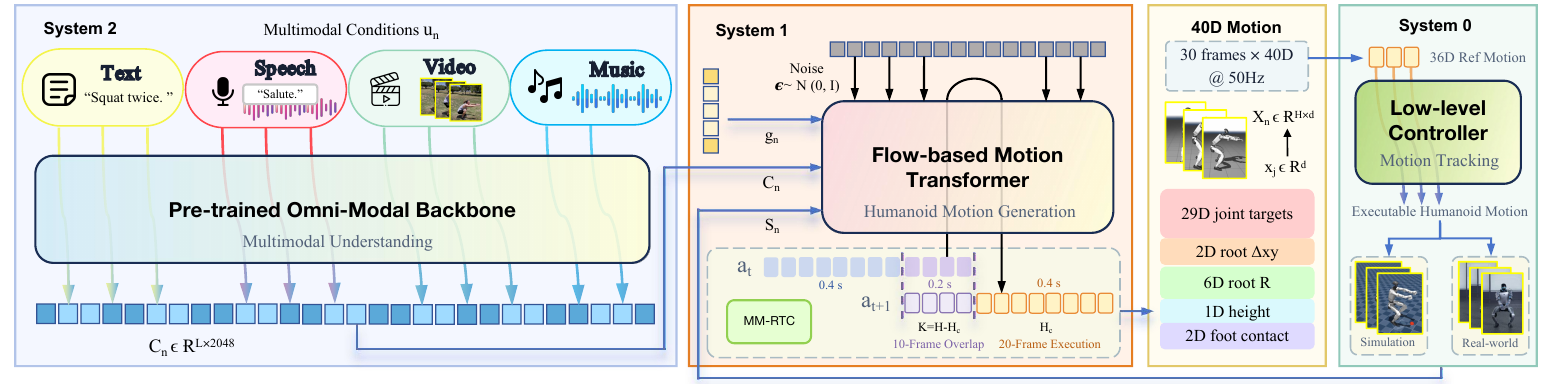}
\caption{Architecture of \movenszero. Its understanding--generation--execution hierarchy translates multimodal human intent directly into physically executable humanoid behaviors in real time: an omni-modal backbone encodes interleaved multimodal inputs, infers human intent, and produces context tokens; a flow-based transformer generates overlapping state-conditioned motion chunks; and a low-level controller executes the resulting references in simulation and the real world.}
\label{fig:model}
\end{figure*}

\paragraph{Understanding--generation--execution.}
\label{sec:understanding_generation_execution}
The architecture contains three stages connected through the shared robot-centric motion interface.
We first describe how the omni-modal backbone encodes text, video, and audio into context tokens, then explain how the joint-attention action expert combines this context with the current robot state to generate reference motion, and finally detail how the tracking controller executes the reference and returns an updated state to the generator.
\textit{\textbf{(a)~Omni-modal context understanding.}}
We initialize $E_{\psi}$ from Qwen2.5-Omni-3B~\citep{xu2025qwen25omni}, whose shared interface accepts language, visual, and audio inputs.
Text instructions are tokenized directly; video conditions are represented by sampled RGB frames, optionally followed by a text prompt; audio conditions, including speech and music, use the audio front end, again with an optional text prompt.
We take the last-layer token sequence $\mathbf{C}_n\in\mathbb{R}^{L_n\times 2048}$, where $L_n$ denotes the number of context tokens, and retain its padding mask.
During action-expert training, the omni-modal encoder is frozen and these token sequences are precomputed, so all optimization capacity is spent on learning the motion generator rather than repeatedly encoding the same conditions.
The cached condition-token dimensions, mask handling, and audio and video preprocessing settings are detailed in Appendix~\ref{app:training_details}.
\textit{\textbf{(b)~Joint-attention action expert.}}
The action expert follows the joint-attention design of $\Psi_0$~\citep{wei2026psi0}; its scale-specific configurations are reported in Appendix Table~\ref{tab:model_scale_architecture}.
Condition tokens are linearly projected from 2048 to the model width $D$, and the 40-dimensional state is appended as an additional context token.
For action position $j$ at flow time $\sigma$ (defined in Section~\ref{sec:training_strategy}), a noisy motion frame $\mathbf{x}^{\sigma}_{n,j}$ is embedded by a two-layer MLP and augmented with a learned positional embedding and a progress residual,
\begin{equation}
    \mathbf{a}^{0}_{n,j}
    =F_{\mathrm{act}}(\mathbf{x}^{\sigma}_{n,j})
     +\mathbf{e}^{\mathrm{pos}}_j
     +F_{\mathrm{prog}}(\mathbf{g}_{n,j})
    \in\mathbb{R}^{D}.
    \label{eq:action_tokenization}
\end{equation}
Here, $\mathbf{g}_{n,j}\in\mathbb{R}^{5}$ provides token-wise temporal progress.
Let $p_{n,j}\in[0,1]$ denote the normalized position of the frame predicted at action position $j$ on the reference-motion timeline, and let $p^{\mathrm{start}}_n$ denote the normalized start position of the current chunk.
We define
\begin{equation}
    \mathbf{g}_{n,j}
    =
    \left[
        p_{n,j},
        1-p_{n,j},
        p^{\mathrm{start}}_n,
        \sin(2\pi p_{n,j}),
        \cos(2\pi p_{n,j})
    \right].
    \label{eq:progress_encoding}
\end{equation}
Unlike the chunk-local positional embedding $\mathbf{e}^{\mathrm{pos}}_j$, this encoding provides explicit temporal context on the reference-motion timeline, helping distinguish similar poses or token positions that occur at different stages of a motion.
Transformer blocks jointly update action tokens and the context--state tokens, with the number of blocks and attention heads determined by the model scale.
The flow time is embedded separately and modulates the action stream through adaptive layer normalization.
The final block reads from the context without unnecessarily updating it, and an adaptive output head maps each action token to a 40-dimensional velocity prediction.
We use zero-initialized adaptive residual gates and output projection, together with per-head Q/K normalization computed in full precision, to stabilize optimization.
\textit{\textbf{(c)~Tracking execution.}}
The predicted 40-dimensional sequence is decoded into joint and root references and sent to a high-frequency whole-body tracking controller.
The generator itself runs at the slower chunk rate.
After each committed segment, the newest robot state becomes the proprioceptive token for the next prediction, closing the feedback loop without requiring the semantic generator to output low-level torques.

\paragraph{Multimodal real-time chunking.}
\label{sec:mmrtc}
Multimodal Real-Time Chunking (\mmrtc) turns chunk-based motion generation into responsive closed-loop execution by alternating overlapping prediction, short-horizon execution, and condition and state updates.
This subsection first defines the commit-and-overlap schedule, then explains how the unexecuted tail becomes a clean prefix for the next prediction, and finally connects this inference pattern to the variable-length prefix training introduced later.
Concretely, each prediction contains $H$ frames, of which only the first $H_{\mathrm{c}}$ frames are committed for execution. The remaining $K=H-H_{\mathrm{c}}$ frames are retained as the prefix of the next chunk. At the next generation step, the retained tail is re-encoded using the latest observed state and the current anchor, with its predicted contact channels preserved. The resulting prefix is kept fixed while the remaining action positions are newly generated under the updated conditioning input and robot state.
To make the action expert support such prefix-conditioned continuation, we expose it to variable-length clean prefixes during training. The corresponding modification to the flow-matching objective is described in Section~\ref{sec:training_strategy}.
The prediction, commitment, and overlap horizons, together with the inference-time flow solver configuration, are provided in Appendix~\ref{app:training_details}.

\section{Training Strategy}
\label{sec:training_strategy}

We jointly train a single action expert in one task-mixed stage. The training data are mixed in proportion to dataset size across seven text-, speech-, music-, and video-conditioned generation tasks, all of which share the robot-centric motion representation, conditional flow-matching objective, and MM-RTC-compatible clean-prefix construction. We next describe the task mixture, optimization objective, and prefix-training procedure.

\paragraph{Seven-task data mixture.}
The training mixture contains seven generation tasks spanning text, video, and audio conditions: text-to-motion; music-to-motion with and without a text instruction; speech-to-motion with and without a text instruction; and video-to-motion with and without a text instruction.
Rather than using equal task weights, each enabled task is weighted by the number of examples in its training split, which we term data-proportional task sampling. The full task mixture and exposure ratios are reported in Appendix Table~\ref{tab:training_task_mix}.
All seven tasks predict the same robot-motion chunk, allowing every batch to update a shared motion prior and condition--action interface.

\paragraph{Conditional flow matching.}
Let $\mathbf{X}$ be a normalized target chunk and $\boldsymbol{\epsilon}\sim\mathcal{N}(\mathbf{0},\mathbf{I})$ be Gaussian noise.
For $\sigma\in[0.05,1)$, we construct the linear probability path
\begin{equation}
    \mathbf{X}^{\sigma}
    =(1-\sigma)\mathbf{X}+\sigma\boldsymbol{\epsilon},
    \qquad
    \mathbf{V}^{\star}=\boldsymbol{\epsilon}-\mathbf{X}.
    \label{eq:flow_path}
\end{equation}
The action expert predicts $\mathbf{V}^{\star}$ from $\mathbf{X}^{\sigma}$, the flow time, the condition tokens, the current state $\mathbf{s}_n$, and future progress $\mathbf{g}_n$.
With validity mask $\mathbf{M}$ and per-channel weights $w_k$, the training objective is
\begin{equation}
    \mathcal{L}_{\mathrm{flow}}
    =\mathbb{E}_{\mathbf{X},\boldsymbol{\epsilon},\sigma}
      \left[
        \sum_{j,k}M_{j,k}w_k
        \left(
          V_{\theta,j,k}-V^{\star}_{j,k}
        \right)^2
      \right].
    \label{eq:flow_objective}
\end{equation}

\paragraph{MM-RTC-compatible prefix training.}
To train the action expert for the prefix-conditioned generation required by MM-RTC (Section~\ref{sec:mmrtc}), we augment the standard flow-matching procedure with randomly sampled clean prefixes. Specifically, we sample a clean-prefix length $\ell$ uniformly from $\{0,\ldots,K\}$.
The first $\ell$ target frames remain clean and receive flow time zero, while the remaining frames follow Equation~\ref{eq:flow_path}:
\begin{equation}
\begin{array}{@{}l@{\;}c@{\;}l@{}}
\widetilde{\mathbf{x}}^{\sigma}_{j}
&=&
\begin{cases}
\mathbf{x}_{j}, & j<\ell,\\
(1-\sigma)\mathbf{x}_{j}+\sigma\boldsymbol{\epsilon}_{j}, & j\geq \ell.
\end{cases}
\\[2pt]
M^{\mathrm{RTC}}_{j,k}
&=&
M_{j,k}\,\mathbf{1}[j\geq \ell].
\end{array}
\label{eq:rtc_training}
\end{equation}
All $H$ positions are processed in one forward pass, but the loss is evaluated only on the unknown suffix.
Randomizing $\ell$ exposes the model to both initial generation without a motion prefix and continuation with different clean-prefix lengths.
The seven-task sampling ratios, loss-channel weights, optimizer, distributed batch size, and validation protocol are summarized in Appendix~\ref{app:training_details}.

\endgroup

\section{Experiments}
\label{sec:experiments}

We organize the experiments around three questions:
\textbf{(1)} How well does \movenszero generate semantically aligned and executable motion from text, video, and audio?
\textbf{(2)} Do larger training subsets of \movensdata and larger models yield predictable gains across these modalities?
\textbf{(3)} How do data curation and modality-aligned training affect performance, and how does scaling change physical executability?
This section first describes the evaluation setup (Section~\ref{sec:experimental_setup}) and then evaluates omni-modal generation (Section~\ref{sec:omnimodal_results}).
It next presents qualitative comparisons (Section~\ref{sec:qualitative_comparisons}) and finally investigates scaling behavior and ablations (Section~\ref{sec:in_depth_analysis}).
Additional real-robot rollouts across all five conditioning interfaces are provided in Appendix~\ref{app:real_robot_cases}.

\subsection{Experimental Setup}
\label{sec:experimental_setup}

\paragraph{Benchmarks.}
We evaluate on the public OMG benchmark, the EMDB and RICH datasets, and the source-disjoint \movensintent benchmark.
For OMG, we use the text- and audio-conditioned tracks from the released benchmark manifests~\citep{huang2026omg}, containing 1,024 text-conditioned and 512 audio-conditioned test samples.
Each sample contains 60 future frames at 30 FPS, corresponding to a two-second prediction horizon.
For direct comparison, we retain OMG's metric definitions and published baseline results, and evaluate \movenszero with the same manifest, evaluator, and HoloMotion execution path.
To prevent contamination, all source identities and near-duplicate motions associated with the OMG test manifests are excluded from \movenszero training data.
For video-conditioned evaluation, we follow the camera- and world-space protocols of GVHMR~\citep{shen2024gvhmr} on EMDB~\citep{kaufmann2023emdb} and RICH~\citep{huang2022rich}.
Camera-space evaluation uses RICH and EMDB-1, while world-space evaluation uses RICH and EMDB-2.
For \movenszero, we retarget the predicted G1 motions back to SMPL and compare them with the native SMPL ground truth. We exclude all benchmark videos and near duplicates from training data.
We additionally reserve the \movensintent benchmark before temporal segmentation and model training.
The split contains four tracks---text, speech, music, and video---with 2,048 condition--motion pairs per track and 8,192 samples in total, stratified by motion type, scene, and condition family.
No original video, recovered motion identity, or near-duplicate clip is shared with the training set.
The benchmark uses Unitree G1 as the target motion representation across all four tracks.
We use the training split only to train learned evaluators and determine metric thresholds; the \movensintent benchmark is used exclusively for final evaluation.
Complete benchmark manifests, source-disjointness rules, and track-specific evaluation settings are provided in Appendix~\ref{app:evaluation_protocols}.

\paragraph{Baselines.}
For text-conditioned generation, both benchmarks include GENMO~\citep{li2025genmo}, HY-Motion~\citep{wen2025hymotion}, TextOp~\citep{xie2026textop}, and three Kimodo variants~\citep{rempe2026kimodo}. We additionally compare against OMG-L/XL~\citep{huang2026omg} on the \movensintent benchmark.
For video conditioning on RICH, EMDB, and the \movensintent benchmark, we evaluate GENMO~\citep{li2025genmo}, WHAM~\citep{shin2024wham}, TRAM~\citep{wang2024tram}, GVHMR~\citep{shen2024gvhmr}, PromptHMR~\citep{wang2025prompthmr}, and OnlineHMR~\citep{zhao2026onlinehmr}, which recover SMPL or SMPL-X motion directly from RGB video.
For audio-conditioned generation, we compare against GENMO~\citep{li2025genmo}, LODGE~\citep{li2024lodge}, and Bailando~\citep{siyao2022bailando} on both benchmarks. We additionally compare against OMG-L/XL~\citep{huang2026omg} on the \movensintent benchmark.

\paragraph{Metrics.}
For text, we report FID, Matching Score, R-Precision, and Diversity in the shared text-motion embedding space.
For video conditions, we follow GVHMR~\citep{shen2024gvhmr}: camera-space evaluation reports PA-MPJPE, MPJPE, PVE, and acceleration error, while world-space evaluation reports WA-MPJPE$_{100}$, W-MPJPE$_{100}$, root translation error, jitter, and foot sliding.
For audio, we report BeatAlign, FID-k, FID-g, and PFC.
For speech, we report R@3 with the evaluator used in Table~\ref{tab:speech_results}.
Shared tracking metrics include contact sliding (C-Slide), body jerk, MPJPE, global MPJPE, velocity and acceleration errors, fall rate, and joint-limit violation.
Tracking metrics for \movenszero are computed using HoloMotion.
A common policy-independent evaluator assigns each rollout a tracking score based on fall-free completion and normalized tracking fidelity.
We report Tracking Score as the average score across trajectories and compute all rollout-dependent metrics from the corresponding HoloMotion rollouts.
Exact metric definitions, units, valid-frame handling, and tracking-policy settings are detailed in Appendix~\ref{app:evaluation_protocols}.

\subsection{Omni-Modal Generation}
\label{sec:omnimodal_results}

We evaluate text, video, and audio conditions with modality-specific alignment protocols.
Text and audio results include the public OMG benchmark and the source-disjoint \movensintent benchmark, while video is evaluated on RICH, EMDB, and the \movensintent benchmark.
Every generated motion is decoded into the same Unitree G1 representation.

\paragraph{Text-conditioned generation.}

\begin{table*}[htbp]
\centering
\small
\setlength{\tabcolsep}{3pt}
\caption{Text-conditioned motion generation on the OMG and \movensintent benchmarks. R@K, Fall, and J-Limit are percentages. FID and C-Slide are scaled by $10^{-2}$ and $10^{-1}$, respectively.}
\label{tab:text_results}
\resizebox{\textwidth}{!}{
\begin{tabular}{lcccccccc|cccccc}
\toprule
\multirow{2}{*}{\textbf{Method}} &
\multicolumn{8}{c|}{\textbf{Motion Generation Quality}} &
\multicolumn{6}{c}{\textbf{Tracking Fidelity and Execution Failures}} \\
\cmidrule(lr){2-9}\cmidrule(lr){10-15}
& \textbf{FID$\downarrow$} & \textbf{Matching$\downarrow$} & \textbf{R@1$\uparrow$} & \textbf{R@2$\uparrow$} & \textbf{R@3$\uparrow$} & \textbf{Diversity$\uparrow$} &
\textbf{C-Slide$\downarrow$} & \textbf{Jerk$\downarrow$} & \textbf{MPJPE$\downarrow$} & \textbf{g-MPJPE$\downarrow$} & \textbf{$E_{\mathrm{vel}}\downarrow$} & \textbf{$E_{\mathrm{acc}}\downarrow$} & \textbf{Fall$\downarrow$} & \textbf{J-Limit$\downarrow$} \\
\midrule
\multicolumn{15}{l}{\textit{OMG benchmark}} \\
GENMO~\citep{li2025genmo}                    & 43.78 & 1.33 & 12.99 & 23.83 & 30.86 & 1.28 & \textbf{2.66} & 173.54 & 50.35 & 150.71 & 6.91 & 2.51 & \underline{1.46} & 29.10 \\
HY-Motion~\citep{wen2025hymotion}            & \underline{24.68} & \underline{1.30} & \underline{31.84} & \underline{48.73} & \underline{57.71} & \underline{1.29} & 12.20 & 112.56 & 69.05 & 287.71 & 11.94 & 3.24 & 4.98 & 55.57 \\
TextOp~\citep{xie2026textop}                 & 59.91 & \underline{1.30} & 11.52 & 19.43 & 25.00 & 1.15 & 3.80 & 219.67 & 38.00 & \underline{99.08} & 6.56 & 2.62 & 2.15 & 20.70 \\
Kimodo-SMPLX-RP~\citep{rempe2026kimodo}      & 38.44 & 1.31 & 26.27 & 39.36 & 46.48 & 1.25 & 4.52 & 189.39 & 43.51 & 198.96 & 7.80 & 2.64 & 2.64 & 23.63 \\
Kimodo-G1-RP~\citep{rempe2026kimodo}         & 46.36 & 1.32 & 25.59 & 36.13 & 44.63 & 1.21 & 3.92 & \underline{81.44} & \underline{37.25} & 125.04 & \underline{5.72} & \textbf{1.68} & \underline{1.46} & \underline{19.14} \\
Kimodo-G1-SEED~\citep{rempe2026kimodo}       & 50.22 & 1.33 & 22.46 & 33.69 & 41.11 & 1.20 & 4.30 & 99.47 & 37.26 & 161.63 & 6.38 & \underline{1.84} & 1.56 & 20.70 \\
\textbf{\movenszero (Ours)} & \textbf{10.13} & \textbf{1.24} & \textbf{62.60} & \textbf{77.83} & \textbf{85.06} & \textbf{1.37} & \underline{3.75} & \textbf{75.63} & \textbf{36.31} & \textbf{87.38} & \textbf{5.40} & 2.83 & \textbf{0.29} & \textbf{10.45} \\
\midrule
\multicolumn{15}{l}{\textit{\movensintent benchmark}} \\
GENMO~\citep{li2025genmo}                    & 75.18 & 1.34 & 13.23 & 23.93 & 31.93 & 1.19 & 0.89 & 245.45 & 68.39 & 133.71 & 3.00 & 2.01 & 0.34 & 11.38 \\
HY-Motion~\citep{wen2025hymotion}            & \underline{30.19} & \underline{1.26} & \underline{34.67} & \underline{51.76} & \underline{61.87} & \underline{1.33} & 1.55 & \underline{48.06} & 47.06 & 124.73 & 3.40 & 0.95 & 1.71 & 14.65 \\
TextOp~\citep{xie2026textop}                 & 105.96 & 1.49 & 6.25 & 10.60 & 15.33 & 0.86 & 1.38 & 85.30 & 37.22 & 129.08 & 2.92 & 1.05 & \textbf{0.00} & \underline{4.20} \\
Kimodo-SMPLX-RP~\citep{rempe2026kimodo}      & 44.28 & 1.28 & 25.54 & 37.35 & 47.31 & 1.25 & 1.65 & 240.06 & 37.80 & 108.51 & 4.00 & 2.16 & 0.83 & 9.13 \\
Kimodo-G1-RP~\citep{rempe2026kimodo}         & 39.56 & 1.29 & 24.17 & 36.57 & 45.90 & 1.22 & 1.80 & 214.20 & 36.19 & 91.16 & 3.33 & 1.87 & 0.44 & 9.33 \\
Kimodo-G1-SEED~\citep{rempe2026kimodo}       & 38.69 & 1.28 & 24.22 & 35.45 & 45.70 & 1.22 & 1.84 & 230.77 & \underline{35.26} & \underline{89.94} & 3.45 & 1.99 & 0.68 & 7.57 \\
OMG-L~\citep{huang2026omg}                   & 76.39 & 1.31 & 15.33 & 24.66 & 31.20 & 1.13 & 1.79 & 163.98 & 47.63 & 178.54 & 4.49 & 1.99 & 3.71 & 35.40 \\
OMG-XL~\citep{huang2026omg}                  & 124.90 & 1.39 & 5.57 & 9.86 & 14.36 & 0.84 & \textbf{0.40} & 70.95 & 47.18 & 96.21 & \textbf{1.96} & \textbf{0.77} & \underline{0.15} & \textbf{2.64} \\
\textbf{\movenszero (Ours)} & \textbf{17.65} & \textbf{1.25} & \textbf{34.83} & \textbf{51.88} & \textbf{62.57} & \textbf{1.34} & \underline{0.69} & \textbf{45.59} & \textbf{33.40} & \textbf{89.15} & \underline{2.82} & \underline{0.87} & \textbf{0.00} & 7.28 \\
\bottomrule
\end{tabular}
}
\end{table*}

As shown in Table~\ref{tab:text_results}, \movenszero achieves the best result on 12 of the 14 metrics of the OMG benchmark, demonstrating that unified intent understanding and motion generation can jointly preserve semantic alignment, motion quality, and physical executability. Relative to the strongest baseline for each metric, FID decreases from 24.68 to 10.13, a 59.0\% reduction, while R@3 increases from 57.71\% to 85.06\%, a gain of 27.35 percentage points. MPJPE and g-MPJPE decrease by 2.5\% and 11.8\%, the fall rate reaches 0.29\%, and J-Limit decreases from the best baseline value of 19.14\% to 10.45\%. These results show that the advantage of \movenszero extends beyond text--motion matching to tracking accuracy and execution stability after embodiment mapping.
On the source-disjoint \movensintent benchmark, \movenszero achieves the best or tied-best result on 10 of the 14 metrics. Its FID decreases from the strongest baseline value of 30.19 to 17.65, a 41.5\% reduction. R@1, R@2, and R@3 reach 34.83\%, 51.88\%, and 62.57\%, with R@3 exceeding the strongest baseline by 0.70 percentage points. C-Slide reaches 0.69, the second-best result. MPJPE decreases from the best baseline value of 35.26 to 33.40, a 5.3\% reduction, while g-MPJPE decreases from 89.94 to 89.15, a 0.9\% reduction. The fall rate is 0.00\%, tied for the best result. Overall, these results characterize cross-source generalization in text alignment, motion fidelity, and robot execution stability.

\paragraph{Video-conditioned generation.}

\begin{table*}[htbp]
\centering
\small
\setlength{\tabcolsep}{2.2pt}
\renewcommand{\arraystretch}{0.96}
\caption{Video-conditioned motion reconstruction on RICH and EMDB in SMPL space, and on \movensintent in G1 space.}
\label{tab:video_results}
\resizebox{\textwidth}{!}{
\begin{tabular}{lccccc|cccc}
\toprule
\multirow{2}{*}{\textbf{Method}} &
\multicolumn{5}{c|}{\textbf{World-space}} &
\multicolumn{4}{c}{\textbf{Camera-space}} \\
\cmidrule(lr){2-6}\cmidrule(lr){7-10}
& \textbf{WA-MPJPE$_{100}$}$\downarrow$ & \textbf{W-MPJPE$_{100}$}$\downarrow$ &
\textbf{RTE$\downarrow$} & \textbf{Jitter$\downarrow$} & \textbf{Foot-Sliding$\downarrow$} &
\textbf{PA-MPJPE$\downarrow$} & \textbf{MPJPE$\downarrow$} & \textbf{PVE$\downarrow$} & \textbf{Accel$\downarrow$} \\
\midrule
\multicolumn{10}{l}{\textit{RICH}} \\
GENMO~\citep{li2025genmo} & \underline{75.3} & \underline{118.6} & \underline{1.9} & 15.0 & 6.7 & 39.1 & 66.8 & 75.4 & \underline{4.1} \\
WHAM~\citep{shin2024wham} & 113.0 & 191.7 & 4.3 & 18.6 & 3.7 & 44.3 & 80.0 & 91.2 & 5.3 \\
TRAM~\citep{wang2024tram} & 103.5 & 168.0 & 2.5 & 18.5 & 12.5 & 44.1 & 70.5 & 78.7 & 5.4 \\
GVHMR~\citep{shen2024gvhmr} & 78.8 & 126.3 & 2.4 & 12.8 & 3.0 & 39.5 & 66.0 & 74.4 & \underline{4.1} \\
PromptHMR~\citep{wang2025prompthmr} & 78.2 & 125.9 & 2.3 & \underline{11.3} & \underline{2.7} & \underline{37.0} & \underline{57.4} & \textbf{65.8} & 4.2 \\
OnlineHMR~\citep{zhao2026onlinehmr} & 123.1 & 210.0 & 3.1 & 31.0 & 16.3 & 47.2 & 88.6 & 99.2 & 7.9 \\
\textbf{\movenszero (Ours)} & \textbf{72.5} & \textbf{115.3} & \textbf{1.7} & \textbf{10.5} & \textbf{2.5} & \textbf{36.4} & \textbf{53.3} & \underline{72.4} & \textbf{4.0} \\
\midrule
\multicolumn{10}{l}{\textit{EMDB}} \\
GENMO~\citep{li2025genmo} & 74.3 & \underline{202.1} & \underline{1.2} & 17.8 & 8.8 & 42.5 & 73.0 & 84.8 & 3.8 \\
WHAM~\citep{shin2024wham} & 135.6 & 354.8 & 6.0 & 22.5 & 4.4 & 50.4 & 79.7 & 94.4 & 5.3 \\
TRAM~\citep{wang2024tram} & 76.4 & 222.4 & 1.4 & 18.1 & 11.0 & 45.7 & 74.4 & 86.6 & 4.9 \\
GVHMR~\citep{shen2024gvhmr} & 111.0 & 276.5 & 2.0 & 16.7 & \underline{3.5} & 42.7 & 72.6 & 84.2 & \underline{3.6} \\
PromptHMR~\citep{wang2025prompthmr} & \underline{71.0} & 216.5 & 1.3 & \underline{16.3} & \underline{3.5} & \textbf{40.1} & \textbf{68.1} & \underline{79.2} & \underline{3.6} \\
OnlineHMR~\citep{zhao2026onlinehmr} & 93.5 & 310.4 & 2.2 & 33.5 & 16.4 & 46.0 & 74.0 & 86.1 & 9.0 \\
\textbf{\movenszero (Ours)} & \textbf{70.2} & \textbf{192.4} & \textbf{1.1} & \textbf{15.0} & \textbf{3.2} & \underline{42.0} & \underline{72.0} & \textbf{78.0} & \textbf{3.5} \\
\midrule
\multicolumn{10}{l}{\textit{\movensintent benchmark}} \\
GENMO~\citep{li2025genmo} & 55.1 & 76.2 & 5.5 & \underline{13.3} & 3.3 & 34.4 & 50.9 & -- & 3.5 \\
WHAM~\citep{shin2024wham} & 63.7 & 93.7 & 6.9 & 16.5 & 3.4 & 34.5 & 65.2 & -- & 4.1 \\
TRAM~\citep{wang2024tram} & 85.6 & 142.5 & 13.3 & 95.8 & 15.3 & 36.3 & 66.5 & -- & 4.6 \\
GVHMR~\citep{shen2024gvhmr} & \underline{42.5} & \underline{58.6} & \underline{3.9} & 16.5 & \underline{2.8} & \textbf{28.0} & \underline{41.8} & -- & \underline{2.9} \\
PromptHMR~\citep{wang2025prompthmr} & 66.0 & 98.1 & 8.2 & 17.2 & 4.5 & 36.1 & 63.2 & -- & 3.3 \\
OnlineHMR~\citep{zhao2026onlinehmr} & 144.7 & 357.0 & 62.1 & 135.9 & 36.3 & 39.3 & 68.2 & -- & 7.7 \\
\textbf{\movenszero (Ours)} & \textbf{40.2} & \textbf{49.1} & \textbf{2.0} & \textbf{11.4} & \textbf{2.4} & \underline{30.5} & \textbf{40.9} & -- & \textbf{2.5} \\
\bottomrule
\end{tabular}
}
\end{table*}

For the RICH and EMDB blocks of Table~\ref{tab:video_results}, the baselines recover SMPL motion directly from RGB video, whereas \movenszero predicts G1 motion and retargets it back to SMPL for evaluation against the same ground truth. \movenszero achieves the best result on 8 of the 9 metrics on RICH and 7 of the 9 metrics on EMDB. On RICH, WA-MPJPE$_{100}$ and W-MPJPE$_{100}$ are 72.5 and 115.3\,mm, improving over the strongest baselines by 3.7\% and 2.8\%, respectively. Jitter, PA-MPJPE, and Accel improve by 7.1\%, 1.6\%, and 2.4\%, respectively. RTE, Foot-Sliding, and MPJPE are 1.7\%, 2.5\,mm/frame, and 53.3\,mm, improving over the strongest baselines by 10.5\%, 7.4\%, and 7.1\%, respectively. PVE is second best at 72.4\,mm. On EMDB, WA-MPJPE$_{100}$ and W-MPJPE$_{100}$ are 70.2 and 192.4\,mm, improving by 1.1\% and 4.8\%. RTE, Jitter, and Foot-Sliding decrease by 8.3\%, 8.0\%, and 8.6\%, while PVE and Accel improve by 1.5\% and 2.8\%. PA-MPJPE and MPJPE are second best at 42.0 and 72.0\,mm. These results quantify how well video-aligned human motion is preserved after G1 generation and SMPL retargeting.
On the source-disjoint \movensintent benchmark, we retarget all six video baselines with the same GMR pipeline.
Under this protocol, \movenszero achieves the best result on 7 of the 8 defined metrics. WA-MPJPE$_{100}$ and W-MPJPE$_{100}$ decrease to 40.2 and 49.1\,mm, improving over the strongest baselines by 5.4\% and 16.2\%, respectively. RTE, Jitter, and Foot-Sliding decrease by 48.7\%, 14.3\%, and 14.3\%, while MPJPE and Accel decrease by 2.2\% and 13.8\%. PA-MPJPE is second best at 30.5\,mm, compared with GVHMR's 28.0\,mm.
SMPL PVE is not applicable to this block.

\paragraph{Music-conditioned generation.}

\begin{table*}[htbp]
\centering
\small
\setlength{\tabcolsep}{3pt}
\caption{Music-conditioned humanoid motion generation on the OMG and \movensintent benchmarks. Fall and J-Limit are percentages.}
\label{tab:music_results}
\resizebox{\textwidth}{!}{
\begin{tabular}{lcccc|cccccccc}
\toprule
\multirow{2}{*}{\textbf{Method}} &
\multicolumn{4}{c|}{\textbf{Music Alignment and Target Motion Quality}} &
\multicolumn{8}{c}{\textbf{Target Physical Quality and Tracker Fidelity}} \\
\cmidrule(lr){2-5}\cmidrule(lr){6-13}
& \textbf{BeatAlign$\uparrow$} & \textbf{FID-k$\downarrow$} & \textbf{FID-g$\downarrow$} & \textbf{PFC$\downarrow$} &
\textbf{C-Slide$\downarrow$} & \textbf{Jerk$\downarrow$} & \textbf{MPJPE$\downarrow$} & \textbf{g-MPJPE$\downarrow$} & \textbf{$E_{\mathrm{vel}}\downarrow$} & \textbf{$E_{\mathrm{acc}}\downarrow$} & \textbf{Fall$\downarrow$} & \textbf{J-Limit$\downarrow$} \\
\midrule
\multicolumn{13}{l}{\textit{OMG benchmark}} \\
GENMO~\citep{li2025genmo}                    & 0.54 & 207.39 & 17.70 & \textbf{0.08} & \underline{0.13} & 185.99 & \underline{39.79} & \underline{58.51} & \underline{3.54} & 1.80 & \textbf{0.00} & \textbf{2.05} \\
LODGE~\citep{li2024lodge}                    & 0.58 & \underline{80.46} & \underline{2.66} & 0.21 & 0.30 & 213.86 & 48.60 & 101.19 & 7.55 & 3.06 & 1.59 & 17.95 \\
Bailando~\citep{siyao2022bailando}           & \underline{0.59} & 328.15 & 16.85 & 0.19 & 0.17 & \underline{103.60} & 40.02 & 80.01 & 4.55 & \textbf{1.48} & \underline{0.23} & \underline{7.95} \\
\textbf{\movenszero (Ours)} & \textbf{0.60} & \textbf{53.44} & \textbf{2.31} & \underline{0.10} & \textbf{0.11} & \textbf{87.25} & \textbf{37.79} & \textbf{50.39} & \textbf{3.40} & \underline{1.75} & \textbf{0.00} & 15.32 \\
\midrule
\multicolumn{13}{l}{\textit{\movensintent benchmark}} \\
GENMO~\citep{li2025genmo}                    & 0.58 & 91.02 & 5.32 & \underline{0.09} & \underline{0.20} & \underline{187.90} & 51.46 & 107.24 & \underline{4.39} & \textbf{1.97} & \underline{0.10} & \textbf{5.27} \\
LODGE~\citep{li2024lodge}                    & 0.57 & \underline{59.43} & \textbf{2.14} & 0.20 & 0.28 & 206.20 & 64.02 & 180.19 & 7.53 & 2.89 & 3.52 & 28.17 \\
Bailando~\citep{siyao2022bailando}           & \underline{0.60} & 94.40 & 10.30 & 0.33 & 0.27 & 219.76 & 90.68 & 274.77 & 7.08 & 2.63 & 18.31 & 50.05 \\
OMG-L~\citep{huang2026omg}                   & \underline{0.60} & 75.69 & 6.71 & 0.27 & 0.38 & 262.15 & 44.25 & 115.39 & 6.00 & 2.75 & \textbf{0.00} & 24.37 \\
OMG-XL~\citep{huang2026omg}                  & 0.57 & 68.70 & 5.83 & 0.32 & 0.34 & 266.26 & \underline{40.79} & \underline{98.61} & 5.25 & 2.56 & 0.44 & 33.01 \\
\textbf{\movenszero (Ours)} & \textbf{0.61} & \textbf{58.43} & \underline{4.58} & \textbf{0.07} & \textbf{0.18} & \textbf{142.31} & \textbf{35.95} & \textbf{90.28} & \textbf{4.01} & \underline{1.99} & \textbf{0.00} & \underline{14.72} \\
\bottomrule
\end{tabular}
}
\end{table*}

As shown in Table~\ref{tab:music_results}, \movenszero achieves the best or tied-best result on 9 of the 12 metrics of the OMG benchmark. Its BeatAlign reaches 0.60, compared with the strongest baseline value of 0.59. FID-k and FID-g decrease to 53.44 and 2.31, improving over their corresponding strongest baselines by 33.6\% and 13.2\%. For physical quality and tracking, Jerk, MPJPE, and g-MPJPE are 87.25, 37.79, and 50.39, corresponding to improvements of 15.8\%, 5.0\%, and 13.9\%. The fall rate is 0.00\%, tied for the best result, and J-Limit is 15.32\%, compared with the best baseline value of 2.05\%. These results show gains in music--motion alignment and tracking fidelity, with remaining limitations in joint-limit compliance.
On the source-disjoint \movensintent benchmark, \movenszero achieves the best or tied-best result on 9 of the 12 metrics. Its BeatAlign reaches 0.61, while FID-k decreases from the best baseline value of 59.43 to 58.43, a 1.7\% reduction. FID-g is 4.58, the second-best result. MPJPE and g-MPJPE decrease to 35.95 and 90.28, corresponding to reductions of 11.9\% and 8.4\% over the strongest baselines for these metrics. The fall rate is 0.00\%, tied for the best result, and J-Limit is 14.72\%, the second-best result. Overall, these results characterize temporal alignment, motion fidelity, and humanoid executability on unseen audio sources.

\paragraph{Speech-conditioned generation.}

\begin{table*}[htbp]
\centering
\small
\setlength{\tabcolsep}{3pt}
\caption{Speech-conditioned humanoid motion generation on the \movensintent benchmark. R@K, Fall, and J-Limit are percentages. FID and C-Slide are scaled by $10^{-2}$ and $10^{-1}$, respectively.}
\label{tab:speech_results}
\resizebox{\textwidth}{!}{
\begin{tabular}{lcccccccc|cccccc}
\toprule
\multirow{2}{*}{\textbf{Method}} &
\multicolumn{8}{c|}{\textbf{Motion Generation Quality}} &
\multicolumn{6}{c}{\textbf{Tracking Fidelity and Execution Failures}} \\
\cmidrule(lr){2-9}\cmidrule(lr){10-15}
& \textbf{FID$\downarrow$} & \textbf{Matching$\downarrow$} & \textbf{R@1$\uparrow$} & \textbf{R@2$\uparrow$} & \textbf{R@3$\uparrow$} & \textbf{Diversity$\uparrow$} &
\textbf{C-Slide$\downarrow$} & \textbf{Jerk$\downarrow$} & \textbf{MPJPE$\downarrow$} & \textbf{g-MPJPE$\downarrow$} & \textbf{$E_{\mathrm{vel}}\downarrow$} & \textbf{$E_{\mathrm{acc}}\downarrow$} & \textbf{Fall$\downarrow$} & \textbf{J-Limit$\downarrow$} \\
\midrule
GENMO~\citep{li2025genmo}                    & 77.30 & 1.35 & 14.70 & 23.83 & 31.45 & 1.21 & 0.86 & 237.09 & 67.68 & 134.19 & 2.94 & 1.95 & 0.10 & 11.18 \\
HY-Motion~\citep{wen2025hymotion}            & \underline{30.66} & \underline{1.25} & \underline{36.67} & \underline{52.49} & \underline{62.40} & \textbf{1.33} & 1.44 & \underline{49.35} & 45.63 & 117.10 & 3.30 & 0.94 & 1.03 & 13.62 \\
TextOp~\citep{xie2026textop}                 & 105.46 & 1.48 & 5.91 & 10.35 & 14.89 & 0.90 & 1.43 & 89.54 & 38.05 & 129.41 & 3.04 & 1.10 & 0.20 & 5.42 \\
Kimodo-SMPLX-RP~\citep{rempe2026kimodo}      & 45.18 & 1.29 & 25.00 & 37.26 & 47.17 & 1.25 & 1.60 & 242.13 & 36.89 & 102.85 & 3.93 & 2.17 & 0.24 & 9.28 \\
Kimodo-G1-RP~\citep{rempe2026kimodo}         & 38.84 & 1.29 & 25.00 & 37.21 & 46.24 & 1.23 & 1.76 & 215.91 & 35.50 & 89.83 & 3.28 & 1.87 & 0.39 & 8.25 \\
Kimodo-G1-SEED~\citep{rempe2026kimodo}       & 37.62 & 1.28 & 23.88 & 36.96 & 47.02 & 1.22 & 1.86 & 231.75 & \underline{34.83} & \underline{89.82} & 3.44 & 2.00 & 0.39 & 6.93 \\
OMG-L~\citep{huang2026omg}                   & 73.77 & 1.31 & 11.77 & 21.04 & 28.27 & 1.15 & 1.82 & 167.23 & 53.11 & 198.08 & 4.69 & 2.04 & 5.18 & 37.35 \\
OMG-XL~\citep{huang2026omg}                  & 126.68 & 1.39 & 4.83 & 9.91 & 14.26 & 0.83 & \textbf{0.40} & 69.94 & 45.79 & 94.98 & \underline{1.99} & \textbf{0.77} & \underline{0.05} & \textbf{2.00} \\
\textbf{\movenszero (Ours)} & \textbf{23.15} & \textbf{1.23} & \textbf{43.27} & \textbf{59.76} & \textbf{74.30} & \underline{1.32} & \underline{0.75} & \textbf{43.29} & \textbf{32.63} & \textbf{76.38} & \textbf{1.87} & \underline{0.89} & \textbf{0.00} & \underline{4.93} \\
\bottomrule
\end{tabular}
}
\end{table*}

As shown in Table~\ref{tab:speech_results}, \movenszero achieves the best result on 10 of the 14 metrics of the source-disjoint \movensintent benchmark. Its FID decreases from the strongest baseline value of 30.66 to 23.15, a 24.5\% reduction, while R@1, R@2, and R@3 reach 43.27\%, 59.76\%, and 74.30\%, respectively. R@3 exceeds the strongest baseline value of 62.40\% by 11.90 percentage points. MPJPE and g-MPJPE decrease from 34.83 and 89.82 to 32.63 and 76.38, corresponding to reductions of 6.3\% and 15.0\%. The fall rate is 0.00\%. C-Slide is 0.75 and J-Limit is 4.93\%, compared with the best baseline values of 0.40 and 2.00\%, respectively. These results show improved spoken-instruction alignment and tracking fidelity, with remaining limitations in contact consistency and joint-limit compliance.

\subsection{Qualitative Comparisons}
\label{sec:qualitative_comparisons}

\begin{figure*}[p]
\centering
\includegraphics[width=\linewidth,height=0.88\textheight,keepaspectratio]{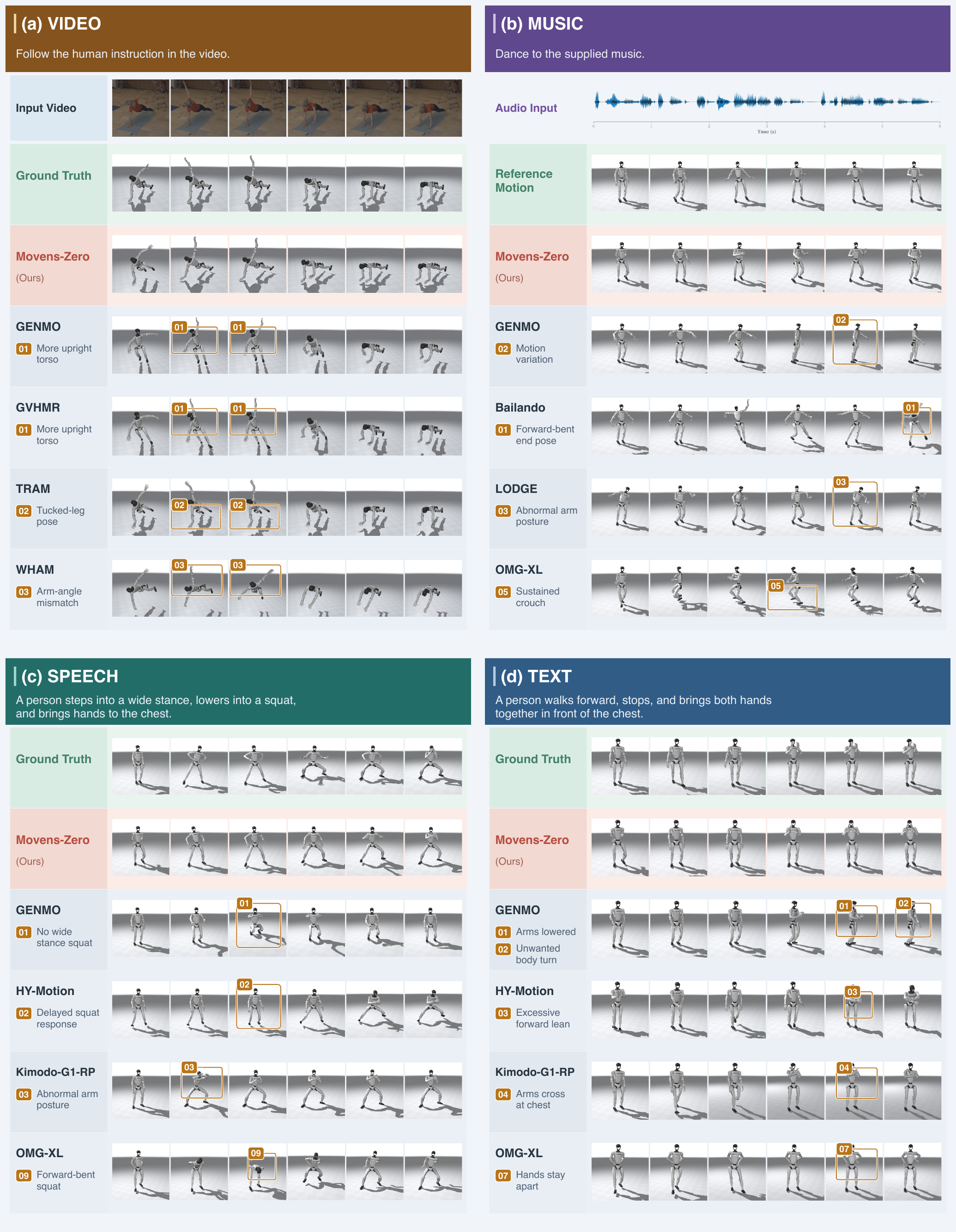}
\caption{Qualitative overview of (a) video, (b) music, (c) speech, and (d) text conditioning. Each panel shows six temporal samples of the ground-truth motion (reference motion for music), \movenszero, and four baselines. Full comparisons appear in Appendix~\ref{app:offline_qualitative}. Conditioning instructions appear in the headers; amber callouts indicate selected visible pose differences.}
\label{fig:qualitative_overview}
\end{figure*}

Figure~\ref{fig:qualitative_overview} presents qualitative comparisons under video, music, speech, and text conditioning. In video panel~(a), \movenszero retains a low torso posture during the arm raise before transitioning into forward support, whereas several baselines appear more upright in the initial phase. In music panel~(b), the generated sequence shows changes in stance and arm position. The reference motion represents one plausible interpretation of the audio. In speech panel~(c), \movenszero widens its stance, lowers into a squat, and brings the hands toward the chest, while some baselines omit or delay the squat. In text panel~(d), \movenszero follows the instructed sequence of walking, stopping, and bringing the hands together, while several baselines finish with separated or crossed arms.
Complete qualitative comparisons for all fourteen cases are provided in Appendix~\ref{app:offline_qualitative}.

\subsection{In-Depth Analysis}
\label{sec:in_depth_analysis}

\paragraph{Data scaling.}

We test the central hypothesis that increasing the amount of diverse, execution-verified data yields predictable improvements in multimodal generation and physical execution.
We construct nested subsets containing $1\%$, $3\%$, $10\%$, $30\%$, and $100\%$ of \movensdata.
Subsampling is performed at the original source-video level before temporal slicing, preventing overlapping windows from inflating the effective data size.
All subsets preserve the scene, motion-type, and modality distributions of the full corpus.
All data-scaling runs use the XL action expert (910.5M parameters).
The splits and controlled variables for the data- and model-scaling experiments are detailed in Appendix~\ref{app:evaluation_protocols}.

\begin{table*}[htbp]
\centering
\small
\setlength{\tabcolsep}{4.5pt}
\caption{Data scaling with the XL action expert (910.5M parameters).}
\label{tab:data_scaling}
\resizebox{\textwidth}{!}{
\begin{tabular}{lcccccc}
\toprule
\textbf{\movensdata fraction} & \textbf{Unique hours} &
\textbf{Text R@3$\uparrow$} & \textbf{Video MPJPE$\downarrow$} &
\textbf{Audio BeatAlign$\uparrow$} &
\textbf{Speech R@3$\uparrow$} & \textbf{Tracking Score$\uparrow$} \\
\midrule
$1\%$   & 122 & 12.72 & 132.82 & 0.23 & 13.98 & 0.21 \\
$3\%$   & 367 & 15.81 & 129.31 & 0.29 & 17.88 & 0.28 \\
$10\%$  & 1,223 & 24.92 & 110.55 & 0.38 & 21.76 & 0.36 \\
$30\%$  & 3,669 & 41.80 & 87.90 & 0.45 & 54.34 & 0.50 \\
$100\%$ & 12,231 & 62.57 & 40.90 & 0.61 & 74.30 & 0.72 \\
\bottomrule
\end{tabular}
}
\end{table*}

As shown in Table~\ref{tab:data_scaling}, all five metrics improve monotonically as the amount of unique training data increases. From $1\%$ to $100\%$, Text R@3 increases from 12.72\% to 62.57\%, and Speech R@3 from 13.98\% to 74.30\%, corresponding to gains of 49.85 and 60.32 percentage points, respectively. Video MPJPE decreases from 132.82 to 40.90\,mm, a 69.2\% reduction, while Audio BeatAlign increases from 0.23 to 0.61 and Tracking Score from 0.21 to 0.72. From $30\%$ to $100\%$, Text and Speech R@3 improve by a further 20.77 and 19.96 percentage points, respectively, while Video MPJPE decreases by 53.5\%. These results support the benefits of broader unique data coverage for multimodal alignment and robot trackability.

\paragraph{Model scaling.}
We train four action-expert variants---S (64.2M), M (187.9M), L (500.1M), and XL (910.5M)---on the full \movensdata training set while keeping the data pipeline unchanged.
Table~\ref{tab:model_scaling} reports validation, modality-specific generation, and physical-execution metrics across model scales, separating gains from unique data coverage and model capacity.

\begin{table*}[htbp]
\centering
\small
\setlength{\tabcolsep}{4.5pt}
\caption{Model scaling of \movenszero on the full \movensdata training set.}
\label{tab:model_scaling}
\resizebox{\textwidth}{!}{
\begin{tabular}{lccccccc}
\toprule
\textbf{Scale} & \textbf{Action expert} & \textbf{Val. FM Loss$\downarrow$} &
\textbf{Text R@3$\uparrow$} & \textbf{Video MPJPE$\downarrow$} &
\textbf{Audio BeatAlign$\uparrow$} &
\textbf{Speech R@3$\uparrow$} &
\textbf{Tracking Score$\uparrow$} \\
\midrule
S  & 64M  & 0.487 & 10.30 & 108.50 & 0.20 & 25.37 & 0.18 \\
M  & 188M & 0.392 & 22.96 & 92.80 & 0.26 & 41.25 & 0.24 \\
L  & 500M & 0.186 & 47.80 & 73.84 & 0.53 & 63.98 & 0.40 \\
XL & 911M  & 0.159 & 62.57 & 40.90 & 0.61 & 74.30 & 0.72 \\
\bottomrule
\end{tabular}
}
\end{table*}

As shown in Table~\ref{tab:model_scaling}, all six metrics improve from S to XL. Validation FM loss decreases from 0.487 to 0.159, a 67.4\% reduction, while Video MPJPE decreases from 108.50 to 40.90\,mm, a 62.3\% reduction. Text R@3 increases from 10.30\% to 62.57\%, and Speech R@3 from 25.37\% to 74.30\%, corresponding to gains of 52.27 and 48.93 percentage points, respectively. Audio BeatAlign increases from 0.20 to 0.61, and Tracking Score increases from 0.18 to 0.72. The L-to-XL step still improves Text and Speech R@3 by 14.77 and 10.32 percentage points, respectively, while further reducing Video MPJPE by 44.6\%. These results indicate that increasing action-expert capacity benefits multimodal alignment and robot trackability.

\begin{figure*}[t]
\centering
\includegraphics[width=\linewidth]{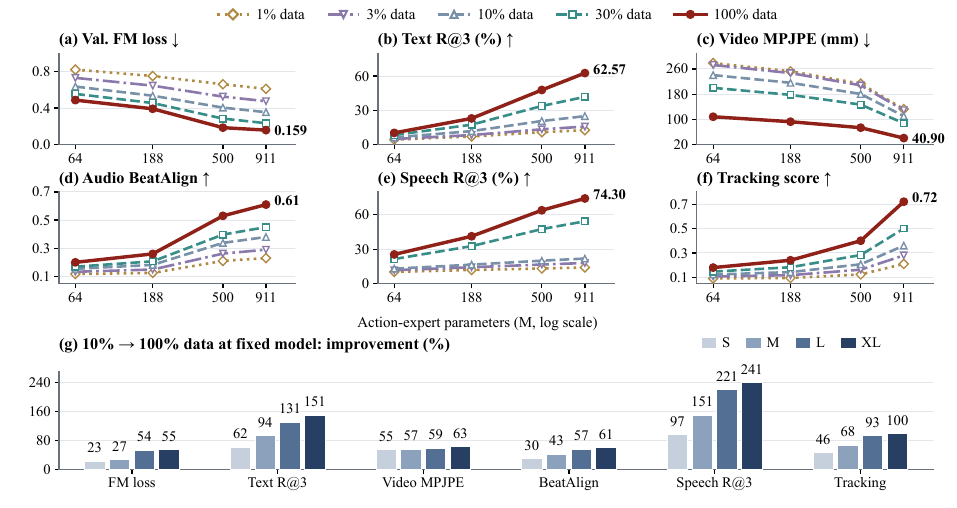}
\caption{Joint data--model scaling. (a--f) Six metrics versus action-expert size for five training-data fractions. (g) Relative improvements from 10\% to 100\% training data at each fixed model size, normalized by the corresponding 10\% value. Reductions in FM loss and MPJPE, and increases in the other scores, are positive.}
\label{fig:joint_scaling}
\end{figure*}

\paragraph{Joint data--model scaling.}
As shown in Figure~\ref{fig:joint_scaling}, we compare data-scaling gains across action-expert capacities using the measured XL results in Table~\ref{tab:data_scaling}, the full-data results in Table~\ref{tab:model_scaling}, and values for the remaining smaller-data results. Expanding the training set from 10\% to 100\% increases Text R@3 from 6.34\% to 10.30\% for S (62.5\% relative improvement). For XL, Text R@3 increases from 24.92\% to 62.57\% (151.1\%). The corresponding reductions in validation FM loss are 23.3\% and 55.2\% for S and XL, respectively. For XL, the same data expansion reduces Video MPJPE from 110.55 to 40.90\,mm (63.0\%) and increases Audio BeatAlign from 0.38 to 0.61 (60.5\%) and Tracking Score from 0.36 to 0.72 (100.0\%). Speech R@3 for XL increases from 21.76\% to 74.30\%, a gain of 52.54 percentage points (241.5\% relative improvement). At a fixed data fraction, scaling from S to XL increases Text R@3 by 18.58 percentage points with 10\% data, compared with 52.27 percentage points with the full dataset. Across all six metrics in Figure~\ref{fig:joint_scaling}(g), the relative improvements from 10\% to 100\% training data increase with action-expert capacity. For example, Speech R@3 improves by 96.5\% for S versus 241.5\% for XL, while Tracking Score improves by 46.3\% and 100.0\%, respectively. These comparisons suggest that larger action experts can benefit more from additional training data, highlighting the complementary gains from scaling data and model capacity.

\paragraph{Data curation and verification.}
To separate the effects of quality and quantity, we construct size-matched training subsets from successive stages of the \movensdata pipeline. All variants contain the same number of motion sequences and use the same optimization-token budget. The upper block of Table~\ref{tab:ablations} follows a cumulative setup. Visual and human-motion filtering first removes clips with missing subjects, temporal anomalies, near-static content, or unreliable motion recovery. Closed-loop physical refinement then repairs local artifacts such as contact errors, jitter, self-intersection, and foot sliding, while converting human motion into reference trajectories that satisfy robot embodiment and joint constraints. Tracking verification subsequently rejects candidate motions that cannot be executed stably under a common closed-loop simulation criterion. Finally, distribution balancing prevents frequent motions and scenes from dominating training, allowing a fixed sample budget to cover more long-tail behaviors. Because each row includes all preceding stages, this experiment measures the marginal change produced by adding each stage rather than independent single-factor effects. Specific gating rules are listed in Appendix Table~\ref{tab:active_data_gates}. Human-motion inspection and repair, as well as motion retargeting and execution verification, are detailed in Appendices~\ref{app:motion_quality} and~\ref{app:retargeting_verification}, respectively.

\paragraph{Modality-aligned training.}
We further use leave-one-modality-out ablations that remove text--motion, video--motion, or audio--motion alignment training. Audio alignment includes both music-conditioned and speech-conditioned generation. Each variant masks only the corresponding condition, while the original motion samples continue to participate in training through the remaining tasks. The action-expert size, optimization steps, and token budget are held fixed. The lower block of Table~\ref{tab:ablations} reports generation metrics for the modalities retained during alignment training, together with Tracking Score. Condition-specific metrics for the excluded modality are not reported in this ablation. The complete seven-task mixture and input forms are provided in Appendix Table~\ref{tab:training_task_mix}, while aligned motion chunks and condition processing are detailed in Appendix~\ref{app:training_details}. These controlled comparisons examine how joint alignment affects generation under the remaining conditioning modalities, although a single leave-one-out experiment is insufficient to attribute cross-metric changes to one independent mechanism.

\begin{table*}[htbp]
\centering
\small
\setlength{\tabcolsep}{5pt}
\caption{Ablations of data curation and modality-aligned training on the \movensintent benchmark.}
\label{tab:ablations}
\resizebox{\textwidth}{!}{
\begin{tabular}{lcccc|c}
\toprule
\textbf{Variant} &
\textbf{Text R@3$\uparrow$} & \textbf{Video MPJPE$\downarrow$} &
\textbf{Audio BeatAlign$\uparrow$} &
\textbf{Speech R@3$\uparrow$} &
\textbf{Tracking Score$\uparrow$} \\
\midrule
\multicolumn{6}{l}{\textit{Data curation}} \\
Raw recovered motion                         & 48.72 & 126.82 & 0.48 & 35.31 & 0.41 \\
\quad + visual and motion filtering          & 54.36 & 84.57 & 0.52 & 49.43 & 0.48 \\
\quad + closed-loop physical refinement      & 57.84 & 64.91 & 0.55 & 64.35 & 0.58 \\
\quad + tracking verification                & 61.92 & 57.62 & 0.58 & 70.23 & 0.67 \\
\quad + distribution balancing \textbf{(full)}        & 62.57 & 40.90 & 0.61 & 74.30 & 0.72 \\
\midrule
\multicolumn{6}{l}{\textit{Model and training}} \\
\movenszero w/o text--motion alignment          & -- & 56.84 & 0.59 & 73.16 & 0.69 \\
\movenszero w/o video--motion alignment         & 61.74 & -- & 0.60 & 72.32 & 0.66 \\
\movenszero w/o audio--motion alignment         & 60.02 & 66.07 & -- & -- & 0.69 \\
\textbf{\movenszero (full)}                    & 62.57 & 40.90 & 0.61 & 74.30 & 0.72 \\
\bottomrule
\end{tabular}
}
\end{table*}

The upper block of Table~\ref{tab:ablations} evaluates the cumulative data pipeline with matched sample counts and training budgets. Relative to raw recovered motion, the full pipeline increases Text R@3 from 48.72\% to 62.57\%, a gain of 13.85 percentage points, and reduces Video MPJPE from 126.82 to 40.90\,mm, a 67.7\% reduction. Audio BeatAlign rises from 0.48 to 0.61, while Speech R@3 increases from 35.31\% to 74.30\%. Tracking Score increases from 0.41 to 0.72. Visual and human-motion filtering adds 5.64 percentage points in Text R@3 and reduces MPJPE by 33.3\%. Closed-loop physical refinement further reduces MPJPE from 84.57 to 64.91\,mm (23.2\%) and improves Tracking Score from 0.48 to 0.58. Tracking verification raises the score to 0.67, and distribution balancing further increases it to 0.72 while adding 0.65 percentage points in Text R@3. These cumulative gains show how data curation improves conditional alignment and robot trackability under fixed sample and training budgets.

The lower block of Table~\ref{tab:ablations} evaluates the effects of modality-aligned training on the remaining conditioning tasks. Removing video--motion alignment lowers Text R@3 from 62.57\% to 61.74\%, a drop of 0.83 percentage points, and Speech R@3 from 74.30\% to 72.32\%, a drop of 1.98 percentage points. Removing audio--motion alignment lowers Text R@3 to 60.02\%, a drop of 2.55 percentage points, and increases Video MPJPE from 40.90 to 66.07\,mm (61.5\%). Without text--motion alignment, Video MPJPE increases to 56.84\,mm (39.0\%), Speech R@3 decreases to 73.16\%, and Audio BeatAlign decreases from 0.61 to 0.59 (3.3\%). These results indicate that joint alignment can benefit generation under other conditioning modalities.

\section{Conclusion}

In this work, we introduce \textbf{\movensdata}, a large-scale omni-modal humanoid motion dataset constructed from Internet videos.
\movensdata employs an automatic data pipeline to convert diverse human videos into multimodally annotated, physically refined, and executable humanoid motions.
We further construct \textbf{\movensintent}, a source-disjoint four-track benchmark containing 8,192 expert-verified condition--motion pairs spanning text, speech, music, and video for evaluating condition alignment, motion fidelity, and robot executability in a unified Unitree G1 representation.
Building on \movensdata, we introduce \textbf{\movenszero}, a scalable understanding--generation--execution architecture that translates multimodal human intent directly into physically executable humanoid behaviors in real time through a unified interface spanning text, video, speech, and music conditions.
Experiments on public benchmarks and \movensintent demonstrate that \movenszero jointly maintains intent alignment, motion fidelity, and stable robot execution under text, video, speech, and music conditions. Data- and model-scaling experiments further show consistent performance improvements as non-duplicated training-data coverage and action-expert capacity increase.
Further analyses highlight the importance of data diversity and physical executability for effective scaling.
In future work, we plan to extend the Movens data and model family to richer human--object interactions, additional humanoid embodiments, and broader real-world applications.

\newpage
\bibliography{references}
\bibliographystyle{movens_references}

\clearpage
\appendix
\section*{Appendix}
\label{sec:appendix}

\subsection*{Contents}
\begingroup
\setlength{\parskip}{0pt}
\begin{enumerate}
    \setlength{\itemsep}{4pt}
    \setlength{\parsep}{0pt}
    \item \hyperref[app:dataset_details]{Appendix~\ref*{app:dataset_details}: Movens-5M Dataset Details}
    \begin{itemize}
        \setlength{\itemsep}{0pt}
        \setlength{\topsep}{1pt}
        \setlength{\parsep}{0pt}
        \item \hyperref[app:dataset_examples]{Appendix~\ref*{app:dataset_examples}: Representative Dataset Examples}
        \item \hyperref[app:motion_quality]{Appendix~\ref*{app:motion_quality}: Human Motion Quality Assessment and Repair}
        \item \hyperref[app:retargeting_verification]{Appendix~\ref*{app:retargeting_verification}: Physics-Aware Retargeting and Execution Verification}
    \end{itemize}
    \item \hyperref[app:method_details]{Appendix~\ref*{app:method_details}: Additional \movenszero Method Details}
    \begin{itemize}
        \setlength{\itemsep}{0pt}
        \setlength{\topsep}{1pt}
        \setlength{\parsep}{0pt}
        \item \hyperref[app:motion_representation]{Appendix~\ref*{app:motion_representation}: Robot-Centric Motion Representation}
        \item \hyperref[app:condition_encoding]{Appendix~\ref*{app:condition_encoding}: Omni-Modal Condition Encoding}
        \item \hyperref[app:scale_configurations]{Appendix~\ref*{app:scale_configurations}: Action-Expert Scale Configurations}
        \item \hyperref[app:task_mixed_generation]{Appendix~\ref*{app:task_mixed_generation}: Task-Mixed Training and Online Generation}
    \end{itemize}
    \item \hyperref[app:training_details]{Appendix~\ref*{app:training_details}: Training Details}
    \begin{itemize}
        \setlength{\itemsep}{0pt}
        \setlength{\topsep}{1pt}
        \setlength{\parsep}{0pt}
        \item \hyperref[app:cached_conditions]{Appendix~\ref*{app:cached_conditions}: Cached Omni-Modal Conditions}
        \item \hyperref[app:motion_chunk_construction]{Appendix~\ref*{app:motion_chunk_construction}: Aligned Motion-Chunk Construction}
        \item \hyperref[app:training_loss]{Appendix~\ref*{app:training_loss}: Loss and Optimization}
        \item \hyperref[app:validation_reproducibility]{Appendix~\ref*{app:validation_reproducibility}: Validation and Reproducibility}
    \end{itemize}
    \item \hyperref[app:evaluation_protocols]{Appendix~\ref*{app:evaluation_protocols}: Evaluation Protocols}
    \begin{itemize}
        \setlength{\itemsep}{0pt}
        \setlength{\topsep}{1pt}
        \setlength{\parsep}{0pt}
        \item \hyperref[app:benchmarks_baselines]{Appendix~\ref*{app:benchmarks_baselines}: Benchmarks and Baselines}
        \item \hyperref[app:intent_examples]{Appendix~\ref*{app:intent_examples}: Movens-Intent Benchmark Examples}
        \item \hyperref[app:evaluation_metrics]{Appendix~\ref*{app:evaluation_metrics}: Semantic, Motion, and Execution Metrics}
        \item \hyperref[app:scaling_protocol]{Appendix~\ref*{app:scaling_protocol}: Scaling Protocol}
    \end{itemize}
    \item \hyperref[app:qualitative_results]{Appendix~\ref*{app:qualitative_results}: Additional Qualitative Results and Failure Cases}
    \begin{itemize}
        \setlength{\itemsep}{0pt}
        \setlength{\topsep}{1pt}
        \setlength{\parsep}{0pt}
        \item \hyperref[app:offline_qualitative]{Appendix~\ref*{app:offline_qualitative}: Additional Offline Comparisons}
        \item \hyperref[app:real_robot_cases]{Appendix~\ref*{app:real_robot_cases}: Real-Robot Capability Cases}
        \item \hyperref[app:failure_cases]{Appendix~\ref*{app:failure_cases}: Failure Cases}
    \end{itemize}
    \item \hyperref[app:prompting_details]{Appendix~\ref*{app:prompting_details}: Prompting Details}
    \begin{itemize}
        \setlength{\itemsep}{0pt}
        \setlength{\topsep}{1pt}
        \setlength{\parsep}{0pt}
        \item \hyperref[app:data_filtering_prompts]{Appendix~\ref*{app:data_filtering_prompts}: Human Motion Video Filtering Prompts}
        \item \hyperref[app:audio_visual_consistency_prompts]{Appendix~\ref*{app:audio_visual_consistency_prompts}: Audio-Visual Semantic Consistency Prompts}
        \item \hyperref[app:motion_classification_prompts]{Appendix~\ref*{app:motion_classification_prompts}: Motion Classification Prompts}
        \item \hyperref[app:scene_classification_prompts]{Appendix~\ref*{app:scene_classification_prompts}: Scene Classification Prompts}
        \item \hyperref[app:human_attribute_prompts]{Appendix~\ref*{app:human_attribute_prompts}: Human Attribute Annotation Prompts}
        \item \hyperref[app:object_interaction_prompts]{Appendix~\ref*{app:object_interaction_prompts}: Object Interaction Annotation Prompts}
        \item \hyperref[app:annotation_prompts]{Appendix~\ref*{app:annotation_prompts}: Video Caption Annotation Prompts}
        \item \hyperref[app:motion_caption_prompts]{Appendix~\ref*{app:motion_caption_prompts}: Motion Caption Annotation Prompts}
        \item \hyperref[app:motion_generation_prompts]{Appendix~\ref*{app:motion_generation_prompts}: Motion Generation Task Prompts}
    \end{itemize}
\end{enumerate}
\endgroup
\clearpage

\section{Movens-5M Dataset Details}
\label{app:dataset_details}

\begin{table}[h!]
\centering
\small
\setlength{\tabcolsep}{5pt}
\caption{Hard gates used by the audited web-video assembly. Audio--visual consistency and generic perceptual scores are retained as diagnostics rather than restrictive final gates.}
\label{tab:active_data_gates}
\begin{tabular}{>{\raggedright\arraybackslash}p{0.18\textwidth}>{\raggedright\arraybackslash}p{0.24\textwidth}>{\raggedright\arraybackslash}p{0.50\textwidth}}
\toprule
\textbf{Gate} & \textbf{Signal} & \textbf{Retention rule} \\
\midrule
Human relevance & Qwen3-VL yes/no likelihood & $s_{\mathrm{human}}\geq0.90$ \\
Temporal rate & RGB-window class posterior & Slow or normal playback/encoding rate \\
Articulated motion & Body-normalized limb amplitude & $A_{\mathrm{limb}}\geq0.08$ \\
Temporal integrity & Human/background flow statistics & No robust discontinuity test fires \\
Executability & Selected rollout tracking score & $q\geq0.20$ \\
Record completeness & Metadata and derived artifacts & All mandatory annotations, motions, rollout states, logs, and paths are present and readable \\
\bottomrule
\end{tabular}
\end{table}

\subsection{Representative Dataset Examples}
\label{app:dataset_examples}

Figure~\ref{fig:appendix_dataset_diverse_examples} presents fourteen \movensdata records spanning daily activities, fitness, sports, dance, locomotion, performance, and gestures, with two motion subtypes per category.
Each case follows the source video through recovered human motion, a retargeted humanoid reference, and a simulation tracking rollout.
Six shared clip-relative timestamps align corresponding poses across the four streams, using the nearest available frame in each.
The original video captions provide scene and action context.
Figure~\ref{fig:appendix_dataset_structured_example} further connects the gardening case to its metadata and temporally localized motion captions, including body-part descriptions.

\begingroup
\setlength{\intextsep}{6pt}
\begin{figure}[!ht]
    \centering
    \captionsetup{skip=4pt}
    \includegraphics[width=0.91\textwidth]{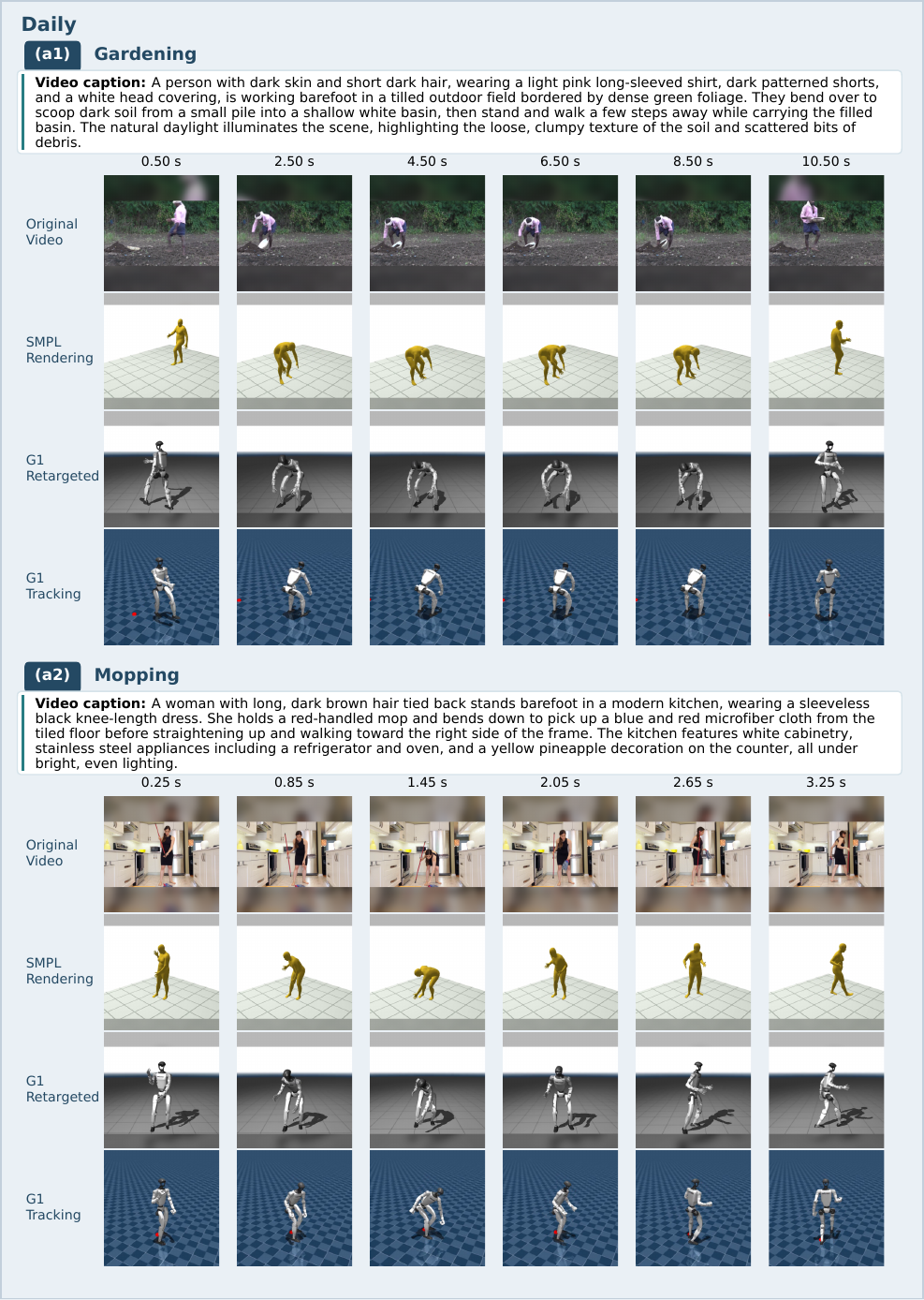}
    \caption{Representative \movensdata examples: Daily, with (a1) gardening and (a2) mopping.}
    \label{fig:appendix_dataset_category_daily}
    \label{fig:appendix_dataset_diverse_examples}
\end{figure}
\endgroup
\clearpage

\begin{figure}[!ht]
    \ContinuedFloat
    \centering
    \includegraphics[width=0.91\textwidth]{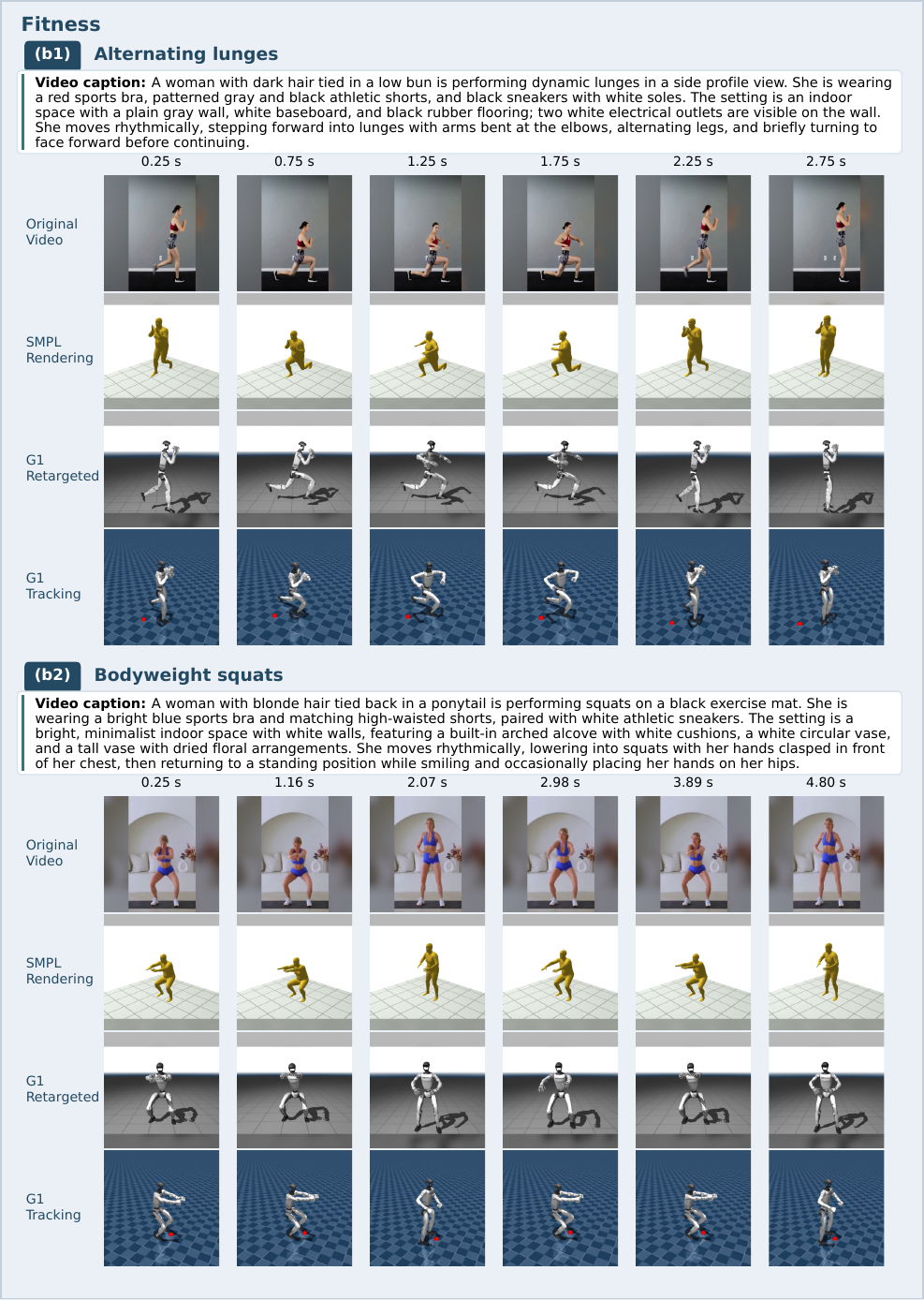}
    \caption[]{Representative \movensdata examples (continued): Fitness, with (b1) alternating lunges and (b2) bodyweight squats. Each half-page case shows four video streams at six clip-relative times.}
    \label{fig:appendix_dataset_category_fitness}
\end{figure}
\clearpage

\begin{figure}[!ht]
    \ContinuedFloat
    \centering
    \includegraphics[width=0.91\textwidth]{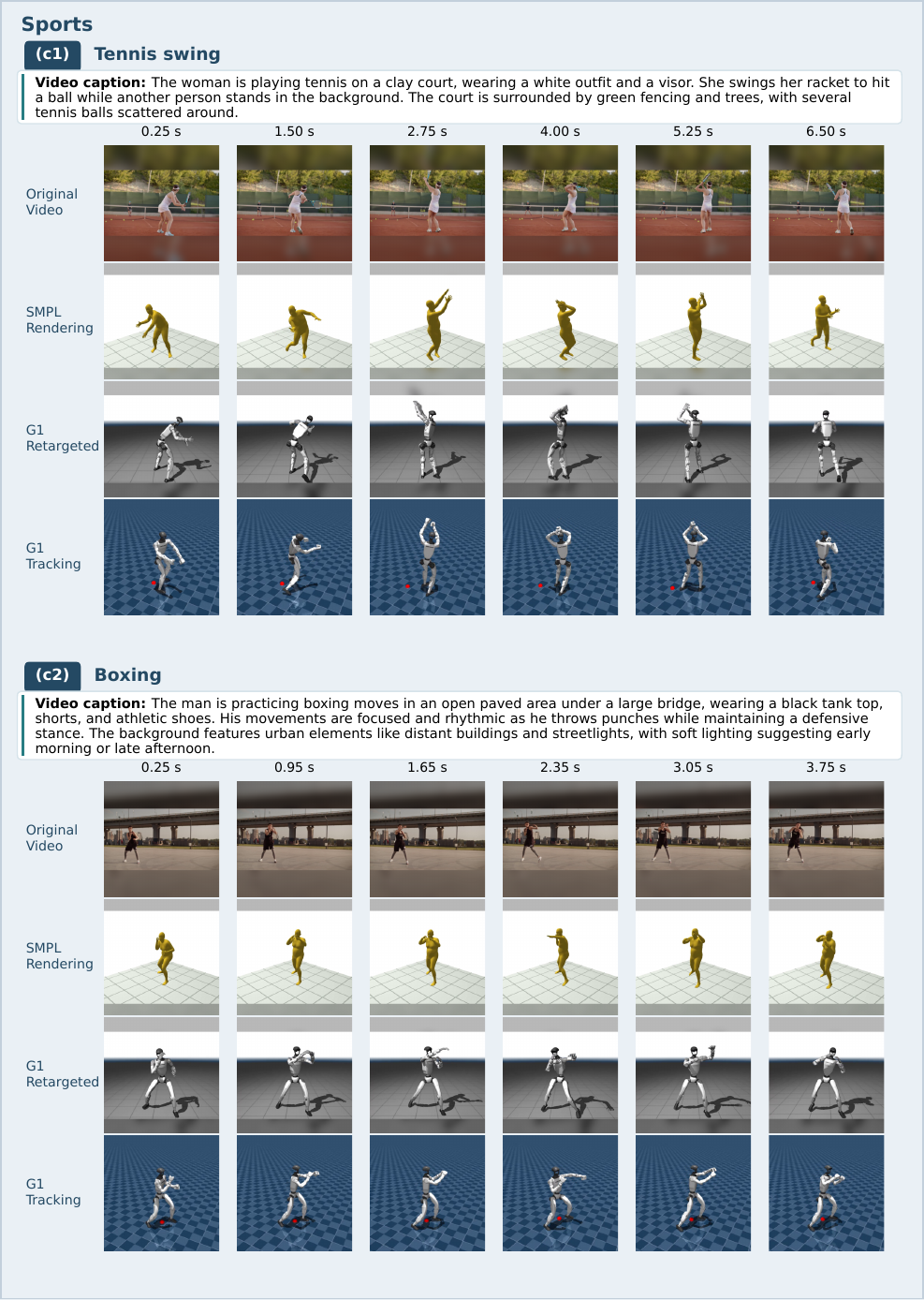}
    \caption[]{Representative \movensdata examples (continued): Sports, with (c1) tennis swing and (c2) boxing. Each half-page case shows four video streams at six clip-relative times.}
    \label{fig:appendix_dataset_category_sports}
\end{figure}
\clearpage

\begin{figure}[!ht]
    \ContinuedFloat
    \centering
    \includegraphics[width=0.91\textwidth]{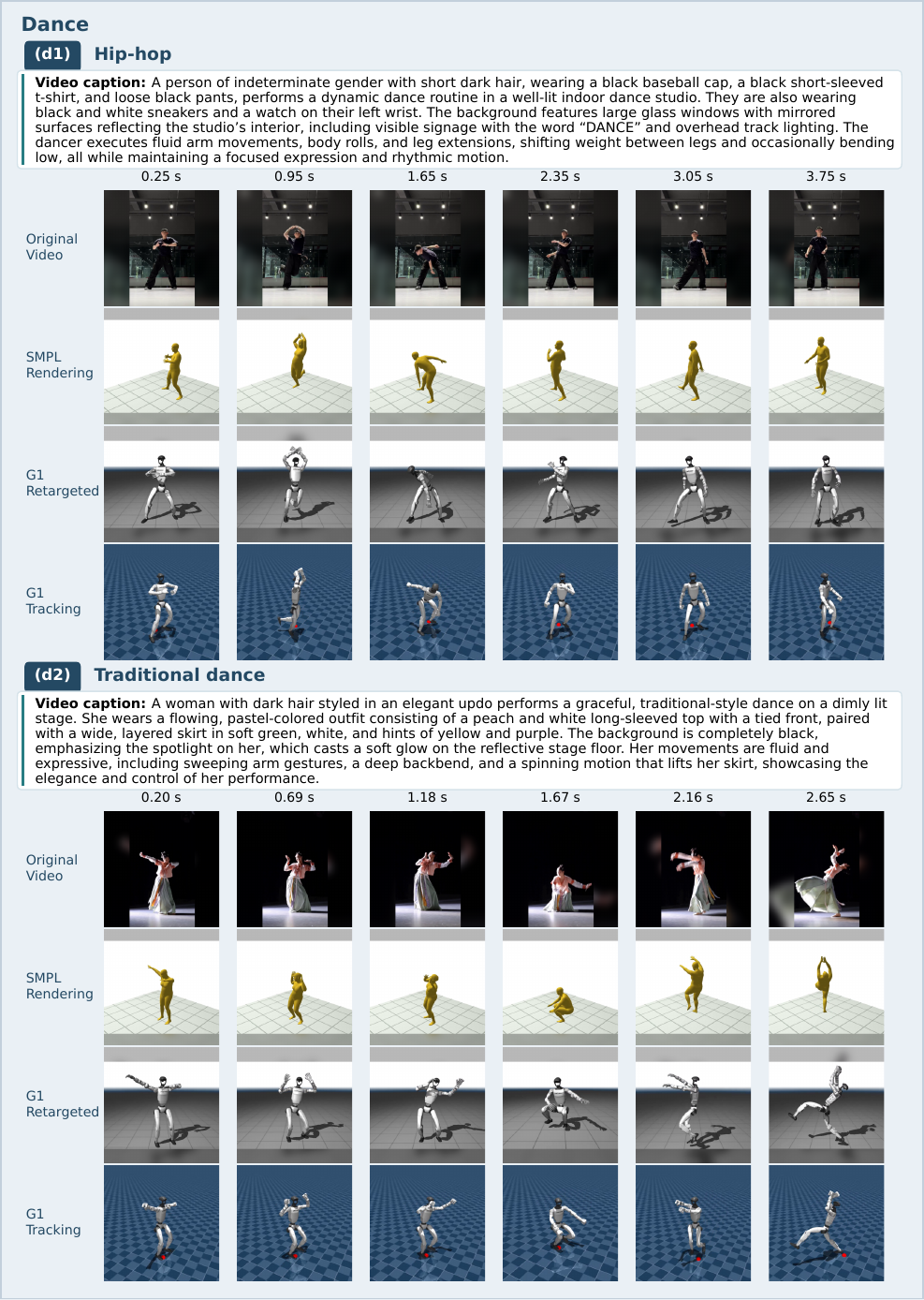}
    \caption[]{Representative \movensdata examples (continued): Dance, with (d1) hip-hop and (d2) traditional dance. Each half-page case shows four video streams at six clip-relative times.}
    \label{fig:appendix_dataset_category_dance}
\end{figure}
\clearpage

\begin{figure}[!ht]
    \ContinuedFloat
    \centering
    \includegraphics[width=0.91\textwidth]{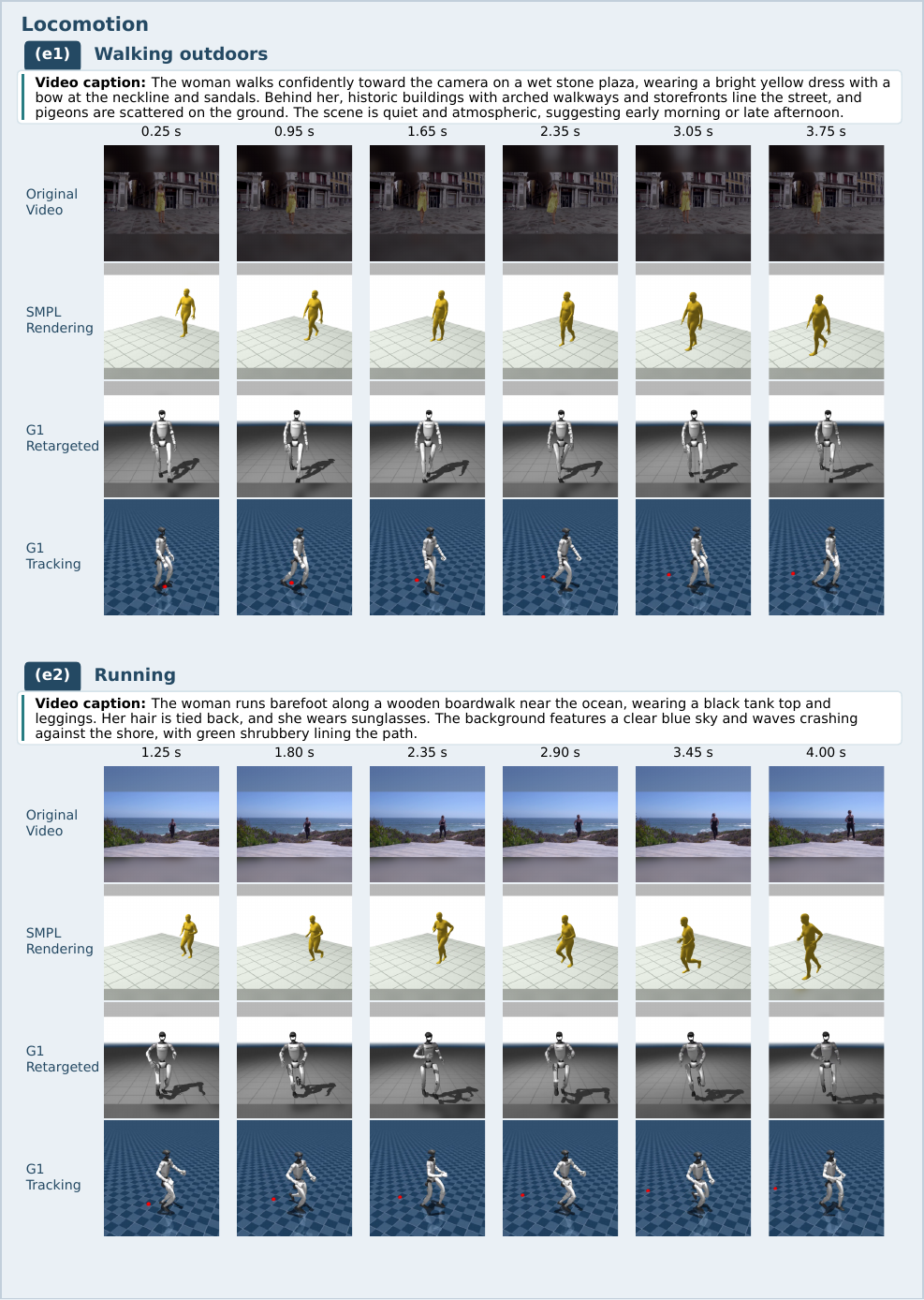}
    \caption[]{Representative \movensdata examples (continued): Locomotion, with (e1) walking outdoors and (e2) running. Each half-page case shows four video streams at six clip-relative times.}
    \label{fig:appendix_dataset_category_locomotion}
\end{figure}
\clearpage

\begin{figure}[!ht]
    \ContinuedFloat
    \centering
    \includegraphics[width=0.91\textwidth]{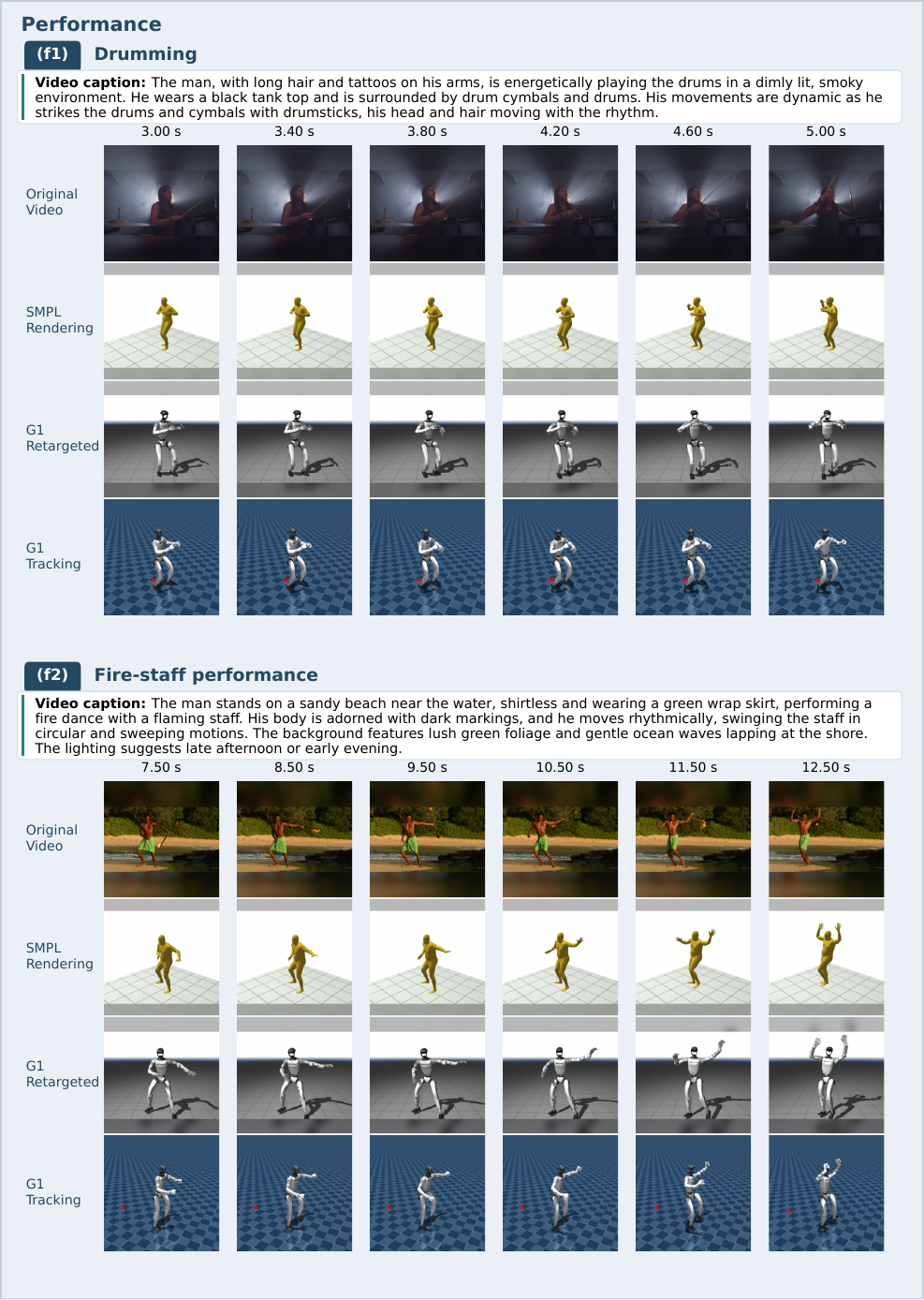}
    \caption[]{Representative \movensdata examples (continued): Performance, with (f1) drumming and (f2) fire-staff performance. Each half-page case shows four video streams at six clip-relative times.}
    \label{fig:appendix_dataset_category_performance}
\end{figure}
\clearpage

\begin{figure}[!ht]
    \ContinuedFloat
    \centering
    \includegraphics[width=0.91\textwidth]{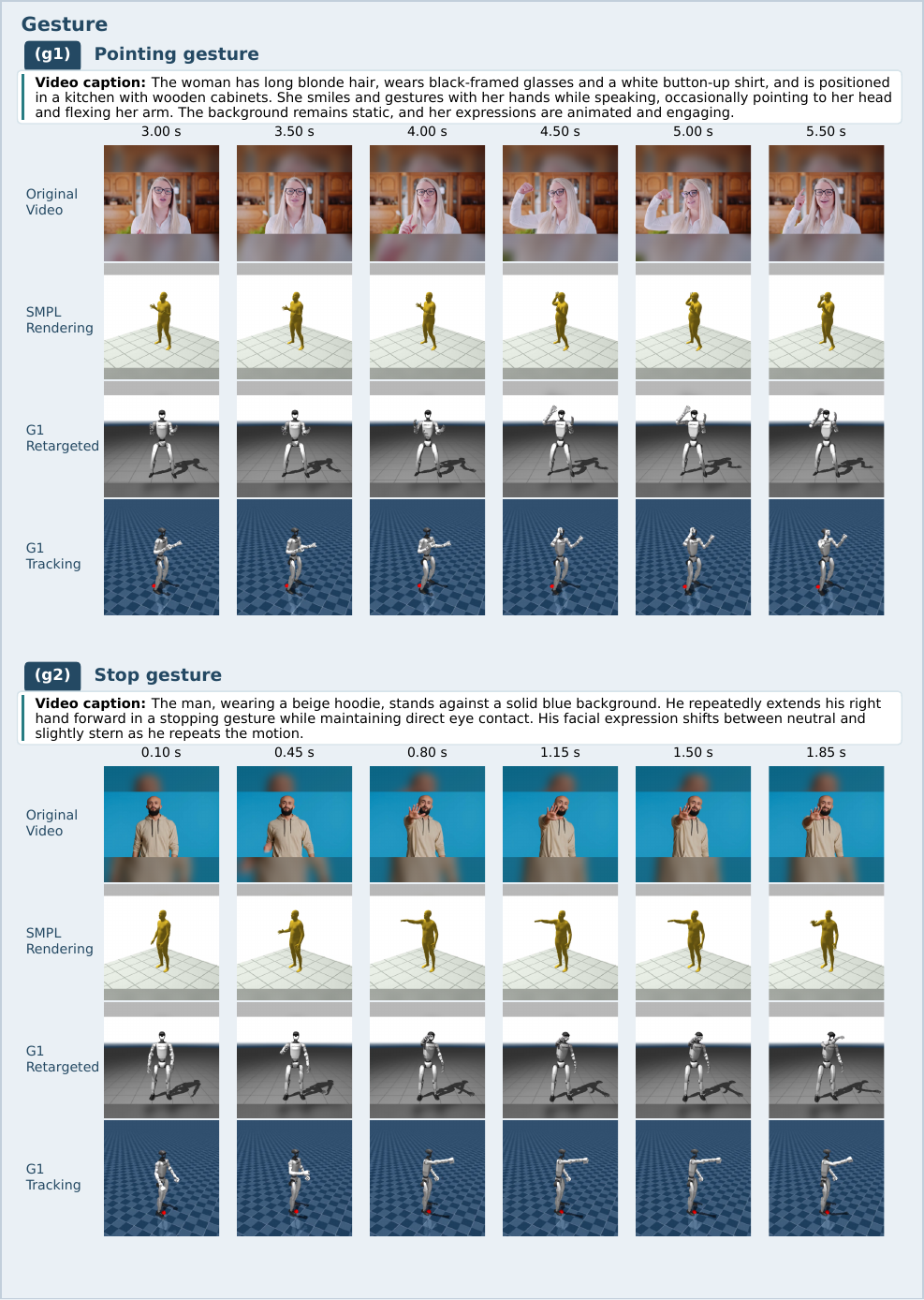}
    \caption[]{Representative \movensdata examples (continued): Gesture, with (g1) pointing gesture and (g2) stop gesture. Each half-page case shows four video streams at six clip-relative times.}
    \label{fig:appendix_dataset_category_gesture}
\end{figure}
\clearpage

\paragraph{Record metadata and video-level descriptions.}
The video examples are accompanied by fields that describe the identity, content, and timing of each record.
Figure~\ref{fig:appendix_dataset_structured_example}(a--c) examines the same gardening sample as Figure~\ref{fig:appendix_dataset_diverse_examples}(a1), so the non-video annotations can be read alongside a concrete visual example.
The six source frames in panel~(a) provide a temporal reference for the record's descriptions.

In panel~(b), \texttt{sample\_id} and \texttt{sample\_name} identify the record, while \texttt{motion\_type} supplies its released action label, \texttt{daily\_gardening}.
The \texttt{objects} field contains the string \texttt{bowl, soil}, connecting the action to the objects involved.
The numeric fields \texttt{duration} and \texttt{fps} describe the clip's temporal extent and sampling rate; \texttt{audio\_valid} retains the released Boolean audio-validity flag, and \texttt{human\_count} records the automatically annotated number of people.
These values are flat Parquet columns: the visual grouping in the figure does not introduce nested metadata objects.
Panel~(b) also displays \texttt{human\_count}, \texttt{human\_age}, \texttt{human\_height}, \texttt{human\_weight}, \texttt{human\_gender}, and \texttt{human\_mood} in a human-attribute group.
These fields contain automatically generated estimates; the figure retains their released values and scalar types.

Panel~(c) reproduces the complete \texttt{video\_caption}.
This description combines the person's appearance, the outdoor setting, the soil and basin, and the overall sequence of filling and carrying the basin.
It supplies context that is difficult to infer from the rendered body alone, and remains associated with the same record as the four video streams.

\begin{figure}[!ht]
    \centering
    \includegraphics[width=\textwidth]{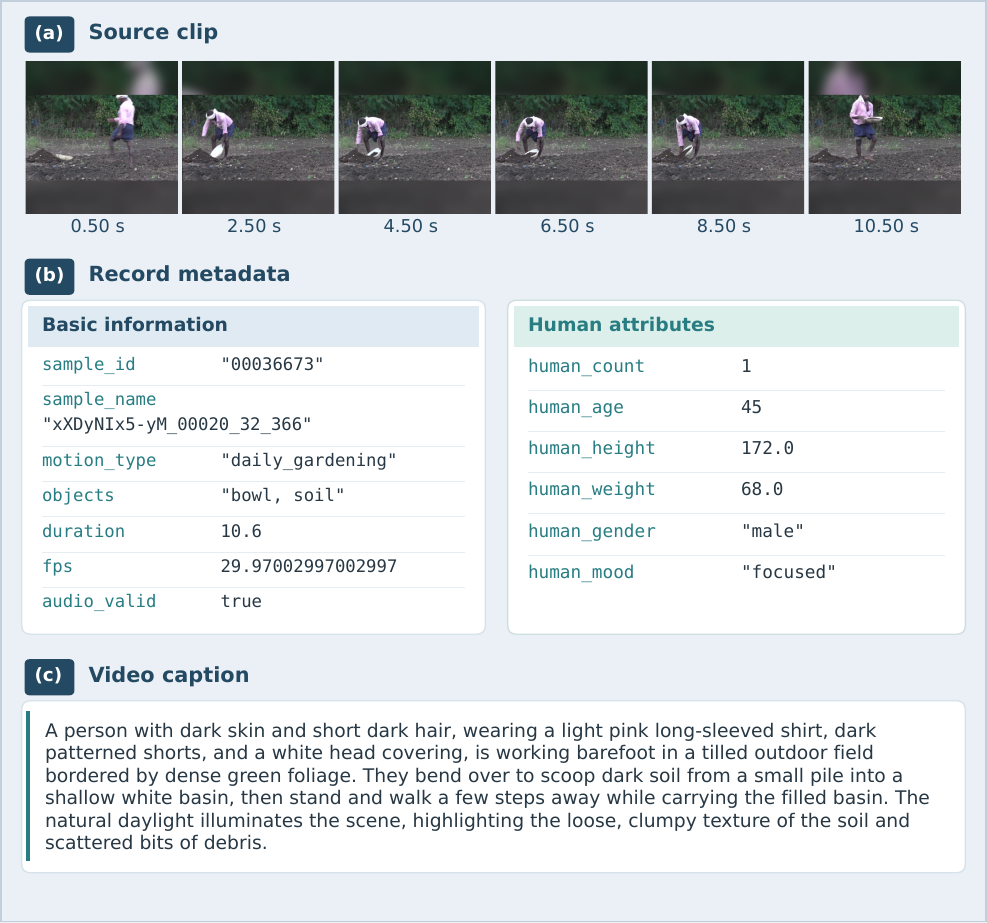}
    \caption{Non-video fields for gardening record \texttt{00036673}: (a) source frames corresponding to Figure~\ref{fig:appendix_dataset_diverse_examples}(a1), (b) all non-caption metadata fields, including automatic human attributes, with original field names and values, and (c) the complete original video caption.}
    \label{fig:appendix_dataset_structured_example}
    \label{fig:appendix_dataset_record_metadata}
\end{figure}
\clearpage

\paragraph{Temporal and body-part motion descriptions.}
Both \texttt{video\_caption} and \texttt{motion\_caption} are stored as strings in the original flat record.
The latter contains three shared intervals: $0.0$--$2.0$\,s, $2.0$--$9.0$\,s, and $9.0$--$10.6$\,s.
Figure~\ref{fig:appendix_dataset_motion_annotations}(d,e) organizes the whole-body and body-part descriptions by these same intervals, preserving their temporal correspondence.
The whole-body annotation progresses from approaching the soil pile, through bending and scooping, to lifting the basin and walking away.

The body-part descriptions change with these phases: the arms transition from walking movements to coordinated scooping and basin support; the legs shift from alternating steps to a planted, bent-knee stance and then resume walking.
The hands change from relaxed and open to scooping and gripping, while the face descriptions shift from looking ahead to looking downward and then forward again.
These are released automatic annotations.
The three intervals cover the full $10.6$\,s clip.

\begin{figure}[!ht]
    \ContinuedFloat
    \centering
    \includegraphics[width=\textwidth]{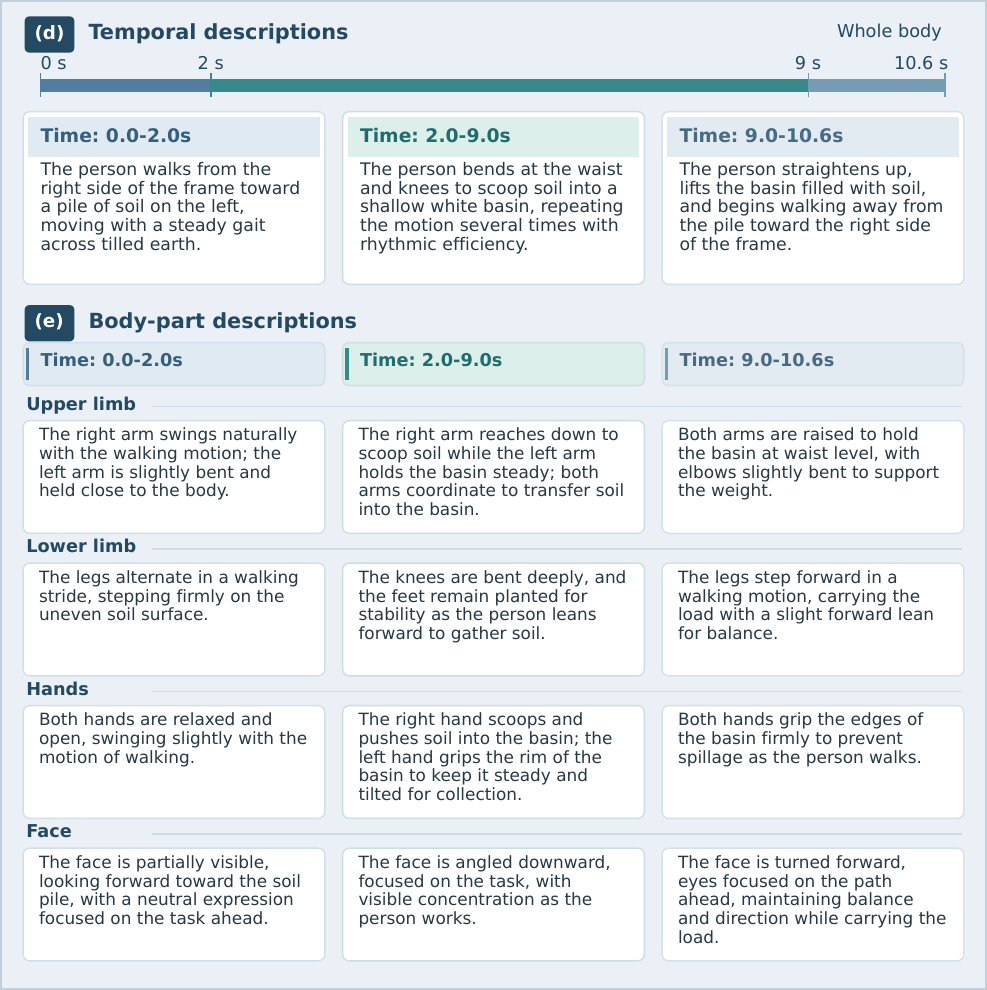}
    \caption[]{Non-video fields (continued): (d) the three original temporal intervals and their whole-body descriptions, and (e) upper-limb, lower-limb, hand, and face descriptions across all three intervals. The columns share the same temporal order, and all motion descriptions reproduce the original \texttt{motion\_caption} text.}
    \label{fig:appendix_dataset_motion_annotations}
\end{figure}
\clearpage

\subsection{Human Motion Quality Assessment and Repair}
\label{app:motion_quality}

Recovered human motions are screened for reconstruction artifacts before humanoid retargeting.
The following quality assessment and repair procedure evaluates local motion chunks, repairs recoverable failures, and rejects sequences that remain invalid after reassessment.

\begingroup
\let\section\subsubsection
\section{Quality Assessment}

We design a hierarchical quality assessment framework for SMPL-X motion sequences by decomposing motion artifacts into five complementary physical and temporal categories: ground contact, temporal jitter, fall risk, body interpenetration, and foot sliding. Following the protocol in~\citep{lee2025phuma}, we partition the input motion into 4-second chunks with a 0.5-second overlap between adjacent chunks, and perform quality assessment on each chunk individually.

\paragraph{Ground Contact.} Following the protocol in~\citep{lee2025phuma}, the ground-contact metric detects motion segments with insufficient foot support. We estimate the per-frame contact state from four foot regions, including the left toe, left heel, right toe, and right heel. For each frame, the maximum contact score over these four regions is used to indicate whether the body has at least one reliable support point. The score is then averaged within each chunk to obtain the Foot Contact Score. A chunk is rejected when this score falls below the threshold, indicating prolonged lack of stable foot contact, which may correspond to floating. Specifically, this can be formulated as:

\begin{equation}
\mathrm{FCS}
=
\frac{1}{T}
\sum_{t=1}^{T}
\max
\left(
s_t^{lh},
s_t^{lt},
s_t^{rh},
s_t^{rt}
\right),
\end{equation}
where $s_t^{lh}$, $s_t^{lt}$, $s_t^{rh}$, and $s_t^{rt}$ denote the graded ground-contact scores at frame $t$ for the left heel, left toe, right heel, and right toe, respectively.

\paragraph{Temporal Jitter.} The temporal-jitter metric captures high-frequency instability in the motion trajectory. We use two complementary measurements. The first is Root Jerk, computed as the mean norm of the third-order finite difference of root translation, which detects abrupt changes in global body motion.

The second is Position High Frequency Ratio. Given a motion sub-clip with $T$ frames and $J$ joints, let
$\mathbf{p}_{t,j} \in \mathbb{R}^{3}$ denote the 3D position of joint $j$ at frame $t$, and let $\mathbf{p}_{t,r}$ be the root joint position. We first remove the global translation by constructing root-relative joint positions:
\begin{equation}
\tilde{\mathbf{p}}_{t,j}
=
\mathbf{p}_{t,j}
-
\mathbf{p}_{t,r}.
\end{equation}

The root-relative joint sequence is then flattened as
$\tilde{\mathbf{y}}(t) \in \mathbb{R}^{3J}$ and transformed into the frequency domain using the short-time Fourier transform (STFT). For a local temporal window centered at frame $i$, the spectrogram magnitude is computed as
\begin{equation}
\mathbf{S}(i, f)
=
\left|
\sum_{k=0}^{L-1}
\tilde{\mathbf{y}}(k)
w(k-i)
e^{-j2\pi f k / N_w}
\right|,
\end{equation}
where $w(\cdot)$ is a Hann window, $L$ denotes the local temporal window size, and $N_w$ is the FFT window length. The magnitude $\mathbf{S}(i,f)$ represents the amplitude of each frequency component at temporal window $i$.

Since natural human motion is expected to be dominated by low-frequency components, we quantify jitter by measuring the proportion of spectral energy above a cutoff frequency $f_c=10$Hz. The frequency threshold is set to 10 Hz, as natural human motion typically contains frequencies below 10 Hz~\citep{crenna2021filtering,zijlstra1996adaptability}. Specifically, the position high frequency ratio is defined as
\begin{equation}
\mathrm{PHFR}
=
\frac{
\sum_{i}
\sum_{f > f_c}
\left\|
\mathbf{S}(i,f)
\right\|_2^2
}{
\sum_{i}
\sum_{f}
\left\|
\mathbf{S}(i,f)
\right\|_2^2
},
\end{equation}

A larger $\mathrm{PHFR}$ indicates that a greater proportion of motion energy lies in high-frequency bands, suggesting stronger temporal jitter or frame-level instability. If either metric exceeds its threshold, the chunk is considered temporally unstable.

\paragraph{Fall Risk.} Following the protocol in~\citep{lee2025phuma}, we use pelvis-height and base-of-support criteria to identify motion segments with abnormal body placement or unstable balance. The pelvis-height criteria exclude segments in which the body is unnaturally positioned with respect to the ground. Specifically, the minimum pelvis-height criterion filters out motions where the body is excessively crouched, collapsed, or lying close to the ground, while the maximum pelvis-height criterion removes segments that exhibit unnatural floating.

The distance-to-base-of-support criteria further enforce physically plausible balance. Since the body center of mass of an SMPL-X mesh is typically correlated with the pelvis and lower-spine region, we use the horizontal-plane projections of the pelvis and spine joints as proxies for body support stability. The base of support is approximated by the convex hull formed by the horizontal-plane projections of the left ankle, right ankle, left foot, and right foot joints. Large distances from the projected pelvis or spine1 joint to this base-of-support region indicate that the body center is insufficiently supported, which often corresponds to imbalance, falling poses, or physically infeasible configurations.

\paragraph{Interpenetration}
The interpenetration metric detects self-collisions in the SMPL-X mesh by measuring distance-field penetration losses over colliding triangle pairs. For each sampled frame, we first use a BVH-based collision search to obtain a set of colliding face pairs
$\mathcal{C}_t=\{(f_s, f_r)\}$, where $f_s$ and $f_r$ denote the source and receiver triangles. For each colliding pair, we follow the distance-field penetration formulation and compute a symmetric penetration energy~\citep{pavlakos2019expressive}:
\begin{equation}
E_c(f_s, f_r)
=
\sum_{\mathbf{v}_s \in f_s}
\left\|
-\Psi_{f_r}(\mathbf{v}_s)\mathbf{n}_s
\right\|_2^2
+
\sum_{\mathbf{v}_r \in f_r}
\left\|
-\Psi_{f_s}(\mathbf{v}_r)\mathbf{n}_r
\right\|_2^2 ,
\end{equation}
where $\Psi_{f_r}(\mathbf{v}_s)$ is the conical distance field induced by triangle $f_r$ and evaluated at vertex $\mathbf{v}_s$, and $\mathbf{n}_s$ and $\mathbf{n}_r$ are the corresponding triangle normals. This symmetric formulation penalizes both directions of penetration, i.e., vertices of the source triangle inside the receiver triangle's distance field and vertices of the receiver triangle inside the source triangle's distance field.

Based on the per-pair penetration energy, we compute two complementary chunk-level interpenetration measurements. The first one, \emph{Interpenetration Max Loss}, captures the deepest local collision:
\begin{equation}
\mathcal{L}_{\mathrm{max}}
=
\max_{t}
\max_{(f_s,f_r)\in\mathcal{C}_t}
E_c(f_s,f_r).
\end{equation}
The second one, \emph{Interpenetration Max Frame Sum Loss}, captures large-area penetration by accumulating all collision losses within each frame and then taking the maximum over time:
\begin{equation}
\mathcal{L}_{\mathrm{sum}}
=
\max_{t}
\sum_{(f_s,f_r)\in\mathcal{C}_t}
E_c(f_s,f_r).
\end{equation}

Based on our empirical observations, these two metrics are complementary and can identify different types of interpenetration that cannot be captured by a single criterion alone. Specifically, $\mathcal{L}_{\mathrm{max}}$ filters interpenetration from the perspective of penetration depth, such as a single finger deeply penetrating the thigh, where the affected region is small but the penetration is severe. In contrast, $\mathcal{L}_{\mathrm{sum}}$ filters interpenetration from the perspective of penetration extent, such as large-area surface overlap between the two hands. Therefore, if either metric exceeds its threshold, we mark the chunk as containing self-interpenetration artifacts.

To reduce false positives caused by SMPL-X topology or part-segmentation artifacts, we ignore stable collision pairs that repeatedly occur in valid poses, as well as predefined local artifact pairs such as eye-head contacts.

\paragraph{Foot Sliding.}
Foot sliding is evaluated using a physical-contact criterion inspired by the Physical Foot Contact (PFC) score~\citep{tseng2023edge}. Instead of assuming that all foot motion is erroneous, this criterion focuses on a physically implausible situation: the body exhibits horizontal acceleration while both feet also move rapidly on the ground plane. In such cases, the feet fail to provide stable support for the body acceleration, which is a strong indicator of sliding artifacts.

In our implementation, the filtering coordinate system is Y-up, and therefore the horizontal plane is the XZ plane. We approximate the body center of mass(COM) by the mean position of all SMPL-X joints. Let $\mathbf{c}_t^{xz}$ denote the XZ-plane projection of this approximate COM at frame $t$, and let $\mathbf{l}_t^{xz}$ and $\mathbf{r}_t^{xz}$ denote the mean XZ-plane positions of the left and right foot contact vertices, respectively. With frame interval $\Delta t$, we compute the horizontal COM acceleration and foot velocities as
\begin{equation}
a_t^{\mathrm{COM}}
=
\left\|
\frac{
\mathbf{c}_{t+1}^{xz}
-
2\mathbf{c}_{t}^{xz}
+
\mathbf{c}_{t-1}^{xz}
}{
\Delta t^2
}
\right\|_2 ,
\end{equation}
\begin{equation}
v_t^{L}
=
\left\|
\frac{
\mathbf{l}_{t+1}^{xz}
-
\mathbf{l}_{t}^{xz}
}{
\Delta t
}
\right\|_2 ,
\quad
v_t^{R}
=
\left\|
\frac{
\mathbf{r}_{t+1}^{xz}
-
\mathbf{r}_{t}^{xz}
}{
\Delta t
}
\right\|_2 .
\end{equation}
The per-frame sliding plausibility score is then defined as
\begin{equation}
s_t
=
a_t^{\mathrm{COM}}
\cdot
v_t^{L}
\cdot
v_t^{R}.
\end{equation}

To make the metric robust to short local artifacts, we compute the score over a temporal window of $0.3$ seconds. For each window $\mathcal{W}$, we normalize the average score by the maximum COM acceleration within that window:
\begin{equation}
\mathrm{PFC}(\mathcal{W})
=
\frac{
\frac{1}{|\mathcal{W}|}
\sum_{t \in \mathcal{W}}
s_t
}{
\max_{t \in \mathcal{W}}
a_t^{\mathrm{COM}}
}.
\end{equation}
Finally, the chunk-level foot-sliding score is the maximum windowed score:
\begin{equation}
\mathrm{PFC}_{0.3}
=
\max_{\mathcal{W}}
\mathrm{PFC}(\mathcal{W}).
\end{equation}
A high $\mathrm{PFC}_{0.3}$ indicates that the body undergoes horizontal acceleration while both feet simultaneously move on the ground plane, suggesting physically implausible foot sliding.

We filter each chunk according to the thresholds summarized in Table~\ref{tab:filtering_thresholds}. For the metrics shared with~\citep{lee2025phuma}, we adopt the thresholds used in the original paper, while the remaining thresholds are determined through a search over a manually curated set of positive and negative samples.

\begin{table}[t]
\centering
\caption{Physics-aware motion filtering metrics and thresholds.}
\label{tab:filtering_thresholds}
\begin{tabular}{ll}
\toprule
Metric & Threshold \\
\midrule
Foot contact score & $> 0.6$ \\
Root jerk & $< 50~\mathrm{m/s^3}$ \\
Position high-frequency ratio & $< 0.003$ \\
Minimum pelvis height & $> 0.6~\mathrm{m}$ \\
Maximum pelvis height & $< 1.5~\mathrm{m}$ \\
Pelvis distance to base of support & $< 6~\mathrm{cm}$ \\
Spine1 distance to base of support & $< 11~\mathrm{cm}$ \\
Interpenetration max loss & $\le 0.0072$ \\
Interpenetration max frame-sum loss & $\le 0.135$ \\
Physical foot contact score, 0.3s window & $< 1.1$ \\
\bottomrule
\end{tabular}
\end{table}

\section{Motion Repair}

For the filtered-out chunks, we use the Kimodo model~\citep{rempe2026kimodo} to search for suitable seed motions and repair the low-quality chunks, thereby obtaining high-quality chunks. Based on the detected failure categories, we formulate motion repair as a localized conditional generation problem centered on each failed chunk. For every abnormal chunk, the system first checks whether its left and/or right neighboring chunks pass the quality filter, and constructs a Kimodo repair job only when at least one reliable neighbor is available. Depending on the available context, the constraint mode is categorized as \textit{both}, \textit{left}, or \textit{right}.

When both neighboring chunks are reliable, the repair job uses 7 keyframes from the left boundary, 7 keyframes from the right boundary, and 5 uniformly sampled keyframes from the original corrupted interval as full-body constraints. When only one side is reliable, the trusted side contributes 7 keyframes, the opposite side contributes 3 keyframes, and the same 5 internal keyframes are added to stabilize the generated motion. These constraints preserve temporal continuity near reliable regions while allowing Kimodo to regenerate the problematic interval.

During the repair process with Kimodo, we observe that when the spatial distance between two constrained frames is small but the temporal interval is relatively long, the generated motion often exhibits an undesirable behavior: the character moves slowly (or at a normal speed) from one constrained frame to the other, followed by a brief pause before continuing the motion. We attribute this phenomenon to the data bias of Kimodo. Since the model is primarily trained on relatively slow motions, it struggles to learn complex transitions required by fast motions, such as dance sequences, and instead tends to generate simple and slow interpolations between the constrained frames. To alleviate this issue, we introduce five additional intermediate constraint frames, which effectively reduce the occurrence of such unnatural transitions.

For each candidate chunk, Kimodo searches over multiple random seeds, stitches the generated SMPL-X segment back into the original motion, and immediately re-applies the quality filter to validate the target chunk and its necessary neighborhood. Once a seed passes validation, the repaired motion is merged into the current sequence, and all successful repairs are re-filtered at the end of the iteration to produce candidates for the next round.

This iterative process continues until all chunks pass, no further repairable candidates remain, or no successful repair is found in the current batch. To improve scalability, the system supports dynamic work queues, enabling localized generative repair to scale to large motion collections while preserving an interpretable validation loop.

\endgroup

\subsection{Physics-Aware Retargeting and Execution Verification}
\label{app:retargeting_verification}

\begingroup
\renewcommand{\section}[1]{}
\let\subsection\subsubsection
\section{Physics-Aware Human-to-Humanoid Motion Retargeting}

The retargeting module converts human motion recovered from in-the-wild videos into joint-space trajectories for a target humanoid. Its input is a canonical SMPL-X sequence~\citep{8953319} recovered by GVHMR for single-person clips or PromptHMR for multi-person clips~\citep{shen2024gvhmr,wang2025prompthmr}, and its output consists of the robot root pose and actuated joint positions at every frame. Direct kinematic retargeting is insufficient for this setting: noisy human estimates and the embodiment gap between SMPL-X and the robot can produce temporally inconsistent motion, foot skating, inaccurate ground contact, and commands close to the robot's actuation limits.

We therefore build on General Motion Retargeting (GMR)~\citep{araujoRetargetingMattersGeneral2025} and introduce an execution-aware, clip-level refinement stage. GMR provides a reliable kinematic initialization, while our refinement jointly optimizes the entire sequence using temporal, contact, and hardware-limit regularizers. This division is important: the GMR-based stage preserves the source motion and bridges the kinematic mismatch, whereas the proposed stage specifically addresses artifacts that cannot be resolved by independent per-frame inverse kinematics (IK).

\subsection{Kinematic Initialization}

We first transform the recovered SMPL-X motion from its $y$-up convention to the $z$-up robot world frame. Following GMR, we compensate for morphological differences with non-uniform local scaling and define a sparse, robot-specific correspondence between human joints and robot bodies. Each correspondence specifies a position target, an orientation target, optional frame offsets, and their relative weights. This explicit mapping is lightweight and interpretable, and can be adapted to a new embodiment without retraining a model.

For each frame, the robot configuration $\mathbf{x}$ is obtained by minimizing weighted Cartesian errors,

\begin{equation}
    \mathcal{E}_{\mathrm{IK}}(\mathbf{x})
    =
    \sum_{k \in \mathcal{K}}
    \left(
        w_k^p \left\|
            \mathbf{p}_k(\mathbf{x})-\hat{\mathbf{p}}_k
        \right\|_2^2
        +
        w_k^R \left\|
            \log\!\left(
                \mathbf{R}_k(\mathbf{x})^{\mathsf{T}}\hat{\mathbf{R}}_k
            \right)
        \right\|_2^2
    \right),
    \label{eq:retarget_ik}
\end{equation}

subject to the robot joint limits. Here, $\mathbf{p}_k$ and $\mathbf{R}_k$ denote robot-body positions and orientations computed by forward kinematics, and $\hat{\mathbf{p}}_k$ and $\hat{\mathbf{R}}_k$ are the scaled human targets. We retain GMR's coarse-to-fine solve: end-effector positions and body orientations are established first, after which all mapped body positions are fine-tuned. The previous frame initializes the current frame. We use the resulting sequence only as a kinematic warm start for the refinement below; the standard IK derivation is omitted for brevity.

\subsection{Physics-Aware Sequence Refinement}

The main extension over GMR is to replace purely frame-wise correction with joint optimization over a complete clip. For a sequence of $T$ frames, we optimize the root translation $\mathbf{r}_{1:T}$ and actuated joint positions $\mathbf{q}_{1:T}$. We keep the root orientations from the IK initialization fixed because their global heading and body attitude are already well determined; doing so also prevents the refinement from changing the semantic content of the motion.

Let $\bar{\mathbf{x}}_{1:T}$ denote the IK initialization and
$\mathbf{x}_t=(\mathbf{r}_t,\bar{\mathbf{R}}^{\mathrm{root}}_t,\mathbf{q}_t)$ the refined state. We optimize
\begin{equation}
\begin{aligned}
    \mathcal{L}_{\mathrm{seq}}
    ={}&
    \lambda_{\mathrm{fid}}\mathcal{L}_{\mathrm{fid}}
    + \lambda_{\mathrm{temp}}\mathcal{L}_{\mathrm{temp}}
    + \lambda_{\mathrm{lim}}\mathcal{L}_{\mathrm{lim}} \\
    &+
    \lambda_{\mathrm{gnd}}\mathcal{L}_{\mathrm{gnd}}
    + \lambda_{\mathrm{skate}}\mathcal{L}_{\mathrm{skate}} .
    \label{eq:retarget_total}
\end{aligned}
\end{equation}

The five terms have complementary roles:

\paragraph{Motion fidelity.}
We preserve the source action by matching the refined robot keypoints to those of the IK initialization,

\begin{equation}
    \mathcal{L}_{\mathrm{fid}}
    =
    \frac{1}{T|\mathcal{K}|}
    \sum_{t=1}^{T}\sum_{k\in\mathcal{K}}
    w_k
    \left\|
        \mathbf{p}_k(\mathbf{x}_t)
        -\mathbf{p}_k(\bar{\mathbf{x}}_t)
    \right\|_2^2 .
\end{equation}
This term acts as a trust region: it permits local corrections required by the other objectives without allowing the optimized trajectory to drift to a different action.

\paragraph{Temporal consistency.}
Frame-wise IK can amplify pose-estimation noise into abrupt joint or root motion. We suppress these high-frequency components by penalizing the temporal variation of joint and root velocities,
\begin{equation}
    \mathcal{L}_{\mathrm{temp}}
    =
    \frac{1}{T-2}
    \sum_{t=2}^{T-1}
    \left(
        \left\|\Delta^2\dot{\mathbf{q}}_t\right\|_2^2
        +
        \left\|\Delta^2\dot{\mathbf{r}}_t\right\|_2^2
    \right),
\end{equation}
where $\Delta^2$ is the centered second-difference operator, and velocities are computed using the input frame rate. Unlike post-hoc frame smoothing alone, this regularizer is optimized together with contact and feasibility, so smoothing cannot independently move a support foot away from its contact target.

\paragraph{Execution limits.}

We use hinge penalties to keep joint positions and velocities inside a safety margin of the robot specifications:

\begin{equation}
\begin{aligned}
    \mathcal{L}_{\mathrm{lim}}
    = \frac{1}{T D}\sum_{t,d}
    \big(
        &[q_d^{\mathrm{lo}}-q_{t,d}]_{+}
        +[q_{t,d}-q_d^{\mathrm{hi}}]_{+} \\
        &+ \Delta t\,[|\dot{q}_{t,d}|-\dot{q}_d^{\max}]_{+}
    \big),
    \label{eq:retarget_limits}
\end{aligned}
\end{equation}

where $[a]_{+}=\max\{a,0\}$, $D$ is the number of actuated joints, and $q_d^{\mathrm{lo}}$ and $q_d^{\mathrm{hi}}$ are conservative limits set slightly inside the mechanical range. We assign this term the largest weight because violations invalidate a trajectory regardless of its visual fidelity.

\paragraph{Contact consistency.}
Foot-contact labels are inferred from the reconstructed SMPL-X mesh. We estimate the ground level robustly from foot vertices and mark a toe or heel as being in contact when its distance to the ground is below a small threshold. For each robot foot point $f\in\mathcal{F}$, the transferred contact indicator
$c_{t,f}\in\{0,1\}$ activates two losses:

\begin{equation}
\begin{aligned}
    \mathcal{L}_{\mathrm{gnd}}
    &=
    \frac{1}{T|\mathcal{F}|}
    \sum_{t,f} c_{t,f}\,h_{t,f}^{\,2}, \\
    \mathcal{L}_{\mathrm{skate}}
    &=
    \frac{1}{\sum_{t,f}c_{t,f}+\epsilon}
    \sum_{t,f}c_{t,f}
    \left\|\mathbf{v}_{t,f}^{xy}\right\|_2^2 ,
    \label{eq:retarget_contact}
\end{aligned}
\end{equation}

where $h_{t,f}$ is the foot-point height above the ground and
$\mathbf{v}_{t,f}^{xy}$ is its horizontal velocity. The first term removes floating and penetration during support, while the second anchors the support foot and reduces skating. Contact is applied selectively rather than forcing both feet onto the ground throughout the sequence, which preserves flight and swing phases in dynamic motions.

\subsection{Optimization and Post-Processing}

We initialize the optimized variables with the GMR result and minimize the loss function using Adam~\citep{kingmaAdamMethodStochastic2017}.

The loss weights follow a feasibility-first hierarchy: joint limits receive the highest priority, followed by ground contact, motion fidelity and temporal consistency, with skating used as a finer correction. This ordering allows the method to make small pose adjustments when the human reference is incompatible with the robot, while retaining the recognizable structure of the action.

Finally, we apply a fourth-order Butterworth low-pass filter~\citep{butterworth1930theory} to the root translation, root orientation, and joint trajectories, using component-specific cutoff frequencies. Before filtering rotations, quaternion signs are made temporally consistent; the filtered quaternions are then renormalized. Joint positions are clamped once more to the conservative limits. The resulting procedure is a lightweight physics-aware surrogate rather than full dynamic trajectory optimization: it does not solve for torques or contact forces, but removes the dominant execution artifacts at dataset scale while preserving the motion semantics provided by the kinematic initializer.

\endgroup

\subsubsection{Tracking-Based Execution Verification}

Each retargeted trajectory is verified with a selected tracking policy using the common acceptance threshold.
Full-bank validation can instead evaluate up to four independently trained policies; in this setting, one successful rollout among the evaluated policies is sufficient for acceptance.
Falls or large tracking errors can trigger a new retargeting seed and another rollout before final rejection.
The retained record stores the synchronized human motion, humanoid reference, selected rollout, available per-policy scores, and verification statistics.

\section{Additional Movens-Zero Method Details}
\label{app:method_details}

\subsection{Robot-Centric Motion Representation}
\label{app:motion_representation}
Each Unitree G1 frame uses 40 channels: 29 absolute joint angles, two planar root-displacement channels, six root-orientation channels, one world-frame root height, and two binary foot-contact channels.
The gravity-aligned anchor is derived only from observed poses and is held fixed within each chunk.
For root positions $\mathbf{r}_j$, the planar displacement is the $xy$ part of $\mathbf{A}_n^{\top}(\mathbf{r}_j-\mathbf{r}_{j-1})$.
The six orientation channels concatenate the first two columns of $\mathbf{A}_n^{\top}\mathbf{R}_j$.
Decoding orthonormalizes these columns, restores the world rotation with $\mathbf{A}_n$, and accumulates planar displacements from the latest observed root position.
The contact channels use the representation's leg-motion and vertical-root-motion heuristic.
When only the latest pose is supplied, its state displacement is zero.
Target motions, generated chunks, and the latest robot state use this shared channel layout.

\subsection{Omni-Modal Condition Encoding}
\label{app:condition_encoding}
Text, sampled RGB frames, and audio are encoded by a shared omni-modal context encoder.
Spoken instructions and music use the same audio front end with different task prompts.
The resulting condition tokens are projected into the action-expert width and jointly attended with the current robot-state token and noisy future-motion tokens.

\subsection{Action-Expert Scale Configurations}
\label{app:scale_configurations}
We vary only the capacity of the flow-based action expert in the model-scaling study, while keeping the frozen omni-modal context encoder, robot-centric motion representation, and per-head dimension fixed.
Table~\ref{tab:model_scale_architecture} summarizes the four architectures.

\begin{table*}[htbp]
\centering
\small
\setlength{\tabcolsep}{7pt}
\caption{Action-expert architectures used in the model-scaling experiments. DiT parameters count the transformer backbone, while action-expert parameters count the complete trainable action side.}
\label{tab:model_scale_architecture}
\begin{tabular}{lrrrrrr}
\toprule
\textbf{Scale} & \textbf{Hidden width $D$} & \textbf{Depth} &
\textbf{Heads} & \textbf{Head dimension} & \textbf{DiT parameters} &
\textbf{Action-expert parameters} \\
\midrule
S  & 768   & 3 & 12 & 64 & 56.1M  & 64.2M  \\
M  & 1,152 & 4 & 18 & 64 & 174.0M & 187.9M \\
L  & 1,536 & 6 & 24 & 64 & 479.2M & 500.1M \\
XL & 1,920 & 7 & 30 & 64 & 881.4M & 910.5M \\
\bottomrule
\end{tabular}
\end{table*}

\subsection{Task-Mixed Training and Online Generation}
\label{app:task_mixed_generation}
Training mixes text-, video-, and audio-conditioned examples across seven generation tasks in proportion to their data sizes, jointly optimizing one motion representation, one flow-matching objective, and one action expert. Text-, speech-, music-, and video-conditioned generation are all directly supported by this task-mixed training stage.
At inference time, the model predicts overlapping future-motion chunks and commits only a short prefix of each prediction.
The latest observed robot state conditions the next chunk, allowing the generator to update future motion without discontinuously replacing the segment already being executed.

\section{Training Details}
\label{app:training_details}

\paragraph{Training recipe at a glance.}
We freeze Qwen2.5-Omni-3B and train the motion-generation side of \movenszero from its precomputed last-layer condition tokens.
Task-mixed training samples the seven task templates in Table~\ref{tab:training_task_mix} according to their training-set sizes and directly learns text-, speech-, music-, and video-conditioned generation in a single joint training stage.
For each model scale, all tasks share the same 40-dimensional G1 motion representation, action expert, flow-matching objective, and \mmrtc continuation task.
Tables~\ref{tab:training_task_mix} and~\ref{tab:training_hyperparameters} summarize the complete recipe; the remainder of this section explains the few settings that are specific to multimodal real-time motion generation.

\begin{table}[t]
\centering
\small
\setlength{\tabcolsep}{5pt}
\caption{Seven-task mixture used for action-expert training. A ``stream'' denotes a source-specific dataset, whereas a ``task'' denotes a conditioning objective. Sampling probabilities are computed from the enabled training-split sizes.}
\label{tab:training_task_mix}
\resizebox{\columnwidth}{!}{
\begin{tabular}{lcll}
\toprule
\textbf{Task} & \textbf{Streams} & \textbf{Exposure} & \textbf{Input} \\
\midrule
Text-to-Motion        & 1 & 19.48\% & Generate a motion sequence that matches the following description. Description: [text description]. \\
\midrule
Music-to-Motion       & 2 & 2.83\%  & [Music] \\
(Music, Text)-to-Motion       & 2 & 2.83\%  & [Music] Generate a dance motion that follows the rhythm and energy of this music. \\
\midrule
Speech-to-Motion      & 1 & 9.74\% & [Spoken command] \\
(Speech, Text)-to-Motion      & 1 & 9.74\% & [Spoken command] Parse the utterance in the audio and move according to that instruction. \\
\midrule
Video-to-Motion  & 3 & 27.69\% & [Video] \\
(Video, Text)-to-Motion  & 3 & 27.69\% & [Video] Follow the visual demonstration. \\
\midrule
\textbf{Total} & \textbf{7} & \textbf{100\%} & shared action expert and motion target \\
\bottomrule
\end{tabular}}
\end{table}

\begin{table*}[t]
\centering
\small
\setlength{\tabcolsep}{6pt}
\caption{Training and inference settings for the L action expert.}
\label{tab:training_hyperparameters}
\resizebox{\textwidth}{!}{
\begin{tabular}{lll|lll}
\toprule
\multicolumn{3}{c|}{\textbf{Model and motion}} & \multicolumn{3}{c}{\textbf{Optimization}} \\
\midrule
Context features & Qwen2.5-Omni-3B, frozen & 2,048-D & Optimizer & AdamW & $\beta=(0.9,0.999)$ \\
Action expert & 6 blocks, 24 heads & width 1,536; head dim. 64 & Learning rate & $10^{-4}$ & minimum $5\!\times\!10^{-7}$ \\
Action-expert params & 479.2M DiT & 500.1M total & Schedule & cosine & 1,000-step warmup \\
Motion/state token & 40-D & q01--q99 norm. + clip & Weight decay / $\epsilon$ & 0 & $10^{-8}$ \\
Progress / time embed. & 5-D / 256-D & --- & Maximum steps & $10^{6}$ & seed 292285 \\
\midrule
\multicolumn{3}{c|}{\textbf{Temporal and modality sampling}} & \multicolumn{3}{c}{\textbf{Distributed execution}} \\
\midrule
Motion rate & 50\,Hz & 30-frame chunk & Parallelism & 8-GPU DDP & bfloat16 \\
Execute / overlap & 20 / 10 frames & 0.4 / 0.2\,s & Micro/global batch & 64 / 512 & accumulation 1 \\
Training prefix & $d\sim\mathrm{Unif}\{0,\ldots,10\}$ & per example & Validation interval & 1,000 steps & 20 batches/task \\
Music & 16\,kHz & 0.6\,s local window & Checkpoint interval & 1,000 steps & keep latest + best \\
Video clip & 8\,FPS & 50,176-pixel budget & Euler solver & 10 steps & $\sigma\in[0.05,1)$ \\
\bottomrule
\end{tabular}
}
\end{table*}

\subsection{Cached Omni-Modal Conditions}
\label{app:cached_conditions}
The frozen context encoder produces a sequence of 2,048-dimensional tokens and a padding mask for each text, audio, or video condition.
We cache these tensors before action-expert training and linearly project them to the corresponding action-expert width at runtime.
The scale-specific architectures are listed in Table~\ref{tab:model_scale_architecture}.
Q/K normalization is evaluated in full precision for numerical stability.
Only the action side is optimized, so every update is devoted to learning the shared condition--motion interface rather than repeatedly evaluating or fine-tuning the understanding backbone.

For music, speech, and video, each modality-aligned example is instantiated both with and without a randomized text instruction, yielding the six paired tasks listed alongside text-to-motion in Table~\ref{tab:training_task_mix}.
We use data-proportional sampling instead of equal task weights, and a shard-aware distributed sampler assigns balanced rank-local blocks while preserving the requested task mixture.
Different ranks may therefore process different tasks in the same optimization step, after which DDP aggregates their gradients into the shared action expert.

\subsection{Aligned Motion-Chunk Construction}
\label{app:motion_chunk_construction}
Target motions are sampled at 50\,Hz and encoded with the 40-dimensional channel layout in Appendix~\ref{app:motion_representation}.
Per-channel affine normalization maps the 1st- and 99th-percentile bounds of the G1 training set to $-1$ and $1$.
Both normalized target motions and normalized state inputs are then clipped to $[-1,1]$.
Each example contains a 30-frame target chunk and the latest 40-dimensional robot state, which is the only proprioceptive token supplied to the model.
Inference applies the same scaling and clipping to the observed state.
Music uses a 0.6\,s local window sampled from 16\,kHz audio, speech uses the full spoken command, and video is decoded at 8\,FPS with a 50,176-pixel tokenization budget.
Prompted variants append a randomly sampled text instruction to the modality input, whereas the condition-only variants intentionally leave the user text empty.

For \mmrtc, the training prefix is not fixed to the ten-frame deployment overlap.
Instead, we independently draw $d\in\{0,\ldots,10\}$ for every example, keep the first $d$ target frames clean at flow time zero, and exclude them from the loss.
The remaining suffix follows the flow path in Equation~\ref{eq:flow_path} with a noise level stratified over the global batch in $[0.05,1)$ and rank-independent Gaussian noise.
At deployment, the ten unexecuted frames of the previous prediction are first re-encoded in the current anchor using the latest observed state, while preserving their predicted contact channels. This prefix is restored before every Euler update and after the final update, yielding an exact continuity constraint in the updated representation.

\subsection{Loss and Optimization}
\label{app:training_loss}
The flow loss uses relative weights of 1, 3, 10, 15, and 0.1 for absolute joint angles, local root displacement, root rotation, root height, and foot contact, respectively.
These multipliers are applied after assigning the common base weight $1/40$ to every motion dimension.
We train with AdamW, bfloat16 mixed precision, and the cosine schedule in Table~\ref{tab:training_hyperparameters}; gradient clipping is disabled, while gradient norms are logged for diagnosis.
The configured one-million-step value is an upper bound rather than a claim that every scaling run consumes the entire budget.

\subsection{Validation and Reproducibility}
\label{app:validation_reproducibility}
Before optimization, the launcher audits source-disjoint split metadata and requires all cached condition tensors and runtime indexes to be present.
Every 1,000 optimizer steps, we evaluate 20 fixed mini-batches from each of the seven task templates and report both aggregate loss and breakdowns by task template and motion-channel group.
We save at the same interval, retaining the latest regular checkpoint and the checkpoint with the lowest validation loss.
The run configuration, learning rate, task-specific losses, group losses, and gradient norms are recorded with Weights \& Biases.
The reported configuration starts fresh from the frozen pretrained context representation and does not resume an earlier action-expert run.

\section{Evaluation Protocols}
\label{app:evaluation_protocols}

\subsection{Benchmarks and Baselines}
\label{app:benchmarks_baselines}
We evaluate on the public OMG benchmark, the EMDB and RICH datasets, and the source-disjoint \movensintent benchmark.
Text and audio follow the released OMG manifests when applicable.
Video follows the GVHMR camera-space protocol on RICH and EMDB-1 and its world-space protocol on RICH and EMDB-2.
The \movensintent benchmark uses the same camera- and world-space metrics.
All baseline outputs are converted to the same Unitree G1 representation and executed with identical simulator and controller settings.

\clearpage
\subsection{Movens-Intent Benchmark Examples}
\label{app:intent_examples}
Figure~\ref{fig:appendix_intent_examples} expands the four-track overview in Figure~\ref{fig:intent_benchmark_overview} with two records per track.
Each case presents a user instruction, the original input condition, and six frames from its released G1 reference motion.
The instructions express a user's request to perform described movements, act on spoken language, dance to music, or imitate a person in a video.
Frame labels use reference-relative seconds, and each six-frame reference sequence is arranged from left to right in one row.
The first case of each track also appears in the main-paper overview.

\paragraph{Text conditions.}
The prompts specify coordinated body actions and their temporal order.
The two records illustrate arm movements with turning and a leg lift, and repeated squat-and-jump motion.

\begin{figure}[!ht]
    \centering
    \includegraphics[width=\textwidth]{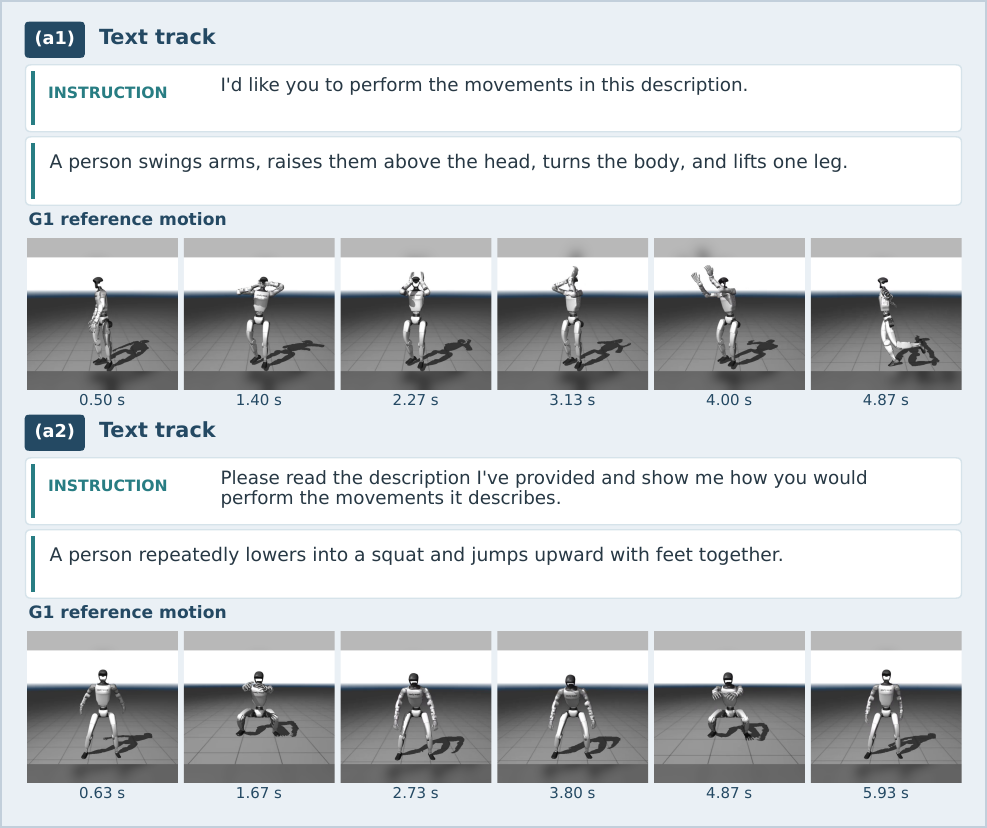}
    \caption{Text benchmark examples: (a1) arm movements with turning and a leg lift; (a2) repeated squat-and-jump motion. Each case pairs a user instruction and the original text description with six G1 reference frames, read from left to right.}
    \label{fig:appendix_intent_examples}
    \label{fig:appendix_intent_text}
\end{figure}
\clearpage

\paragraph{Speech conditions.}
The complete released transcripts accompany the original speech waveforms.
Audio duration is measured from the actual waveform, while reference length follows the motion record; these examples do not assume word-level synchronization between speech and movement.

\begin{figure}[!ht]
    \ContinuedFloat
    \centering
    \includegraphics[width=\textwidth]{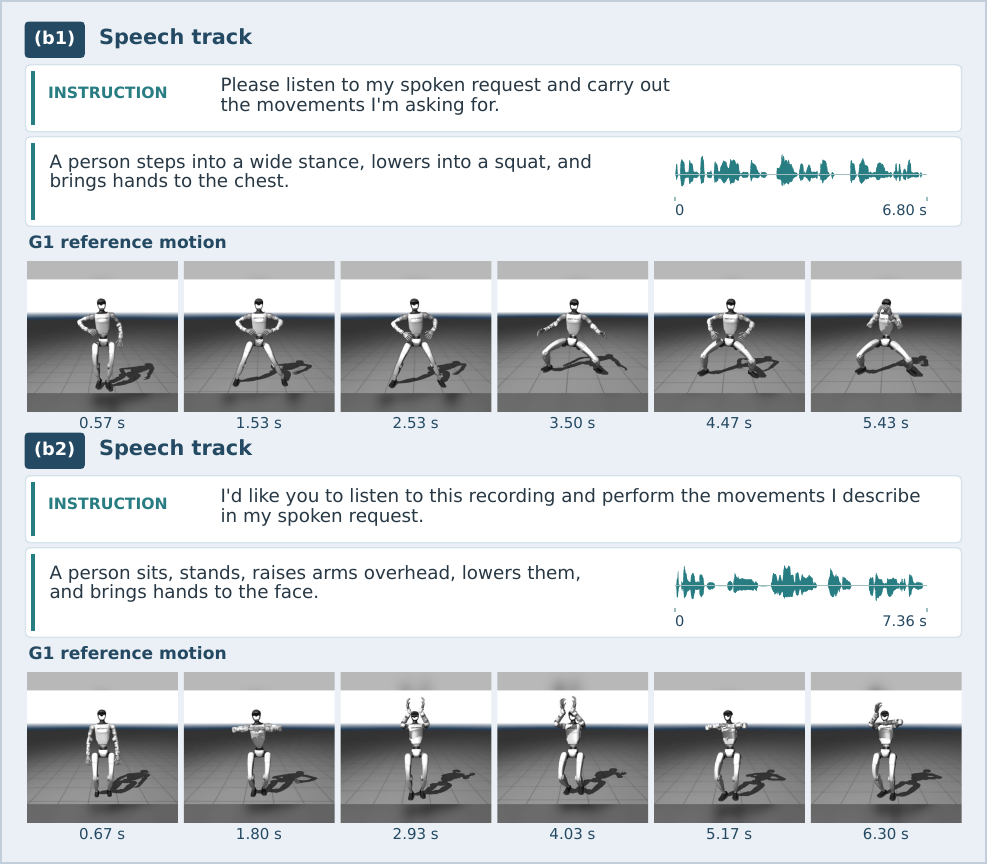}
    \caption[]{Speech benchmark examples (continued): (b1) stepping into a wide squat with hands at the chest; (b2) standing from sitting and moving the arms. Each case shows a user instruction, the original transcript and complete speech waveform, and six G1 reference frames. Audio and reference motion use separate time axes.}
    \label{fig:appendix_intent_speech}
\end{figure}
\clearpage

\paragraph{Music conditions.}
The complete music waveforms are shown above their paired reference motions, retaining each audio file's original sampling rate and duration.
The two records illustrate different combinations of steps, arm movements, and body orientation.
Waveform envelopes describe the input signal and do not substitute for beat-alignment evaluation.

\begin{figure}[!ht]
    \ContinuedFloat
    \centering
    \includegraphics[width=\textwidth]{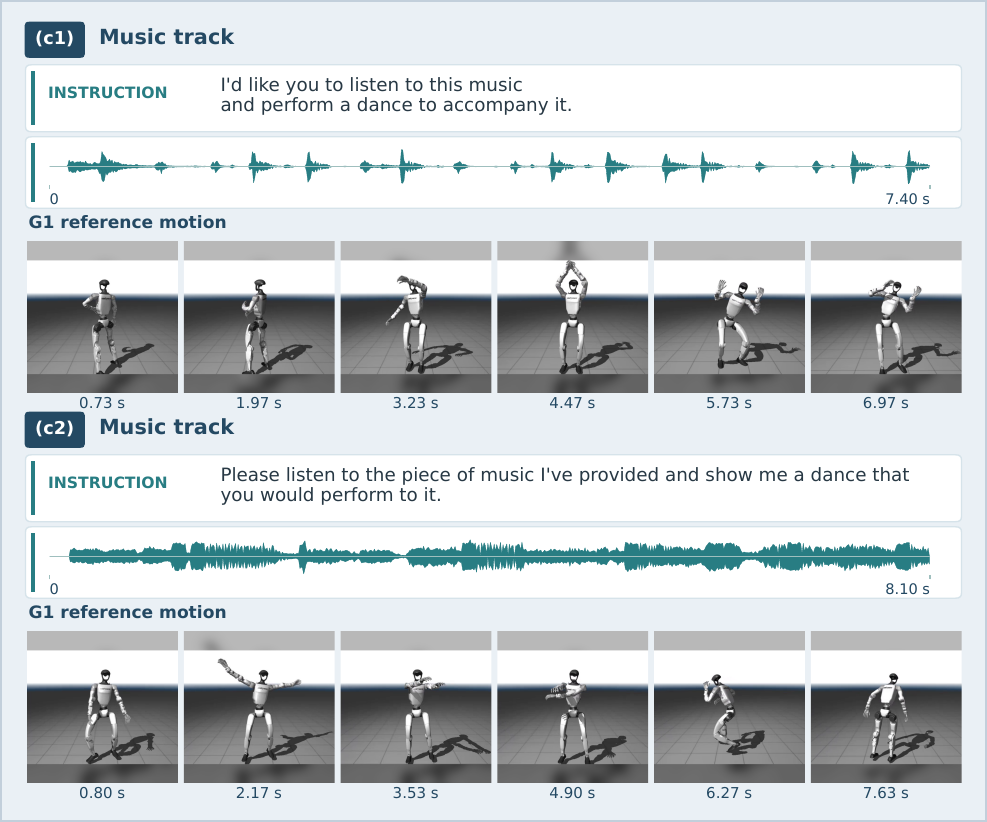}
    \caption[]{Music benchmark examples (continued): (c1--c2) user requests to dance to two music inputs. Each case pairs a dance request and complete music waveform with six G1 reference frames in a single row. Waveform labels report audio time; frame labels report reference time.}
    \label{fig:appendix_intent_music}
\end{figure}
\clearpage

\paragraph{Video conditions.}
RGB input and G1 reference frames are sampled at the same reference-relative times, using the nearest available frame in each media stream.
The two examples cover an indoor dance sequence and a fitness sequence.
The displayed G1 trajectories are expert-verified ground-truth motions paired with the corresponding source videos.

\begin{figure}[!ht]
    \ContinuedFloat
    \centering
    \includegraphics[width=\textwidth]{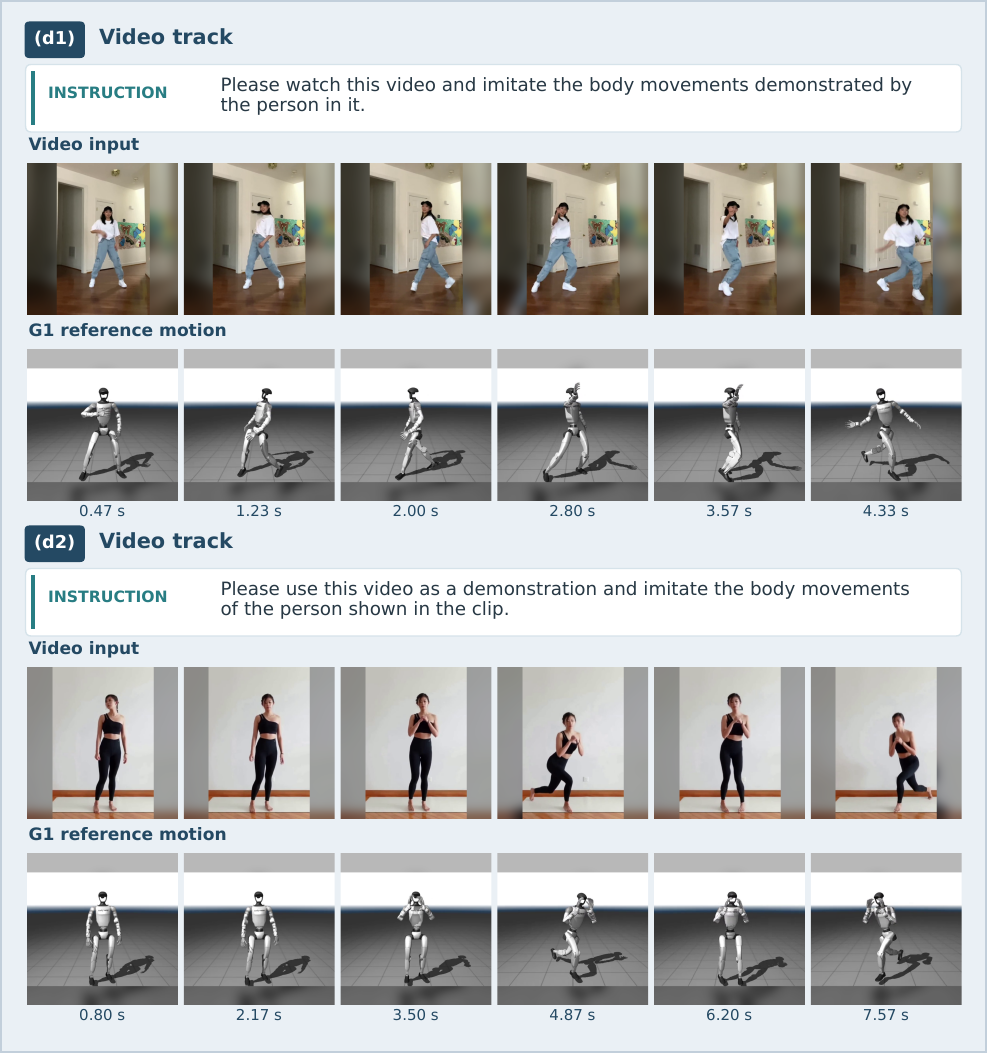}
    \caption[]{Video benchmark examples (continued): (d1) an indoor dance sequence; (d2) a fitness sequence. Each instruction asks the humanoid to imitate the person in the video. Input-video and G1-reference rows share six sampled times within each case.}
    \label{fig:appendix_intent_video}
\end{figure}
\clearpage

\subsection{Semantic, Motion, and Execution Metrics}
\label{app:evaluation_metrics}
Condition-specific metrics measure text--motion retrieval, video-motion reconstruction, and audio--motion alignment.
Shared motion metrics measure smoothness, contact sliding, joint-limit violations, and tracking fidelity.
For \movenszero, Tracking Score is averaged across trajectories, and all rollout-dependent metrics are computed from the corresponding HoloMotion rollouts.

\paragraph{Text and audio benchmark settings.}
The \movensintent results in Tables~\ref{tab:text_results}, \ref{tab:music_results}, and~\ref{tab:speech_results} use 2,048 condition--motion pairs per track, with tracking metrics computed using the canonical HoloMotion single-policy rollout.

\paragraph{Video reconstruction.}
For Table~\ref{tab:video_results}, position errors are in mm, RTE in percent, Foot-Sliding in mm/frame, Jitter in $10\,\mathrm{m/s^3}$, and Accel in $\mathrm{m/s^2}$.
The \movensintent joint and sole-point metrics use G1 forward kinematics.

\paragraph{Dataset quality comparison.}
\label{app:dataset_quality_comparison}
Table~\ref{tab:dataset_motion_quality} groups datasets by their predominant motion acquisition source: MoCap-based AMASS~\citep{mahmood2019amass}, LAFAN1~\citep{harvey2020robust}, BONES-SEED~\citep{bones2026seed}, OmniRetarget~\citep{yang2025omniretarget}, and PHUMA~\citep{lee2025phuma}; and video-based Motion-X++~\citep{zhang2025motionxpp}, Humanoid-X~\citep{mao2024learning}, OMG-Data~\citep{huang2026omg}, and \movensdata.
The scale columns describe the complete datasets or published releases, whereas quality measurements use the sampled populations below.
\# Humanoid Sequences counts published robot-motion trajectories; human-only releases retain a dash.
AMASS, OmniRetarget, and \movensdata frame counts are 30\,Hz equivalents, as specified in Table~\ref{tab:dataset_comparison}.
The eight reference-motion metrics cover temporal smoothness, contact behavior, and joint feasibility; all seven execution metrics come from a common SONIC controller rollout and are displayed in Table~\ref{tab:dataset_motion_quality}.

Table~\ref{tab:dataset_motion_quality} reports an independent seeded sample of at most 1,000 source sequences per dataset, using every available valid sequence when a dataset has fewer than 1,000 candidates.
Each accepted record contributes its complete source trajectory after resampling to 30\,Hz, with no temporal crop or two-second eligibility gate, in a common 29-joint G1 convention.
Unreadable files, malformed representations, and failed conversions are technical rejections; sampling continues along each dataset's seeded candidate order until its target or its available valid population is exhausted.
No motion-quality score, contact label, scene label, or controller result enters selection.
Evaluation uses dataset-specific cohorts, with per-metric sample counts reported in the coverage table.
We evaluate a paired subset of \movensdata containing human motions and retargeted references.
Action categories are not matched across datasets.
Table~\ref{tab:dataset_quality_coverage} reports the resulting coverage; per-window scores, unavailable-field reasons, source identities, crop locations, and hashes accompany the result manifest.

\begin{table}[!htbp]
\centering
\scriptsize
\setlength{\tabcolsep}{4pt}
\caption{Coverage of the independent complete-sequence cohorts. Sampled counts accepted sequences after technical rejection; Sources counts distinct selected recordings. Ref. and Flat count valid G1 sequences on the common floor, Slide counts sequences with measurable support-contact intervals, Contact counts sequences with at least one G1-kinematic sole-contact frame in the shared $[-0.02,0.03]$\,m band, and Exec. counts completed SONIC rollouts including failures. Non-Skating interval coverage is supplied in the result manifest.}
\label{tab:dataset_quality_coverage}
\begin{tabular}{@{}lrrrrrrr@{}}
\toprule
Dataset & Sampled & Sources & Ref. & Flat & Slide & Contact & Exec. \\
\midrule
BONES-SEED & 1000 & 1000 & 1000 & 1000 & 988 & 988 & 1000 \\
OmniRetarget & 1000 & 1000 & 1000 & 1000 & 928 & 938 & 1000 \\
PHUMA & 1000 & 1000 & 1000 & 1000 & 1000 & 1000 & 1000 \\
OMG-Data & 1000 & 1000 & 1000 & 1000 & 297 & 325 & 1000 \\
AMASS & 1000 & 1000 & 1000 & 1000 & 997 & 999 & 1000 \\
LAFAN1 & 77 & 77 & 77 & 77 & 61 & 63 & 77 \\
Motion-X++ & 1000 & 1000 & 1000 & 1000 & 899 & 908 & 1000 \\
Humanoid-X & 1000 & 1000 & 1000 & 1000 & 928 & 972 & 1000 \\
\textbf{\movensdata} & 1000 & 1000 & 1000 & 1000 & 1000 & 1000 & 1000 \\
\bottomrule
\end{tabular}
\end{table}

\textbf{Source preparation.}
Released G1 references undergo unit, quaternion, and joint-order conversion and temporal resampling, without additional smoothing, joint clipping, or ground adjustment.
BONES-SEED uses the original 120\,Hz CSVs restricted to non-mirrored records with no props in the release metadata.
PHUMA uses the 73,076-motion Hub G1 archive, and OMG-Data uses its released state trajectories.
AMASS, Motion-X++, and the publicly released human-pose subset of Humanoid-X are retargeted with GMR~\citep{araujoRetargetingMattersGeneral2025}, using its SMPL-X-to-Unitree-G1 configuration.
The selected source identities, crops, shape parameters, genders, and original scene labels are retained as provenance.
For source sequence $i$, let $N_i$ denote the number of frames after resampling the complete sequence to 30\,Hz.
Translations are linearly interpolated and the first 22 local body rotations are sampled by SLERP at times $t_j=j/30$ seconds, $j=0,\ldots,N_i-1$, through the same SMPL-X body frontend.
This parameter adaptation does not refit SMPL or SMPL-H surfaces to SMPL-X.
Each source sequence starts from the default GMR initialization and processes all $N_i$ frames chronologically, without additional temporal filtering or post-IK joint clipping.
Preceding source-motion history is not replayed.
Following the native GMR SMPL batch-export convention, a constant vertical translation places the minimum FK body-origin height over the complete sequence at zero, and the first-frame root horizontal position is subtracted.
This body-origin convention is not collision-surface calibration.
For \movensdata, the paired human trajectories are retargeted with UMR using its default bidirectional warm start and LQR refinement. After quaternion hemisphere alignment, the root translation and quaternion components are smoothed once with the local three-frame weights $[0.125,0.75,0.125]$, followed by quaternion renormalization. A time-aligned \texttt{smpl2g1} solution then supplies a fixed 0.75 linear prior for the waist and arm joint channels, while the UMR root and 12 leg joints remain unchanged. The blend strength is selected once on a motion-stratified 12-sequence pilot and is not selected per sequence. Finally, all joints are projected inside 97.5\% of the canonical G1 position bounds and 98\% of the adjacent-frame velocity bounds. Refinement is applied to the paired references before SONIC execution.
LAFAN1 uses the native GMR \texttt{bvh\_lafan1}\,$\rightarrow$\,\texttt{unitree\_g1} path with its BVH frame time and a 1.75\,m human-height convention; BONES-SEED, OmniRetarget, PHUMA, and OMG-Data use released G1 references.
These rows therefore measure derived G1 trajectories together with the effects of the stated preparation and initialization conventions.
Contact events are inferred from the prepared G1 reference trajectories using the height-based rule defined below.

Humanoid-X human poses use the published 20\,Hz rate~\citep{mao2024learning}; selected source tracks retain one identity and all consecutive frame indices in the complete track.
Their camera-to-world recovery remains unverified: the measurements concern the canonicalized G1 simulation trajectories, with the pipeline's ground offsets retained in provenance.
Motion-X++ v7 parameter files omit acquisition FPS.
Its row therefore uses 20\,Hz as an explicit playback-time assumption from the official motion-generation renderer at commit \texttt{2a7db28}, rather than a verified capture rate.\footnote{\url{https://github.com/IDEA-Research/Motion-X/blob/2a7db28fc624af736c3eaaa8c10ff375ed69991e/tomato_represenation/plot_3d_global.py}}
Time derivatives and execution difficulty for this row are conditional on that assumption.

\textbf{Reference metrics.}
All datasets use the same 30 G1 body markers.
Jerk is the mean norm of the third finite difference of world positions divided by the cubed frame interval.
MSNR adapts HUMOTO's three-tap velocity SNR~\citep{lu2025humoto} to root-local G1 body positions, discarding convolution boundaries. A higher value means less high-frequency residual relative to the smoothed velocity signal at comparable motion content. Because low-bandwidth or over-smoothed motion can also increase MSNR, it must be interpreted jointly with jerk and the task content; static or noiseless windows are undefined.
C-Slide averages horizontal sole-proxy speed over contacting foot intervals, using the shared height band $[-0.02,0.03]$\,m.
Windows with no such intervals are undefined.
PFC follows the OMG/EDGE implementation~\citep{huang2026omg,tseng2023edge}, combining root acceleration normalized over each complete source sequence with foot displacement; the reported score includes its $10^4$ scaling at the fixed 30\,Hz rate.
Joint Feasibility is the fraction of valid adjacent-frame steps whose 29 joint positions and velocities all satisfy 98\% of the G1 mechanical limits~\citep{lee2025phuma}.

Non-Floating, Non-Penetration, and Non-Skating use PHUMA's 1\,cm, 1\,cm, and 0.1\,m/s robot-side tolerances~\citep{lee2025phuma}.
For Non-Penetration, let $\mathcal{C}_i$ contain the valid G1-kinematic contact frame--region pairs in window $i$, over the left/right heel and toe regions.
Let $d_{t,r}$ be the minimum signed support clearance of the G1 collision proxies in region $r$, including proxy radii.
The window score is
\[
\mathrm{NonPen}_i = 100\,\frac{\sum_{(t,r)\in\mathcal{C}_i}\mathbf{1}[d_{t,r}\geq -0.01\,\mathrm{m}]}{|\mathcal{C}_i|}.
\]
The table averages these scores equally over windows with nonempty $\mathcal{C}_i$.
This measures foot/support penetration, rather than whole-body self-collision; a floating foot can pass this test and is assessed separately by Non-Floating.
Their conditioning events are derived from the G1 trajectories themselves: every cohort is already a canonical G1 sequence on the common $z{=}0$ floor, so a frame--region pair enters $\mathcal{C}_i$ when the region's lowest sole-proxy point lies in the shared $[-0.02,0.03]$\,m height band used by C-Slide.
Contact metrics are computed from G1 trajectories using consistent criteria across datasets.
The 5\,cm conditioning band is deliberately wider than the 1\,cm success tolerance, so the test is not self-fulfilling.
Rates average over contacting foot-region frames or adjacent intervals within each window; windows with no evaluable contact events are undefined, and the per-metric denominators are reported in Table~\ref{tab:dataset_quality_coverage}.

\textbf{Scenes and execution.}
Every reference and controller rollout uses the same horizontal $z=0$ floor, including motions whose original metadata mentions objects, stairs, furniture, or terrain.
Original scene fields remain in provenance but do not exclude a window or add geometry to simulation.
Accordingly, the comparison measures scene-agnostic G1 motion quality and flat-floor executability; it does not reproduce the task success or physical support conditions of interaction motions.
C-Slide and the three G1-kinematic contact rates use the same flat support for every dataset.

Each eligible reference is executed once with the same released SONIC encoder and decoder~\citep{luo2025sonic}, using its G1 motion-conditioning mode.
A synchronous ONNXRuntime adapter preserves the released observation layout, joint mappings, action scales, PD gains, and motor torque limits, with 50\,Hz control and four 0.005\,s MuJoCo substeps.
The native G1 model includes hands; the seven metrics score the same 29 body joints and common body-position kinematics across datasets.
The adapter uses the released hand-control defaults and disables fall-triggered resets, elastic-band support, and outcome-based retries.
No object or terrain geometry is added for any SONIC rollout.
Every window begins at its first reference pose with zero velocity; the ten-step state histories repeat that initial state, and previous actions are zero.
This initialization is held fixed even for crops starting during an ongoing action; it replaces the interactive deployment warmup with a common evaluation reset.
The synchronous runner pairs each reference with the state after its control step, producing one executed state for every complete reference frame, with a 0.02\,s control offset and no temporal or spatial alignment optimization.
MPJPE and g-MPJPE measure root-relative and world-position errors against the window's own reference.
Velocity and acceleration errors are converted from mm/frame and mm/frame$^2$ by factors 0.05 and 2.5 to $\mathrm{m/s}$ and $\mathrm{m/s^2}$, respectively.
A fall is recorded when root height is below 0.20\,m, tilt exceeds $85^\circ$, or height is below 0.35\,m with tilt above $60^\circ$, for at least 11 consecutive control frames.
J-Limit records any executed joint angle beyond its mechanical limit by more than $10^{-4}$\,rad.
Tracking Score aggregates the 29 joint-angle errors over all rollout frames:
\begin{equation}
q=0.35e^{-e_{\mathrm{rms}}/0.12}+0.20e^{-e_{\mathrm{mae}}/0.10}
 +0.30e^{-e_{95}/0.25}+0.15e^{-e_{\max}/0.50},
\label{eq:dataset_quality_tracking_score}
\end{equation}
where $e_{\mathrm{rms}}$, $e_{\mathrm{mae}}$, $e_{95}$, and $e_{\max}$ are the root-mean-square, mean absolute, 95th-percentile absolute, and maximum absolute errors in radians, respectively.

Dataset entries are arithmetic macro-averages of the defined per-sequence values, with coverage reported for each metric.
At most 1,000 complete sequences per dataset contribute, including falls and joint-limit violations; a cohort with fewer than 1,000 available valid sequences contributes all of them.
Every complete sequence is rolled out once from its first reference pose, so the tracking results describe the full available duration rather than a fixed short horizon.
The differing source populations, action mixtures, representation conversions, and contact-label availability prevent interpreting this sampled comparison as a controlled ranking of entire datasets.

\subsection{Scaling Protocol}
\label{app:scaling_protocol}
Data scaling uses nested $1\%$, $3\%$, $10\%$, $30\%$, and $100\%$ subsets sampled before temporal slicing.
All data-scaling runs use the XL action expert (910.5M parameters).
Model scaling compares the S, M, L, and XL action experts in Table~\ref{tab:model_scale_architecture} on the full training set while preserving the data pipeline and evaluation protocol.

\clearpage

\section{Additional Qualitative Results}
\label{app:qualitative_results}
\suppressfloats[t]

\subsection{Additional Offline Comparisons}
\label{app:offline_qualitative}

The fourteen examples below provide the full qualitative comparison set summarized in Figure~\ref{fig:qualitative_overview}: three video cases, three speech cases, five text cases, and three music cases. Each figure shows six frames in temporal order, with ground-truth motion (or reference motion for music), \movenszero, and the modality-specific baselines in separate rows. The spoken instructions are printed as text for readability; the music panels include the supplied audio waveform. Numbered amber callouts identify selected visible discrepancies, with repeated numbers indicating the same discrepancy within each case; they are not measured errors. These offline comparisons are distinct from the hardware rollouts in Appendix~\ref{app:real_robot_cases}.

\begin{figure}[tp]
\centering
\includegraphics[width=\linewidth,height=0.86\textheight,keepaspectratio]{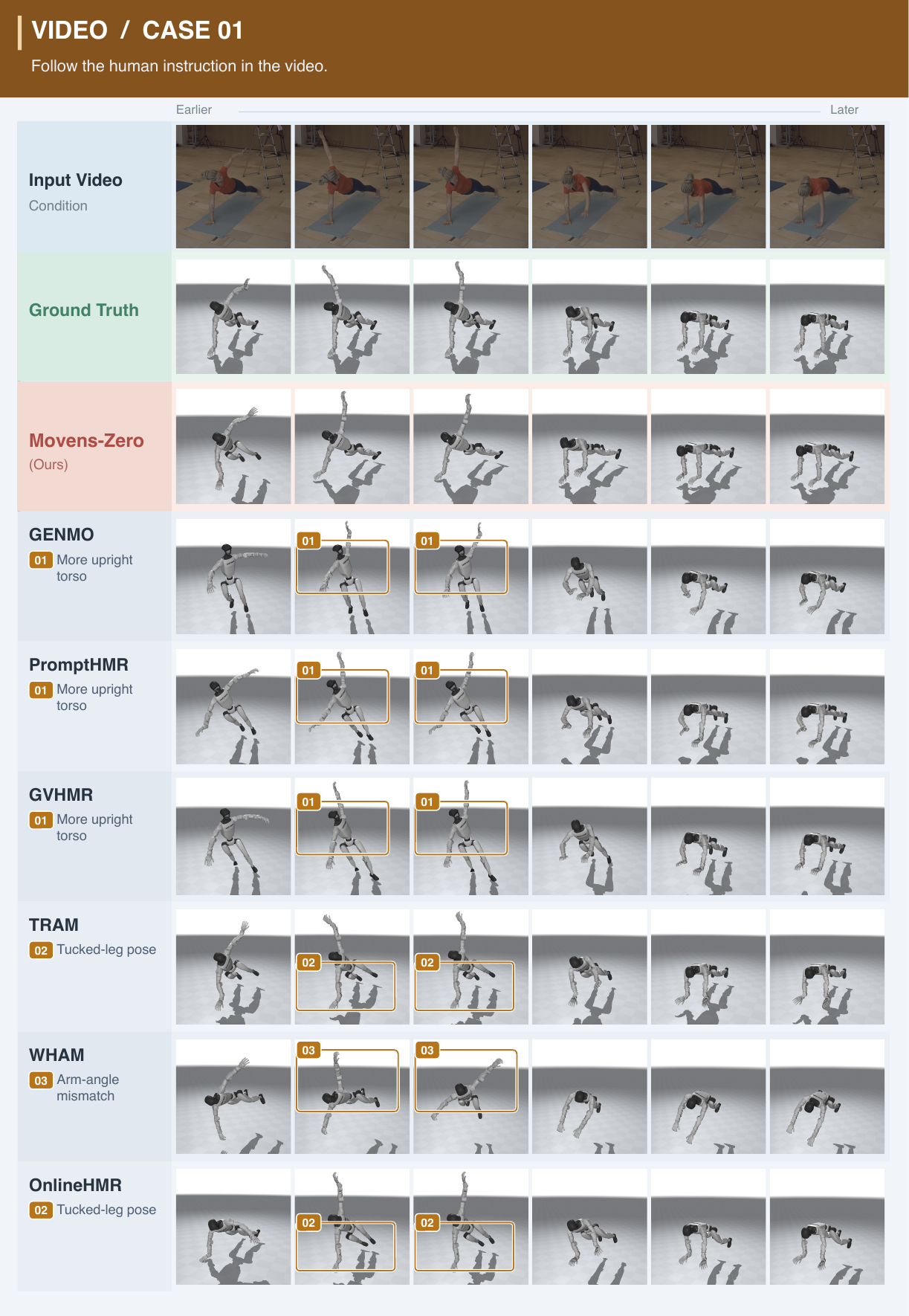}
\caption{Video Case 01: raising one arm from a low supported posture before transitioning to forward support.}
\label{fig:qualitative_main}
\end{figure}

\begin{figure}[tp]
\centering
\includegraphics[width=\linewidth,height=0.86\textheight,keepaspectratio]{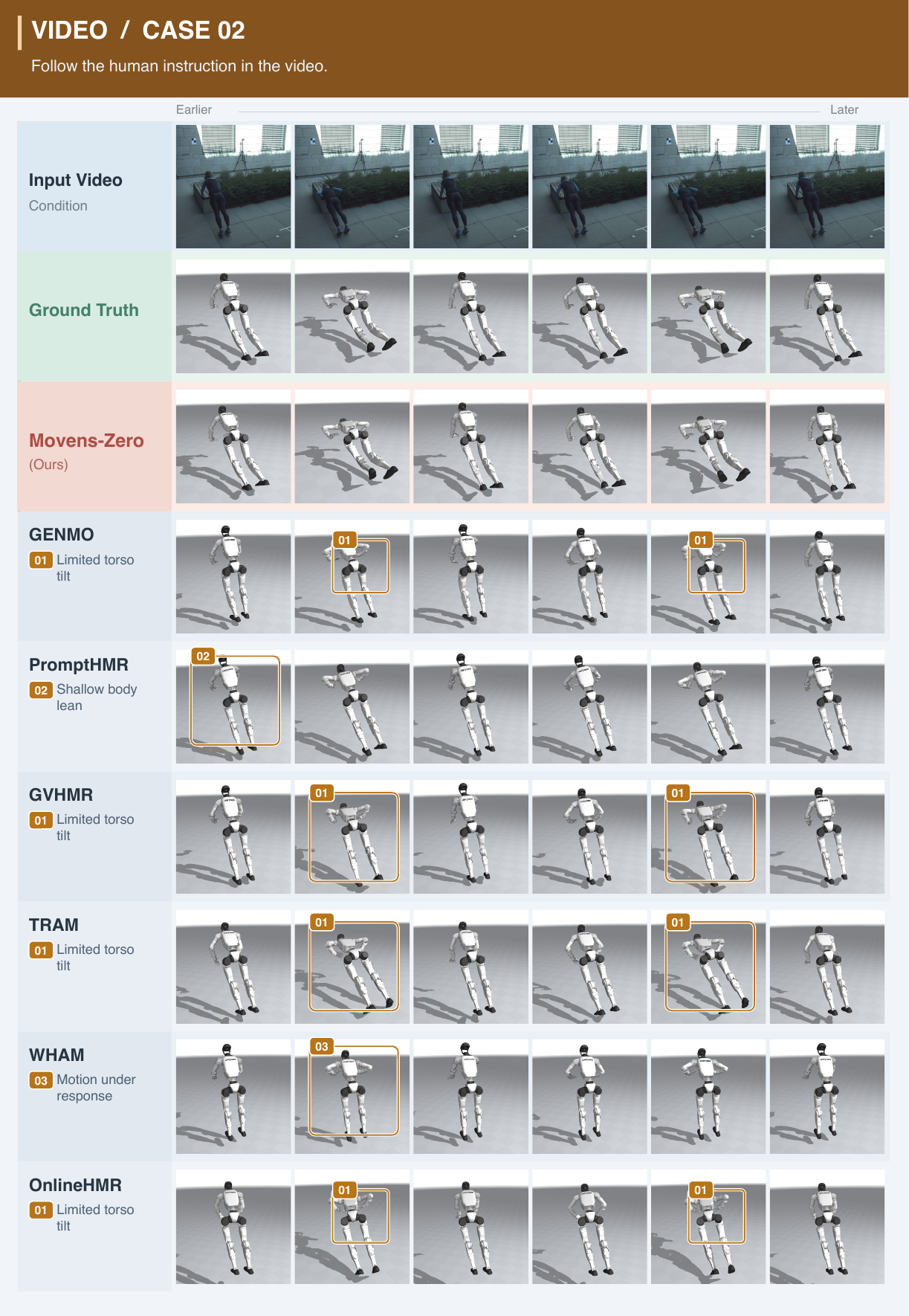}
\caption{Video Case 02: repeated changes in whole-body inclination.}
\label{fig:qualitative_appendix_video_2}
\end{figure}

\begin{figure}[tp]
\centering
\includegraphics[width=\linewidth,height=0.86\textheight,keepaspectratio]{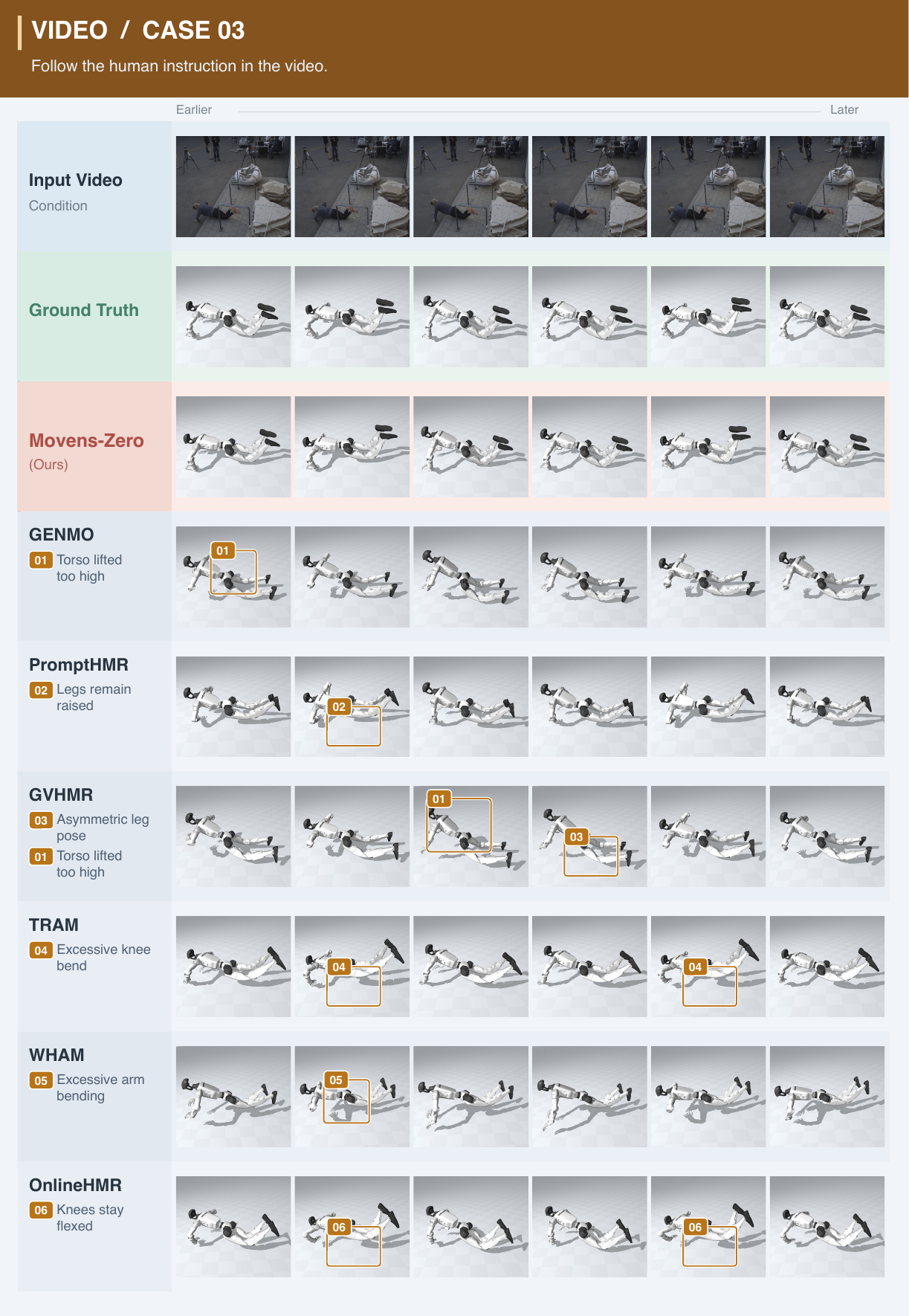}
\caption{Video Case 03: floor exercise with changing support and leg configurations.}
\label{fig:qualitative_appendix_video_3}
\end{figure}

\paragraph{Video-conditioned comparisons.}
\textbf{Case 01 (Figure~\ref{fig:qualitative_main}).} The demonstrated motion raises one arm before transitioning to forward support. During the arm-raising phase, GENMO, PromptHMR, and GVHMR adopt a more upright torso than the reference, while TRAM and OnlineHMR show a more tucked lower-body configuration. WHAM differs in the raised-arm angle at the third sampled moment. \movenszero preserves the low torso posture, the outward lower-body configuration, and the subsequent transition to two-arm support, retaining the main spatial relationships of the demonstrated whole-body motion. \textbf{Case 02 (Figure~\ref{fig:qualitative_appendix_video_2}).} The reference alternates between a shallower lean and a pronounced forward inclination, most clearly at the second and fifth sampled moments. GENMO, GVHMR, TRAM, and OnlineHMR remain more upright at these moments, while PromptHMR also shows a shallower whole-body lean. WHAM exhibits a smaller change in body configuration across the sampled sequence. \movenszero captures the stronger leaning postures and their recurrence, preserving the coordinated changes of the torso, arms, and lower body rather than reducing the demonstrated motion to a nearly upright posture. \textbf{Case 03 (Figure~\ref{fig:qualitative_appendix_video_3}).} GENMO raises the torso higher than the reference in several sampled poses, and GVHMR shows a similar deviation at the third moment together with an asymmetric leg configuration at the fourth. WHAM has more pronounced elbow bending at the second moment, while TRAM and OnlineHMR differ in knee flexion and foot height at the second and fifth moments. \movenszero retains the low supported torso and the corresponding bent-leg configurations, reproducing the main upper- and lower-body changes shown by the reference.

\begin{figure}[tp]
\centering
\includegraphics[width=\linewidth,height=0.86\textheight,keepaspectratio]{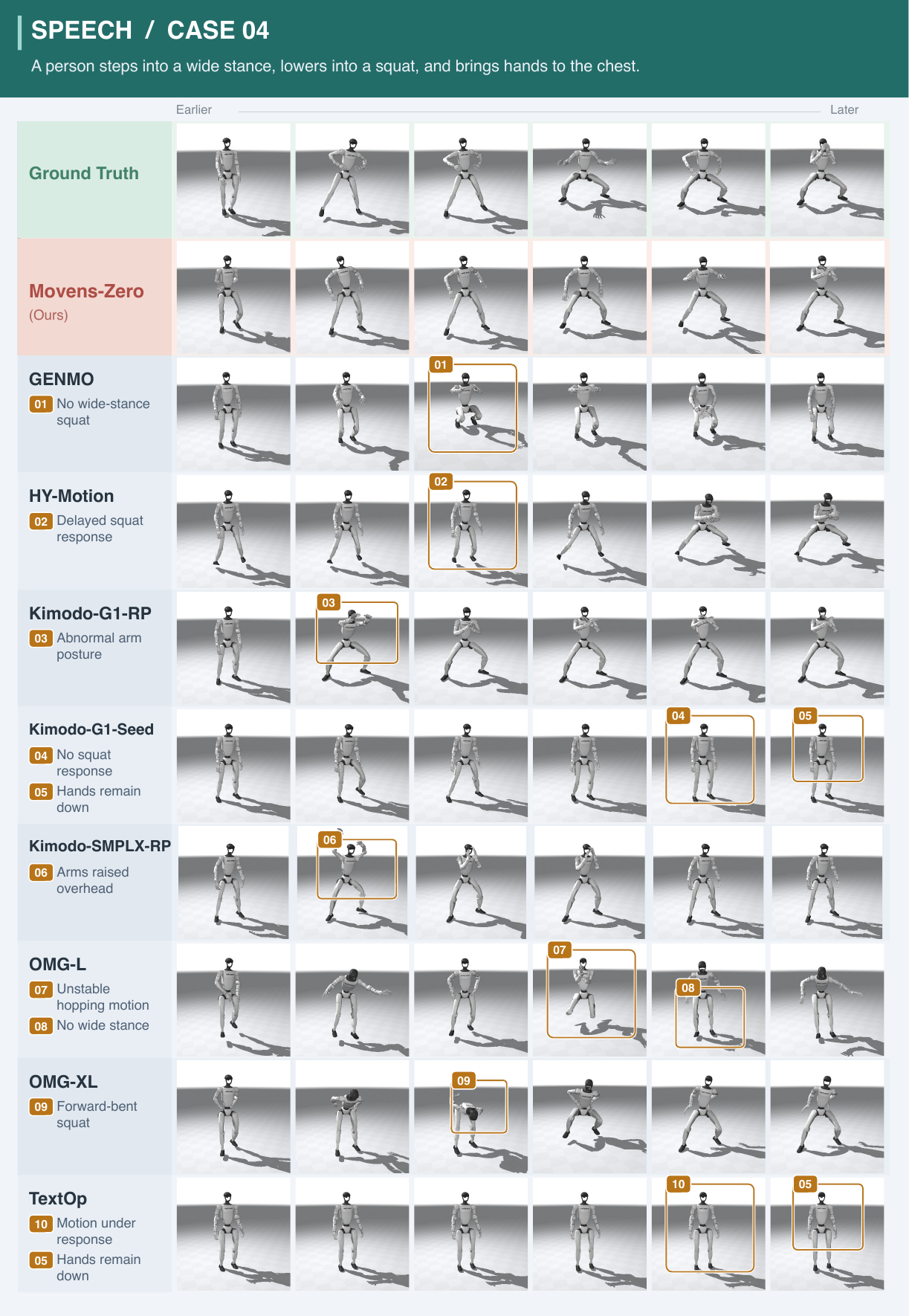}
\caption{Speech Case 04: stepping into a wide stance, lowering into a squat, and bringing the hands to the chest.}
\label{fig:qualitative_speech}
\end{figure}

\begin{figure}[tp]
\centering
\includegraphics[width=\linewidth,height=0.86\textheight,keepaspectratio]{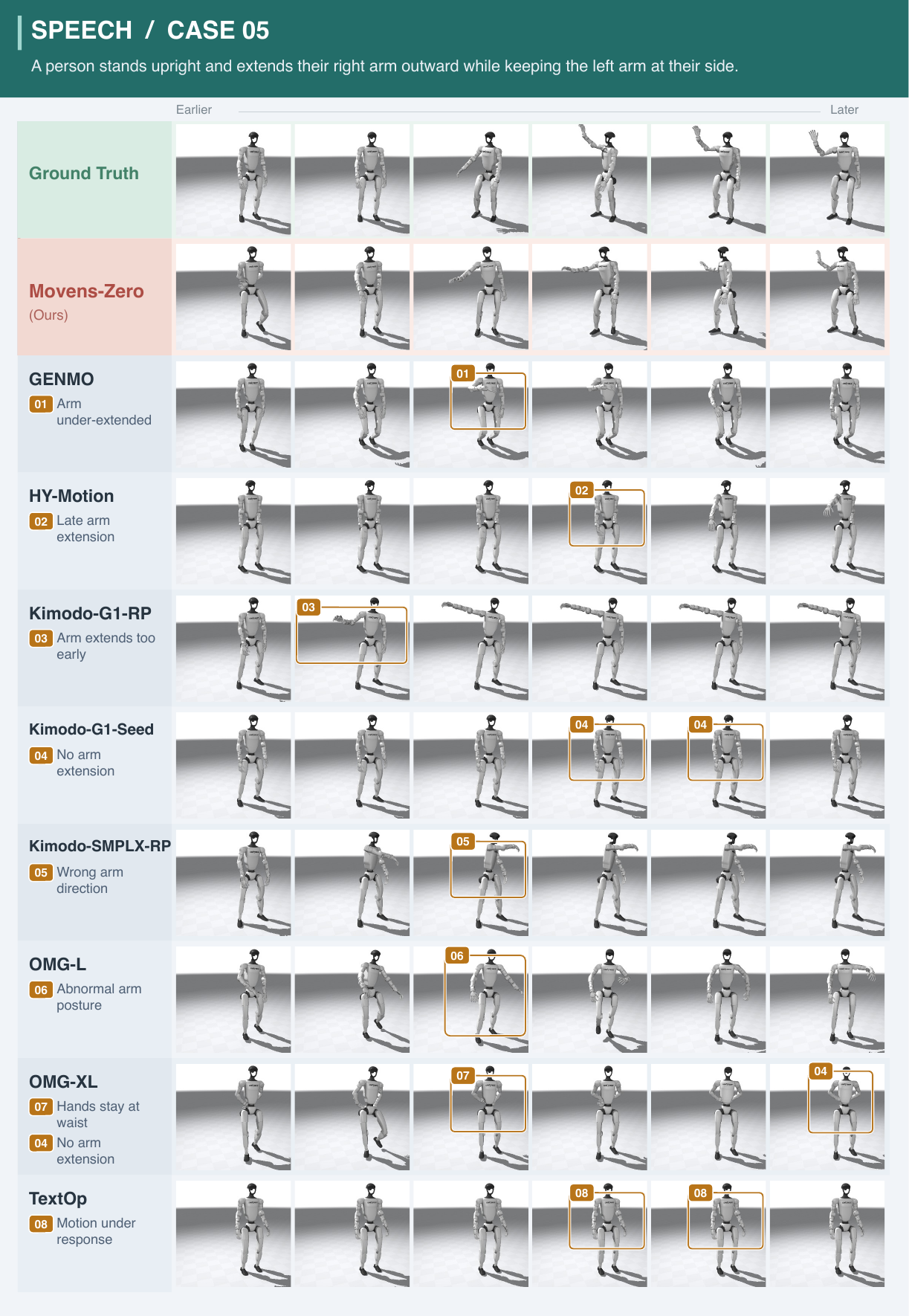}
\caption{Speech Case 05: extending the right arm outward while the left arm remains lowered.}
\label{fig:qualitative_appendix_speech_2}
\end{figure}

\begin{figure}[tp]
\centering
\includegraphics[width=\linewidth,height=0.86\textheight,keepaspectratio]{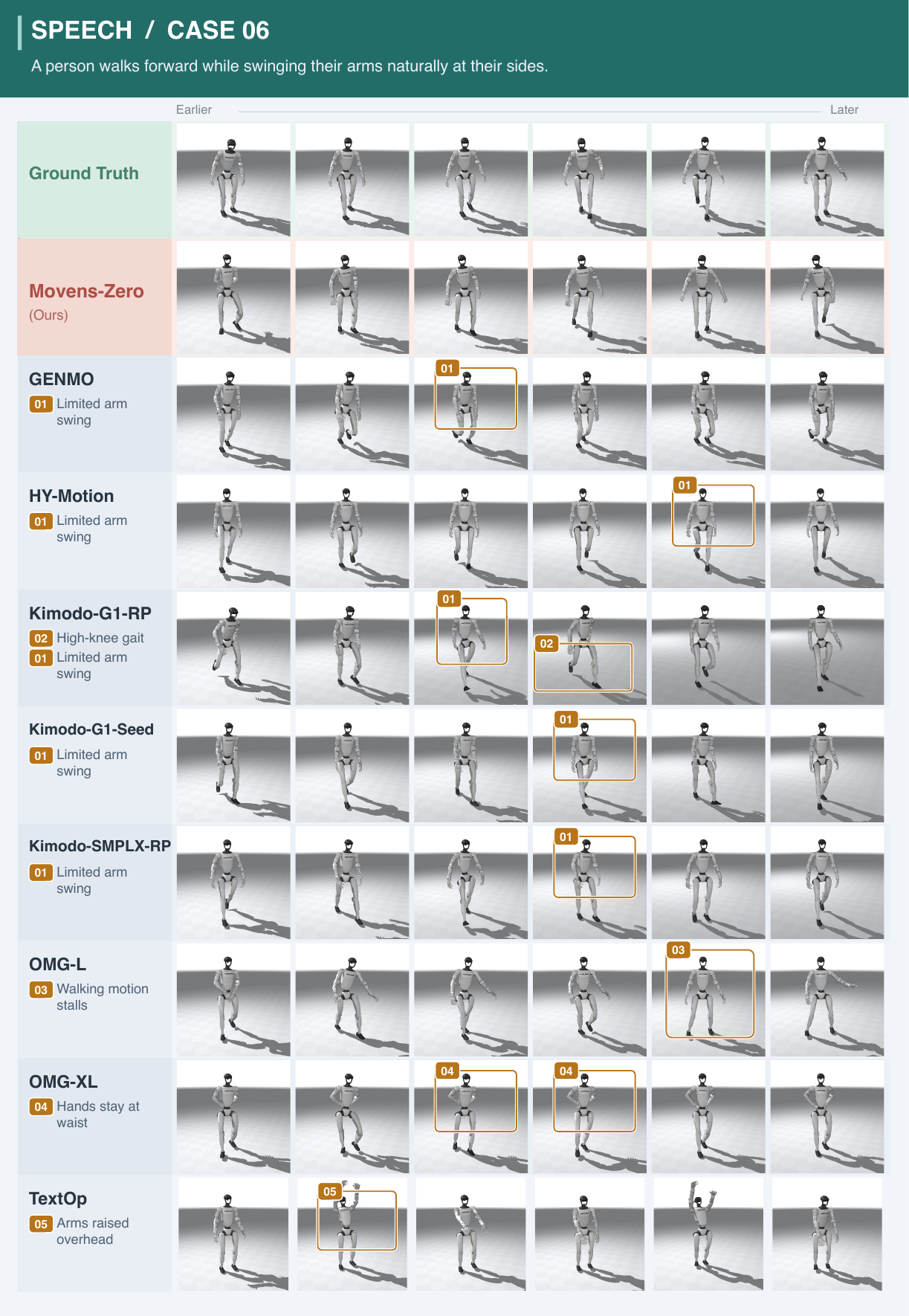}
\caption{Speech Case 06: walking forward with the arms swinging naturally at the sides.}
\label{fig:qualitative_appendix_speech_3}
\end{figure}

\paragraph{Speech-conditioned comparisons.}
\textbf{Case 04 (Figure~\ref{fig:qualitative_speech}).} GENMO bends the knees without reproducing the wide stance, whereas Kimodo-G1-Seed and TextOp remain relatively upright with the hands lowered in the later samples. HY-Motion reaches a pronounced squat at a later sampled phase than the reference, while Kimodo-SMPLX-RP raises the arms overhead early in the sequence. OMG-L maintains a narrower stance, and OMG-XL introduces a stronger forward bend. \movenszero combines the wide stance, knee flexion, and hands-to-chest pose, more fully preserving the action components of the spoken instruction. \textbf{Case 05 (Figure~\ref{fig:qualitative_appendix_speech_2}).} GENMO produces a smaller arm extension, and HY-Motion shows the extension only in the later displayed phase. Kimodo-G1-Seed and TextOp remain close to a neutral standing pose in the sampled sequence; OMG-XL keeps the hands near the waist rather than extending the right arm. \movenszero raises and extends the instructed arm while leaving the opposite arm at the side, preserving both the commanded movement and the asymmetric upper-body configuration specified by the spoken instruction. \textbf{Case 06 (Figure~\ref{fig:qualitative_appendix_speech_3}).} GENMO, HY-Motion, and the Kimodo variants show relatively small changes in arm position across the displayed steps, while Kimodo-G1-RP also uses more pronounced high-knee poses. OMG-XL keeps both hands near the waist, and TextOp raises both arms overhead in some samples instead of maintaining the requested arm placement. \movenszero displays alternating leg configurations together with changes in arm position at the sides, preserving the coordinated upper- and lower-body pattern required by the walking instruction.

\begin{figure}[tp]
\centering
\includegraphics[width=\linewidth,height=0.86\textheight,keepaspectratio]{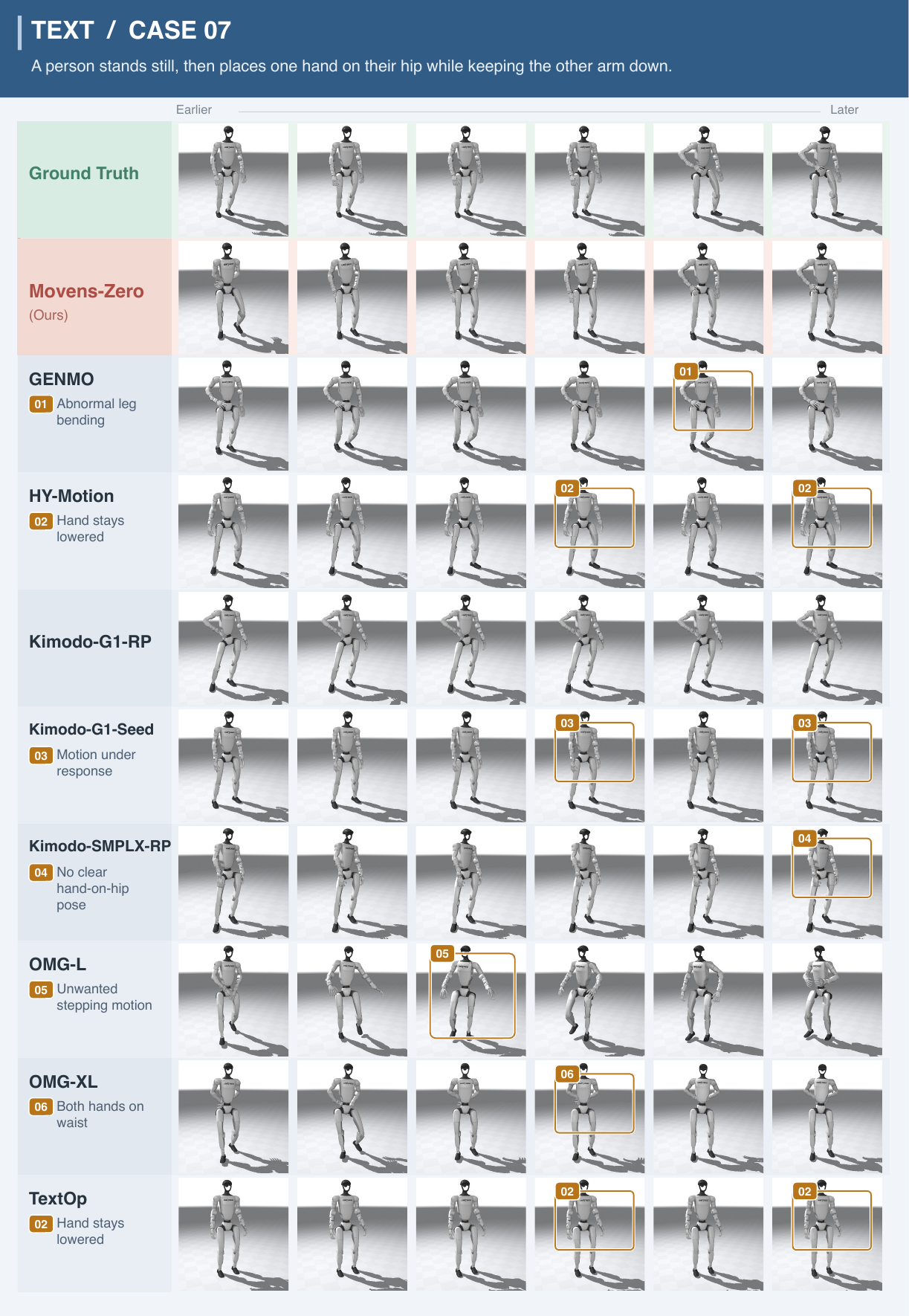}
\caption{Text Case 07: placing one hand on the hip while the other arm remains down.}
\label{fig:qualitative_appendix_text_1}
\end{figure}

\begin{figure}[tp]
\centering
\includegraphics[width=\linewidth,height=0.86\textheight,keepaspectratio]{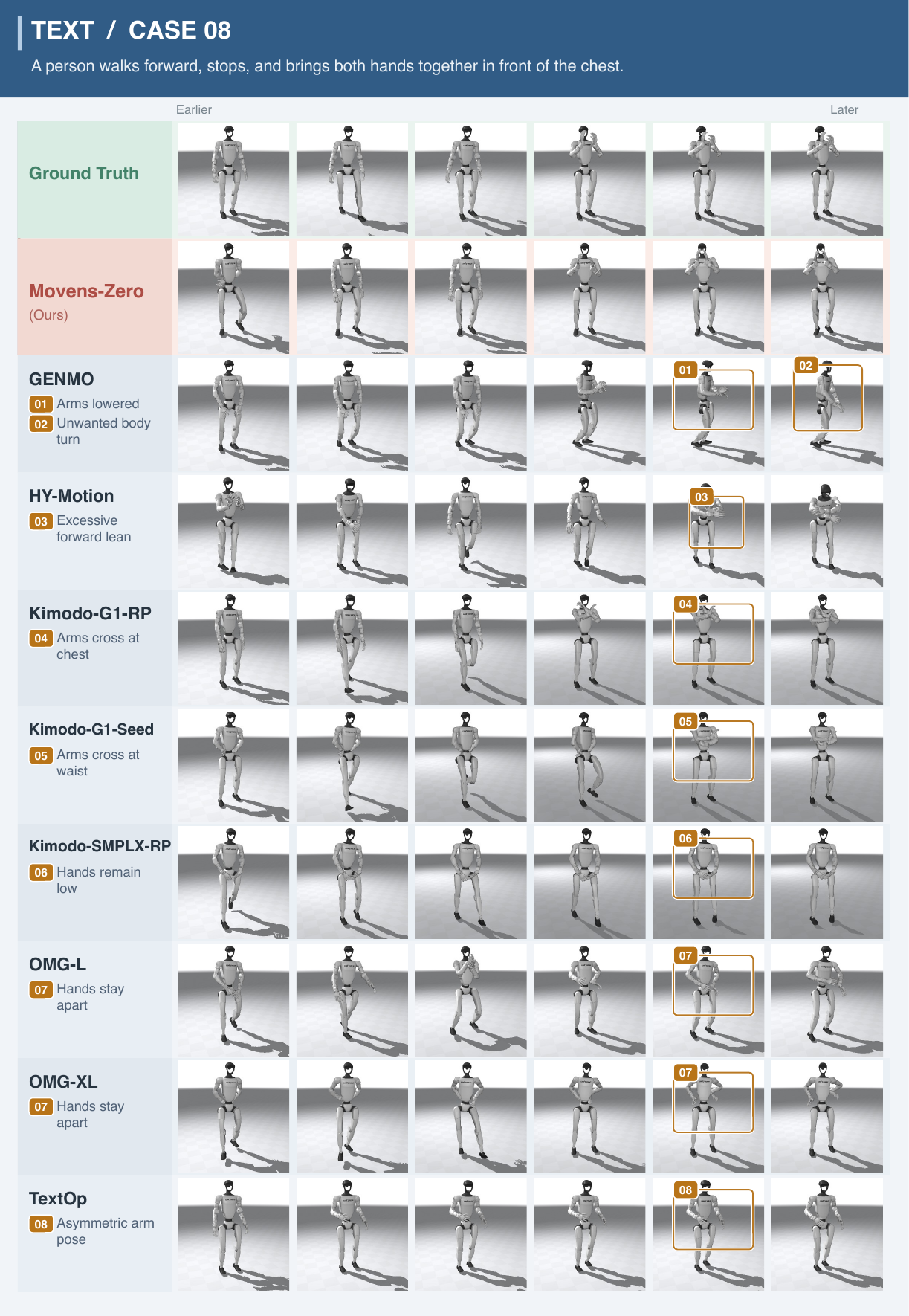}
\caption{Text Case 08: walking forward, stopping, and bringing both hands together in front of the chest.}
\label{fig:qualitative_text}
\end{figure}

\begin{figure}[tp]
\centering
\includegraphics[width=\linewidth,height=0.86\textheight,keepaspectratio]{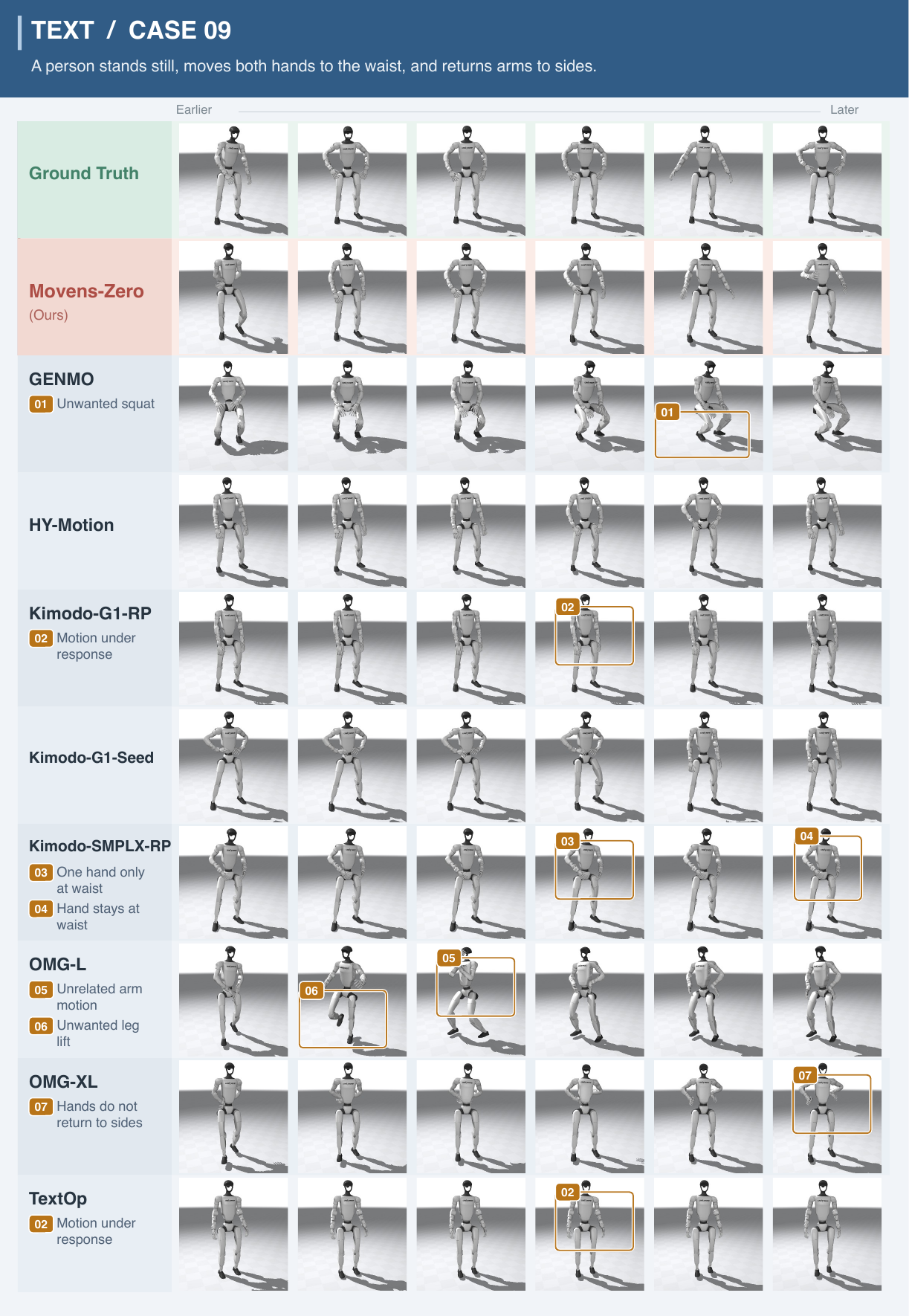}
\caption{Text Case 09: moving both hands to the waist and returning the arms to the sides.}
\label{fig:qualitative_appendix_text_3}
\end{figure}

\begin{figure}[tp]
\centering
\includegraphics[width=\linewidth,height=0.86\textheight,keepaspectratio]{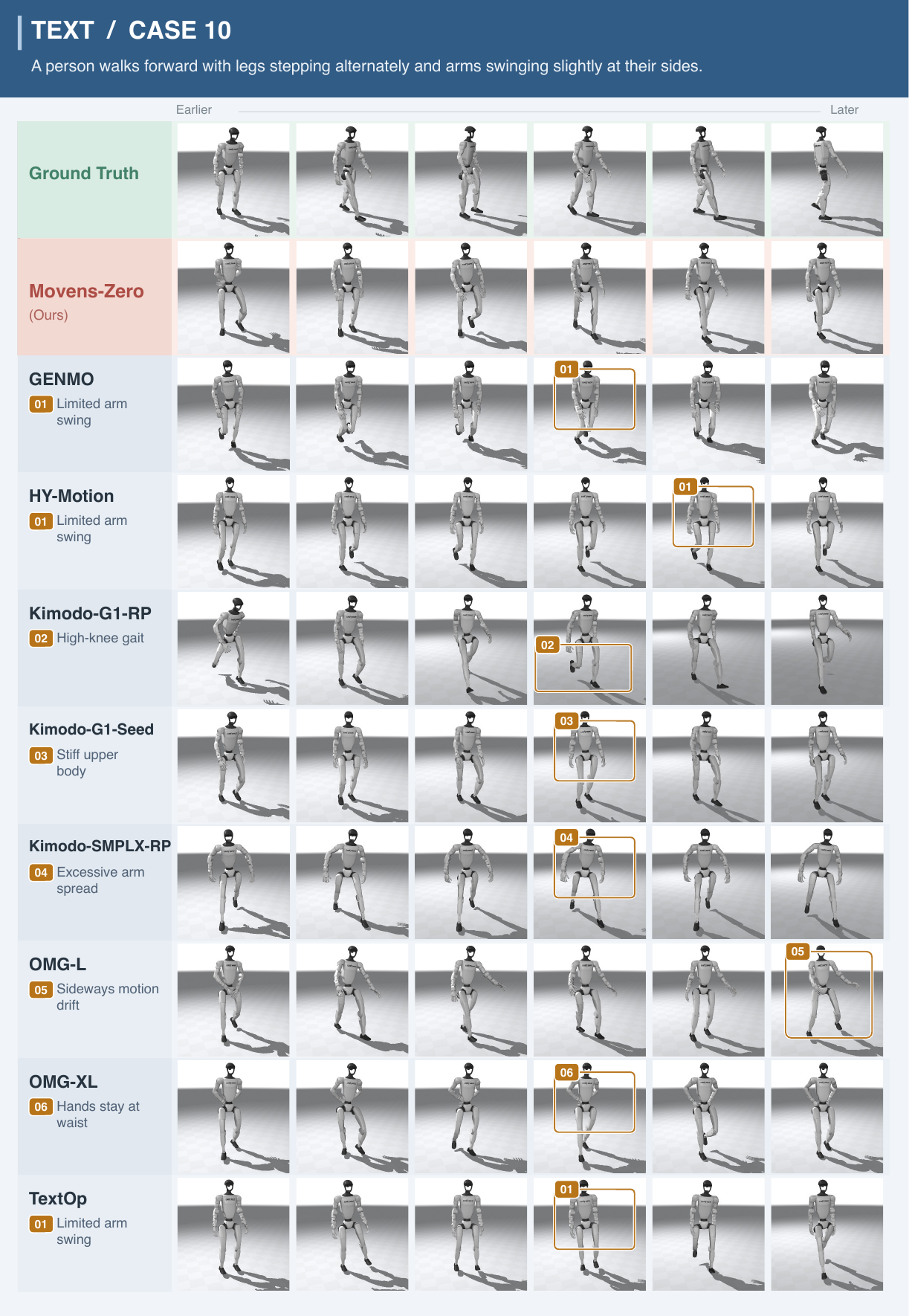}
\caption{Text Case 10: walking forward with alternating steps and slight arm swing.}
\label{fig:qualitative_appendix_text_4}
\end{figure}

\begin{figure}[tp]
\centering
\includegraphics[width=\linewidth,height=0.86\textheight,keepaspectratio]{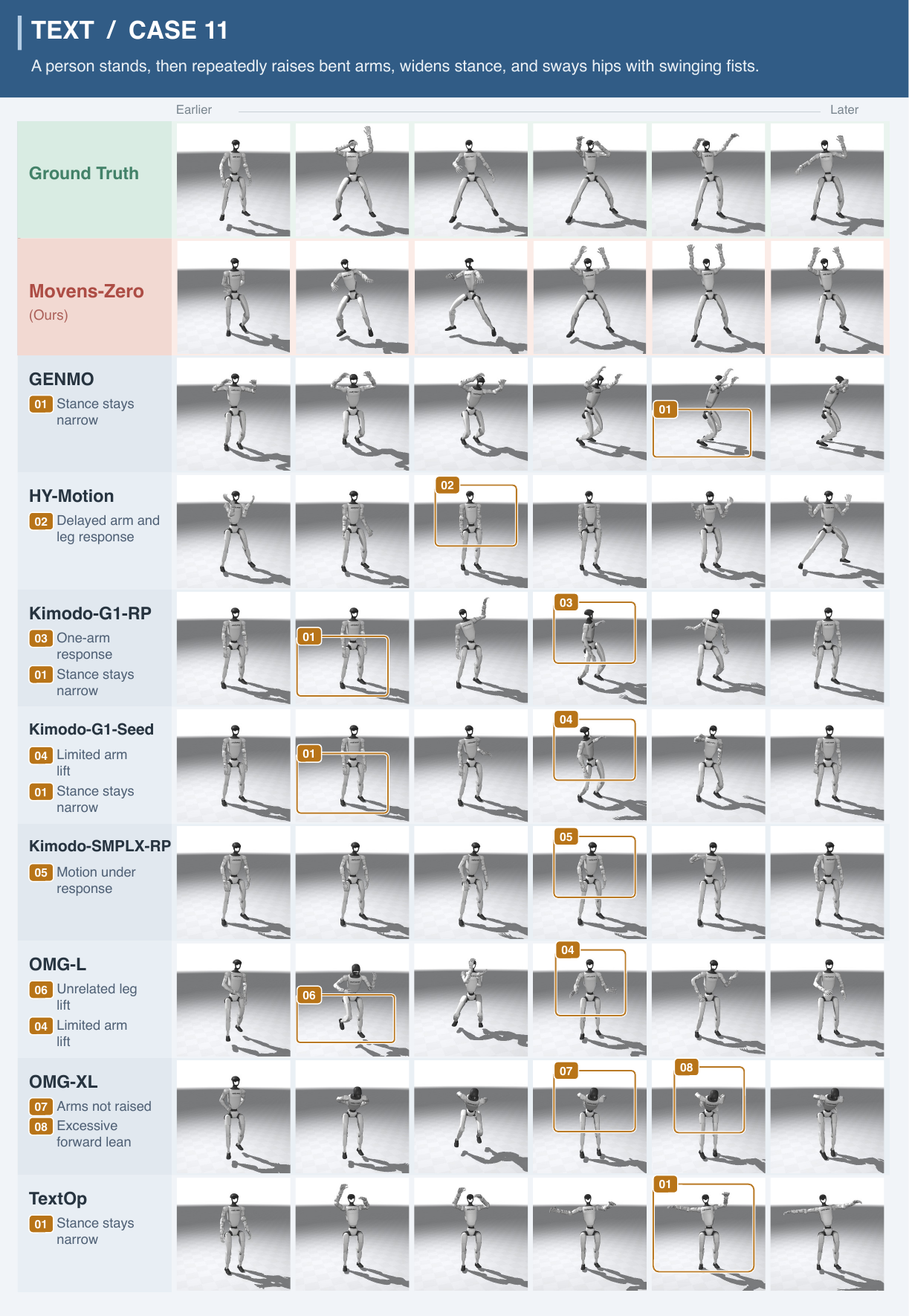}
\caption{Text Case 11: raising bent arms, widening the stance, and swaying the hips while swinging the fists.}
\label{fig:qualitative_appendix_text_5}
\end{figure}

\paragraph{Text-conditioned comparisons.}
\textbf{Case 07 (Figure~\ref{fig:qualitative_appendix_text_1}).} \movenszero places one hand on the hip while keeping the opposite arm lowered in the later samples. HY-Motion and TextOp keep the hands lowered, while Kimodo-G1-Seed remains close to the initial arms-down configuration. OMG-XL instead places both hands at the waist, and OMG-L introduces additional leg lifting. These comparisons highlight the ability of \movenszero to preserve both the commanded arm action and the constraint on the opposite arm, rather than producing only a broadly similar standing pose. \textbf{Case 08 (Figure~\ref{fig:qualitative_text}).} \movenszero retains the transition from alternating stepping poses to a bilateral hand gesture in front of the upper body. In the later samples, Kimodo-G1-Seed and Kimodo-SMPLX-RP keep the hands near the waist, while OMG-L and OMG-XL leave them separated. Kimodo-G1-RP crosses the forearms across the chest rather than clearly bringing the hands together, and HY-Motion introduces pronounced forward trunk flexion. \movenszero better preserves the instructed combination of locomotion and upper-body action. \textbf{Case 09 (Figure~\ref{fig:qualitative_appendix_text_3}).} \movenszero displays both waist-level hand placement and subsequent arm lowering while retaining an upright body configuration. GENMO introduces pronounced knee flexion and a lowered body posture, while Kimodo-G1-RP and TextOp largely retain an arms-down configuration. Kimodo-SMPLX-RP predominantly places one hand at the waist and retains that placement in the last sample; OMG-XL keeps both hands near the waist in the displayed later frames. \movenszero preserves the two distinct upper-body configurations requested by the instruction. \textbf{Case 10 (Figure~\ref{fig:qualitative_appendix_text_4}).} \movenszero shows alternating leg configurations with the arms remaining beside the torso, preserving the requested combination of lower-body movement and upper-body placement. OMG-XL holds the hands at the waist across the samples rather than adopting the requested arm-at-side configuration. Kimodo-SMPLX-RP extends the arms farther from the body, while Kimodo-G1-RP uses more pronounced high-knee poses. These comparisons show how \movenszero preserves the coordination of leg motion and arm placement specified by the text, beyond generating a recognizable walking pattern. \textbf{Case 11 (Figure~\ref{fig:qualitative_appendix_text_5}).} \movenszero displays coordinated changes in arm configuration and stance, including raised arms over a clearly widened base of support. GENMO retains a narrower stance in the front-facing samples. Kimodo-G1-RP primarily raises one arm, while Kimodo-G1-Seed shows less arm elevation; both also retain relatively narrow leg configurations. OMG-XL instead bends the trunk forward while keeping the arms low. Together, the displayed poses show how \movenszero combines arm elevation with stance widening, rather than expressing only one component of the instruction.

\begin{figure}[tp]
\centering
\includegraphics[width=\linewidth,height=0.86\textheight,keepaspectratio]{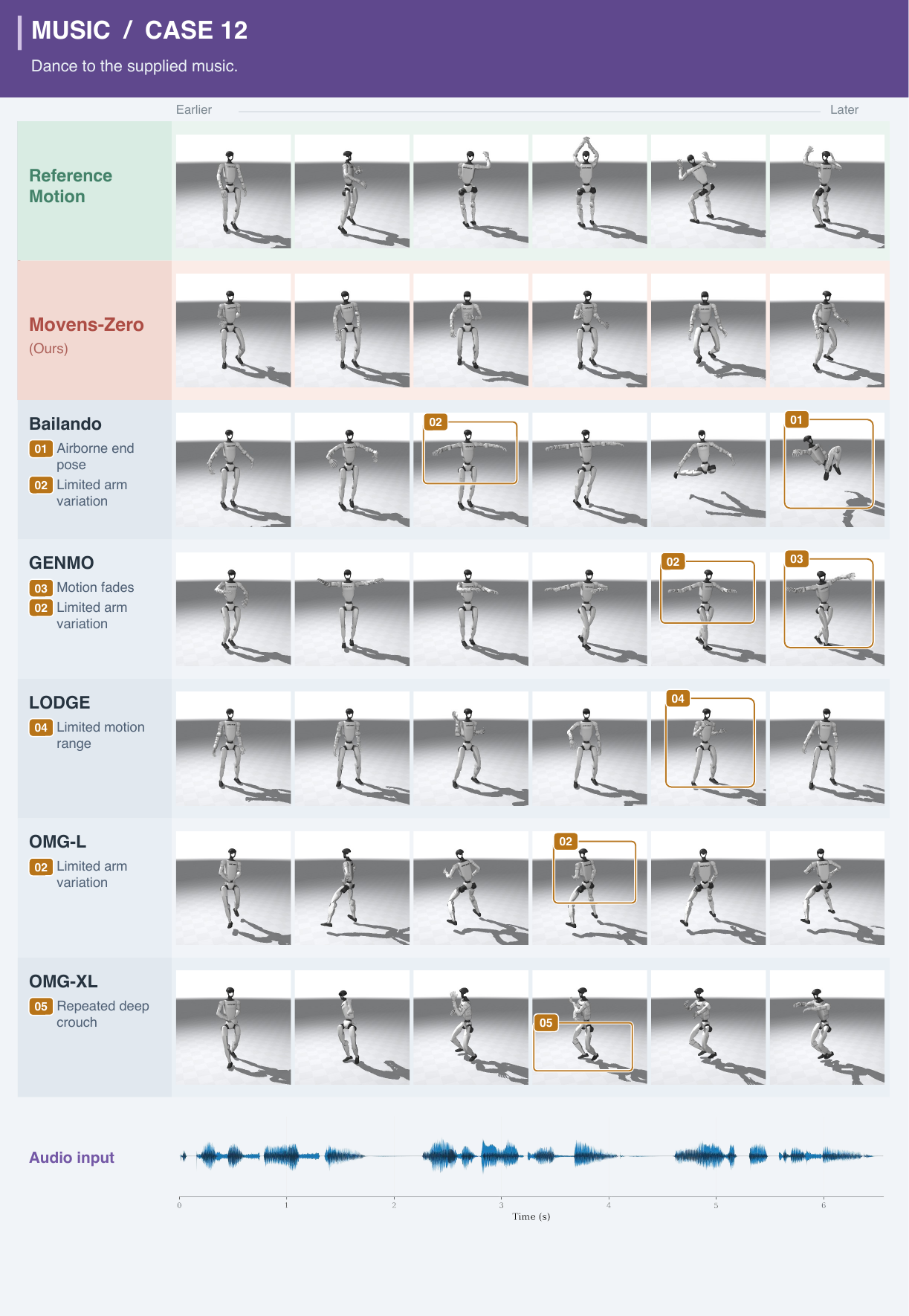}
\caption{Music Case 12: dance with changes in stance, arm placement, and body orientation.}
\label{fig:qualitative_appendix_music_1}
\end{figure}

\begin{figure}[tp]
\centering
\includegraphics[width=\linewidth,height=0.86\textheight,keepaspectratio]{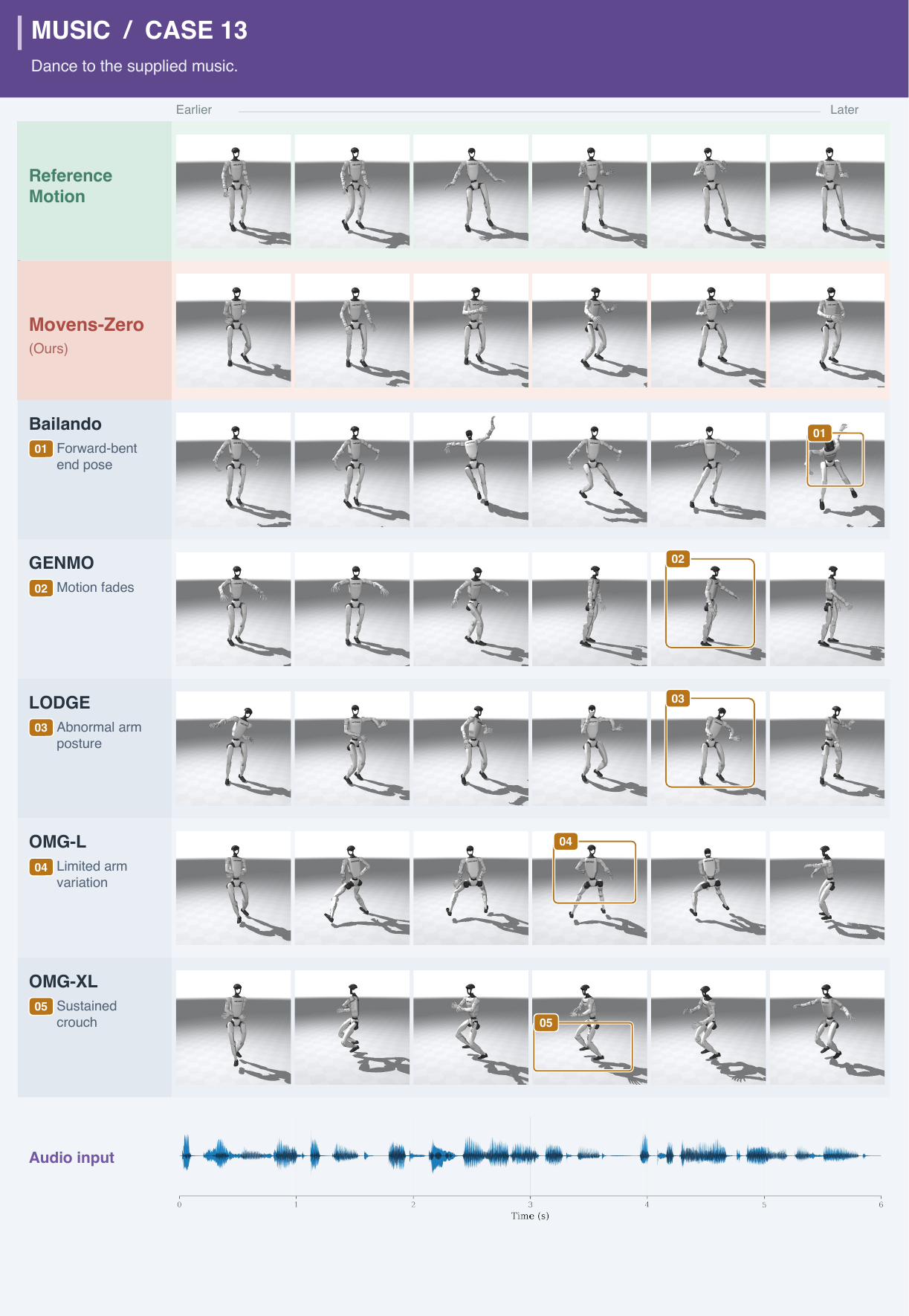}
\caption{Music Case 13: dance with changing arm gestures and lower-body stance.}
\label{fig:qualitative_music}
\end{figure}

\begin{figure}[tp]
\centering
\includegraphics[width=\linewidth,height=0.86\textheight,keepaspectratio]{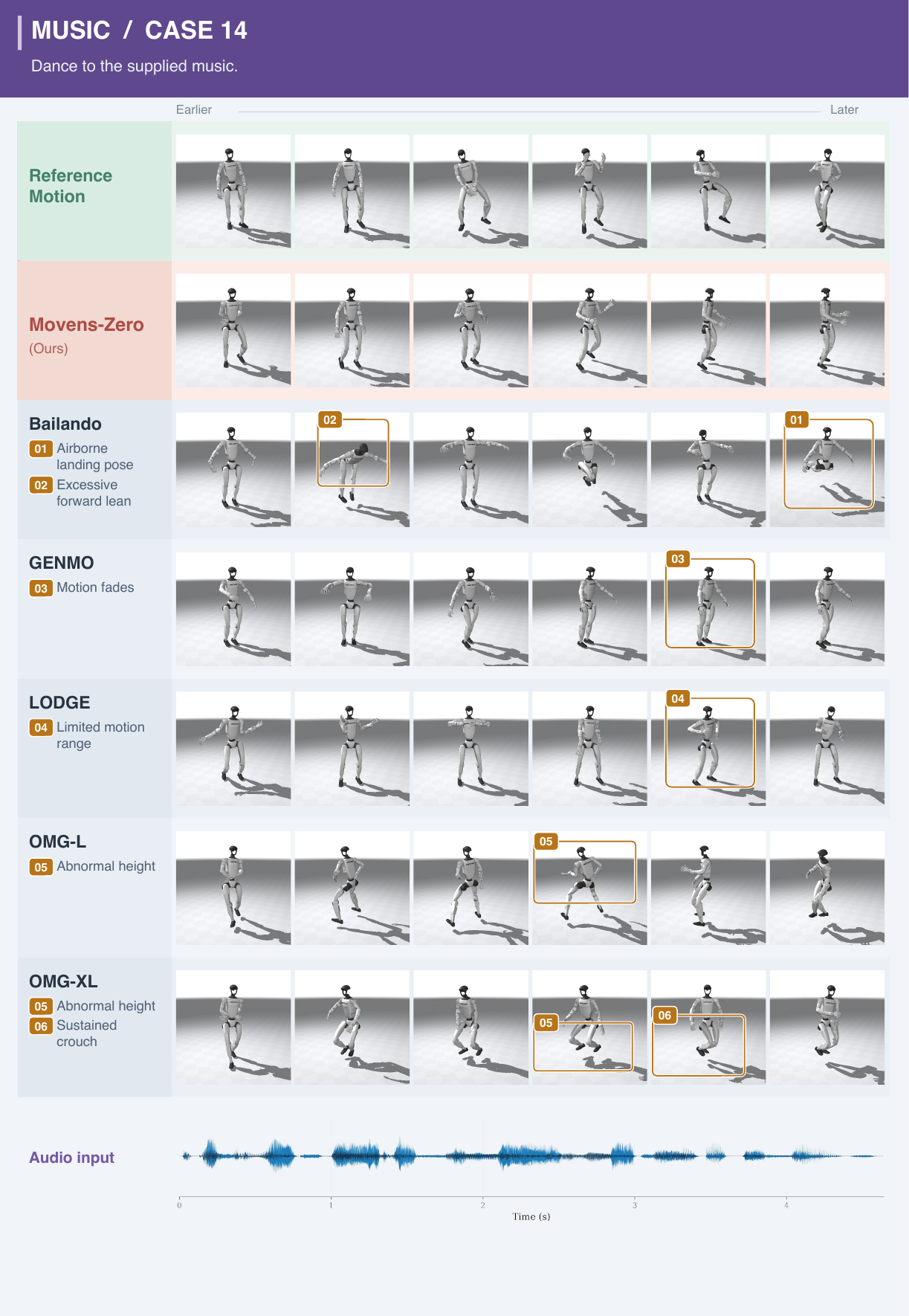}
\caption{Music Case 14: dance with coordinated arm, leg, and facing-direction changes.}
\label{fig:qualitative_appendix_music_3}
\end{figure}

\paragraph{Music-conditioned comparisons.}
\textbf{Case 12 (Figure~\ref{fig:qualitative_appendix_music_1}).} Bailando shows an airborne, strongly tilted configuration in the final samples, whereas GENMO maintains broadly outstretched arms over several moments. LODGE exhibits smaller changes in stance, and OMG-L keeps the arms in a comparatively restricted set of positions. OMG-XL repeatedly adopts a lowered, bent-knee configuration. \movenszero combines alternating steps with changes in torso orientation and arm placement, producing a varied sequence of coordinated whole-body poses in response to the musical input. \textbf{Case 13 (Figure~\ref{fig:qualitative_music}).} Bailando ends in a pronounced forward-bent pose, while GENMO returns to near-upright poses with the arms lowered in the latter samples. OMG-L shows comparatively limited changes in arm placement, and OMG-XL keeps the knees flexed across the later displayed moments. \movenszero continues to vary its stance, arm gestures, and orientation through the later samples, maintaining an articulated upper- and lower-body dance pattern rather than settling into a single dominant posture. \textbf{Case 14 (Figure~\ref{fig:qualitative_appendix_music_3}).} Bailando shows a pronounced forward lean early in the sequence and an airborne configuration in the final sample. GENMO's later poses become more upright with the arms lowered, while LODGE shows relatively small changes in lower-body stance. OMG-XL retains bent-knee configurations across the later samples. \movenszero maintains alternating leg configurations together with upper-body gestures and changes in orientation, sustaining whole-body pose variation across the displayed musical excerpt.

\clearpage

\clearpage

\subsection{Real-Robot Capability Cases}
\label{app:real_robot_cases}

We show all 30 Unitree G1 hardware rollouts from the project page, with six cases for each input type.
Each figure uses a two-column by three-row grid of capability cards.
Every card names the action or reproduces the full instruction, reports the source-video duration, and shows twelve numbered snapshots in a 3-by-4 grid with their audited sample timestamps.
Live-video cards pair an earlier performer crop with a later G1 crop selected at the corresponding action phase, and print the two source times separately.
Their differences are display alignment offsets for qualitative comparison rather than measurements of end-to-end system latency.
Internet-video cards retain a shared composite-video timestamp for the reference and execution, but select clearer key poses instead of ambiguous transition frames.
All evidence lanes use the same snapshot numbers: video and text cards number their timeline markers, speech cards number complete-waveform sample markers, and music cards number detected-onset markers.
For music, each numbered waveform marker is a detected audio onset, and the equally numbered image is the nearest 25-FPS video frame.
The paired times differ by at most 0.02~s.
Only the repeated prompt strip embedded at the bottom of the speech, music, and text videos is cropped.
The displayed inputs and robot frames are only cropped or rescaled; no image content is edited.
These examples illustrate behavioral range rather than estimate a hardware success rate; the complete videos are available on the \href{https://movens-zero.github.io}{project page}.

\paragraph{Live-video-conditioned cases.}
\textbf{Case 01 (Figure~\ref{fig:appendix_real_robot_live_video}a).} The performer alternates forward bows with upright standing. G1 shows corresponding torso lowering and recovery, with knee flexion and the arms held near the body across the repeated transitions. \textbf{Case 02 (Figure~\ref{fig:appendix_real_robot_live_video}b).} Bilateral arm raises are separated by arms-down poses, followed by a single-hand salute. G1 follows this broad ordering and finishes with one hand near its head, moving between symmetric and asymmetric upper-body configurations. \textbf{Case 03 (Figure~\ref{fig:appendix_real_robot_live_video}c).} The sampled sequence covers preparation, a brief airborne phase, and renewed ground contact. In the central samples, both G1 feet are visibly clear of the floor as the arms move with the jump, followed by a standing configuration. \textbf{Case 04 (Figure~\ref{fig:appendix_real_robot_live_video}d).} A one-arm wave precedes an upright pause and a squat. G1's raised-hand configurations give way to increased knee flexion and a lower torso, followed by a return toward standing, preserving the main gesture-to-posture transition. \textbf{Case 05 (Figure~\ref{fig:appendix_real_robot_live_video}e).} Outward arm extension is followed by a side-facing posture and reaches in front of the torso. G1 changes orientation while alternating extended and flexed arms, bringing the demonstrated turning and reaching components into one execution sequence. \textbf{Case 06 (Figure~\ref{fig:appendix_real_robot_live_video}f).} Frontal arm gestures are followed by a turn and walking away in the reference. G1 similarly changes its facing direction and progresses through alternating leg configurations with visible displacement across the floor.

\paragraph{Internet-video-conditioned cases.}
\textbf{Case 01 (Figure~\ref{fig:appendix_real_robot_internet_video}a).} The reference repeatedly brings the arms across the chest, opens the elbows outward, and places the hands near the head. G1 moves among the same broad arm arrangements while remaining in a standing configuration. \textbf{Case 02 (Figure~\ref{fig:appendix_real_robot_internet_video}b).} Lateral dance steps occur with cross-body sweeps, overhead gestures, and a turn in the household reference. G1 varies its support leg and arm placement through the sequence, including a turned orientation in the later samples. \textbf{Case 03 (Figure~\ref{fig:appendix_real_robot_internet_video}c).} Cross-body hand sweeps recur alongside small stepping movements in the reference. G1 alternates arm configurations while changing between flexed-leg support poses, expressing the co-occurrence of the recurring upper-body gesture and the lower-body motion specified by the video. \textbf{Case 04 (Figure~\ref{fig:appendix_real_robot_internet_video}d).} The outdoor reference combines a wide stance and arm sweeps. G1 maintains a separated stance while varying bent-arm positions near the head and in front of the torso, with visible differences from the performer's arm range. \textbf{Case 05 (Figure~\ref{fig:appendix_real_robot_internet_video}e).} Overhead reaches alternate with lower arm positions and steps before a final forward bend in the reference. G1 changes its arm elevation and leg configuration through these phases and finishes with a lowered, inclined torso. \textbf{Case 06 (Figure~\ref{fig:appendix_real_robot_internet_video}f).} Raised-arm gestures lead into a deep crouch, followed by rising, turning, and further changes in hand position. G1 follows the broad ordering of standing, lowered, and turned configurations, although its crouch is shallower than the performer's.

\paragraph{Speech-conditioned cases.}
\textbf{Case 01 (Figure~\ref{fig:appendix_real_robot_speech}a).} G1 moves from arms-down to a broad bilateral arm opening, then brings the forearms inward across the upper torso. The ordered poses make both stages of the spoken instruction visible while the lower body retains a standing configuration. \textbf{Case 02 (Figure~\ref{fig:appendix_real_robot_speech}b).} The sampled execution combines changing foot placement with asymmetric knee bending, while the hands remain near the waist. These poses provide a concrete view of the lower-body exercise elicited by the spoken lunge instruction over an extended clip. \textbf{Case 03 (Figure~\ref{fig:appendix_real_robot_speech}c).} G1 raises the commanded left arm from the side toward the head and retains the bent-elbow salute configuration in the later samples. The opposite arm stays lower, making the asymmetric arm selection in the spoken instruction directly visible. \textbf{Case 04 (Figure~\ref{fig:appendix_real_robot_speech}d).} G1 changes from hands-near-waist poses to an asymmetric raised-arm configuration, then turns the torso before presenting both bent arms beside the head. The sequence illustrates how a broad spoken style request can produce several distinct upper-body configurations within one rollout. \textbf{Case 05 (Figure~\ref{fig:appendix_real_robot_speech}e).} Across the sampled sequence, one hand is raised near the head while the opposite arm remains lower, followed by a switch to the other side. The changing asymmetric configurations make the left-right alternation requested in speech visible in the hardware execution. \textbf{Case 06 (Figure~\ref{fig:appendix_real_robot_speech}f).} G1 keeps the left hand raised beside the head across the sampled clip while the other arm remains lower. The timestamps make the persistence of the selected-arm gesture visible, complementing the earlier speech examples that require a short transition or alternating arm movements.

\paragraph{Music-conditioned cases.}
\textbf{Case 01 (Figure~\ref{fig:appendix_real_robot_music}a).} Alternating leg lifts and sideward foot placements are accompanied by arm movements from a low position to gestures near the chest. The sequence shows variation across both upper and lower body under music conditioning, with the numbered waveform locating each displayed pose. \textbf{Case 02 (Figure~\ref{fig:appendix_real_robot_music}b).} G1 moves through a wide, bent-knee stance while alternating asymmetric arm gestures with bilateral arm lifts. Changes in torso facing accompany the arm movements, providing a concrete example of music-conditioned variation across stance, orientation, and upper-body pose. \textbf{Case 03 (Figure~\ref{fig:appendix_real_robot_music}c).} The sampled poses combine outward arm opening, hands moving toward the head, and later arm lowering with changes in knee flexion. These simultaneous changes in upper- and lower-body configuration illustrate an articulated dance response across several moments of the supplied music. \textbf{Case 04 (Figure~\ref{fig:appendix_real_robot_music}d).} The early samples show both hands near the head, followed by wider arm openings and a change in torso facing. Later poses return to asymmetric arm configurations, making the progression between compact, open, and turned upper-body shapes visible within the music-conditioned rollout. \textbf{Case 05 (Figure~\ref{fig:appendix_real_robot_music}e).} G1 alternates lateral torso inclinations while moving an arm across the chest and changing leg placement. The repeated changes in shoulder angle and stance make the lateral movement pattern visible across the numbered samples, together with the supplied audio waveform. \textbf{Case 06 (Figure~\ref{fig:appendix_real_robot_music}f).} G1 changes from raised-arm poses over shifting leg configurations to a side-facing posture with diagonal reaches and a backward trunk inclination. The ordered samples show variation in facing direction, arm reach, and lower-body posture within one music-conditioned hardware clip.

\paragraph{Text-conditioned cases.}
\textbf{Case 01 (Figure~\ref{fig:appendix_real_robot_text}a).} Under the hip-hop instruction, G1 combines repeated bent-knee steps with alternating arm lifts and changes in torso inclination. The sequence sustains variation across the legs, trunk, and arms over the sampled portion of the clip. \textbf{Case 02 (Figure~\ref{fig:appendix_real_robot_text}b).} The samples show three distinct lowered squat configurations separated by more upright poses. Forward arm motion accompanies the lowering phases, linking the counted instruction to a repeated sequence of whole-body descent and recovery. \textbf{Case 03 (Figure~\ref{fig:appendix_real_robot_text}c).} For the salsa instruction, G1 varies foot placement and alternates its arms around waist and chest height. The snapshots show coordinated changes in leg configuration, torso inclination, and arm position over the sequence. \textbf{Case 04 (Figure~\ref{fig:appendix_real_robot_text}d).} For the counterclockwise-walking request, the samples show changes in knee flexion, arm elevation, and torso inclination. G1 remains near the same floor location and largely faces the camera, so the figure does not establish completion of the requested circular path. \textbf{Case 05 (Figure~\ref{fig:appendix_real_robot_text}e).} G1 walks toward the camera and raises its left arm into a wave during the later part of the clip. The right arm remains lower through the main waving phase, illustrating the combination of locomotion and a side-specific gesture. \textbf{Case 06 (Figure~\ref{fig:appendix_real_robot_text}f).} Under the acting instruction, G1 traverses the view with flexed knees and changing arm positions. Its orientation progresses through frontal, side, and rear views, giving the broad qualitative request a moving whole-body realization.

\clearpage
\begingroup
\makeatletter
\setlength{\@fptop}{0pt}
\setlength{\@fpsep}{8pt}
\setlength{\@fpbot}{0pt plus 1fil}
\setlength{\@dblfptop}{0pt}
\setlength{\@dblfpsep}{8pt}
\setlength{\@dblfpbot}{0pt plus 1fil}
\makeatother

\begin{figure*}[p]
\centering
\includegraphics[width=\linewidth,height=0.92\textheight,keepaspectratio]{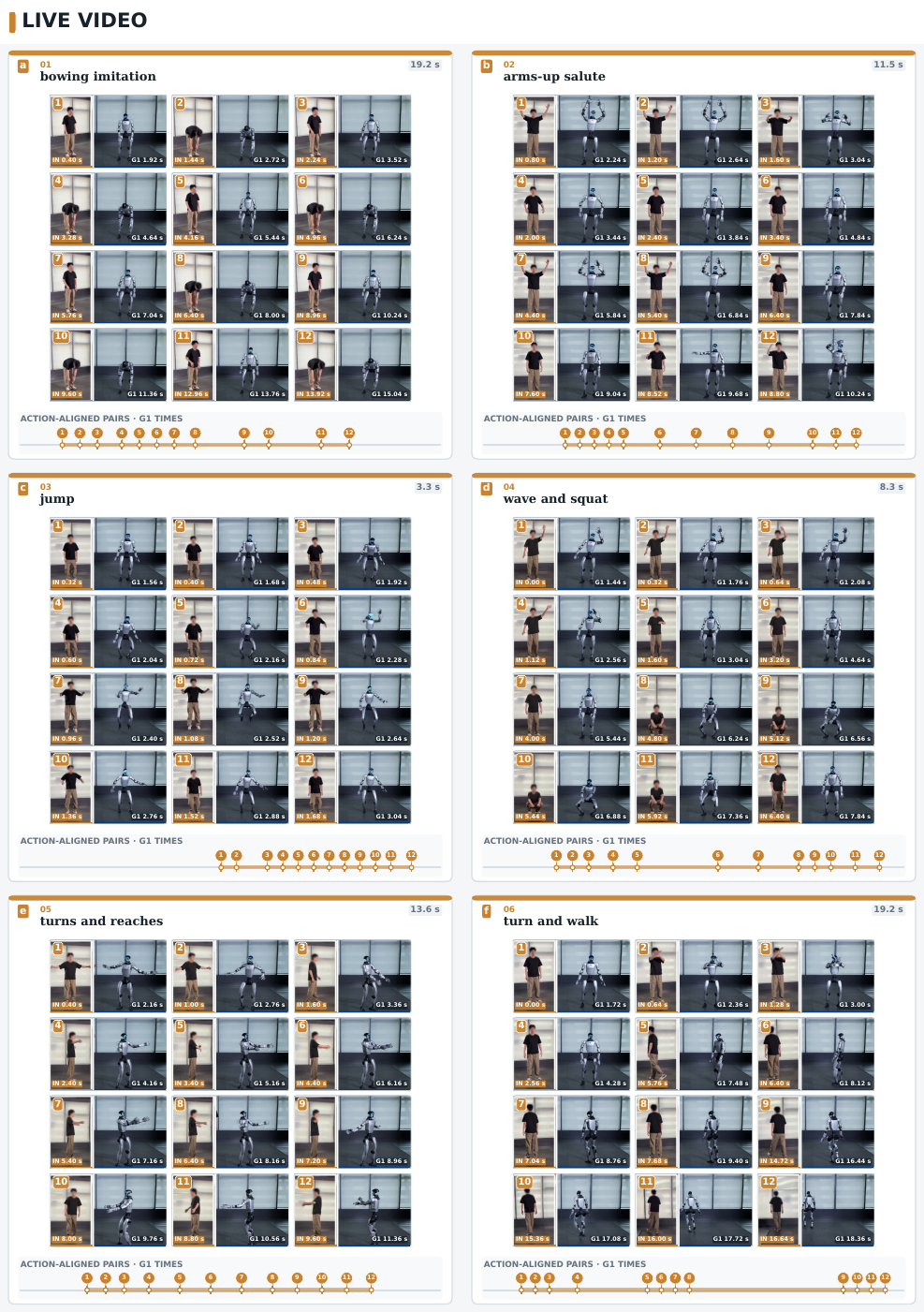}
\caption{Six live-video-conditioned motions executed on Unitree G1. Each capability card shows twelve numbered pairs; every pair combines an earlier performer input with the later G1 response at a corresponding action phase and prints both source times. The time differences are display alignment offsets chosen for qualitative pose comparison, not measurements of end-to-end system latency.}
\label{fig:appendix_real_robot_live_video}
\end{figure*}

\begin{figure*}[p]
\centering
\includegraphics[width=\linewidth,height=0.92\textheight,keepaspectratio]{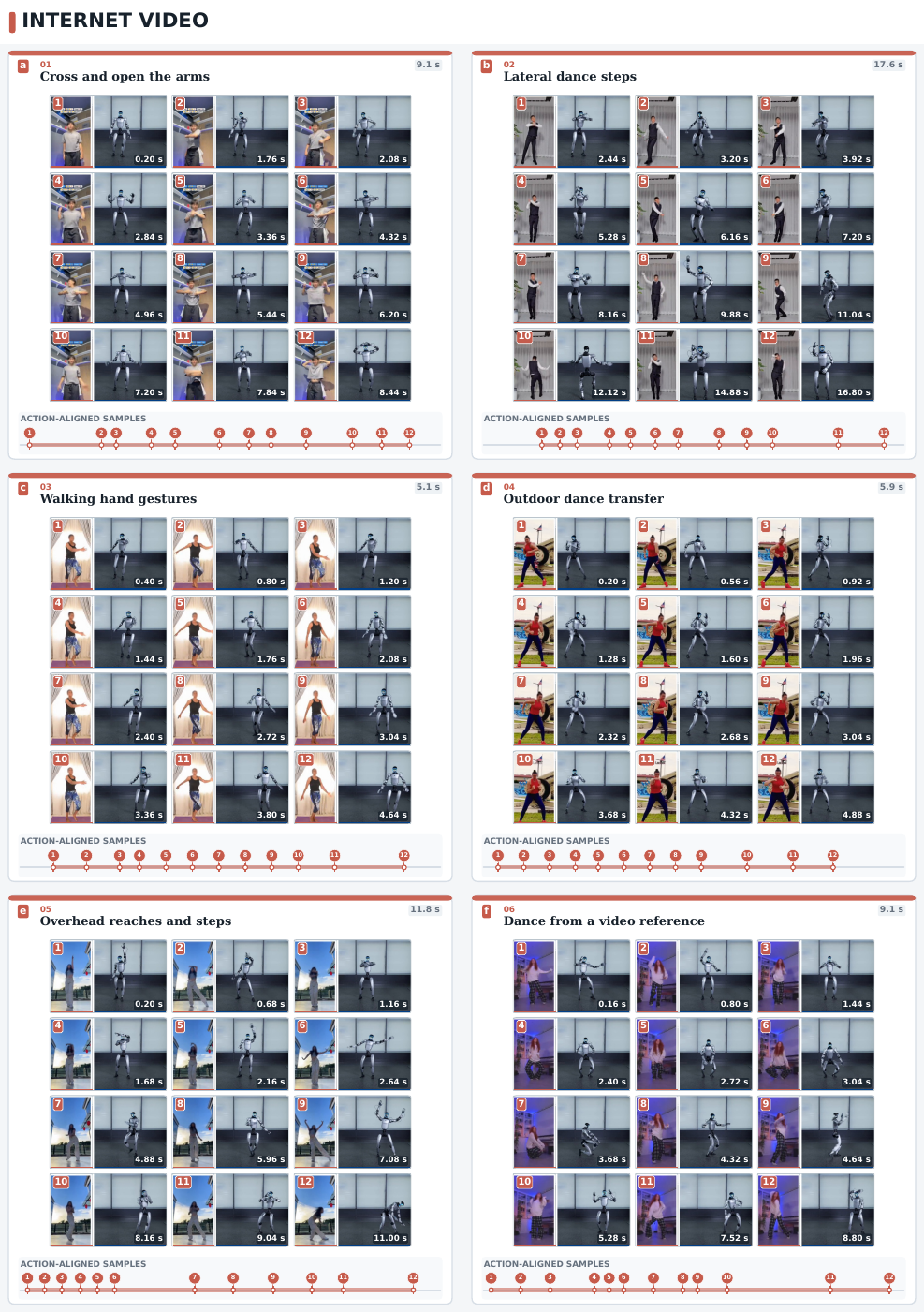}
\caption{Six Internet-video-conditioned motions executed on Unitree G1. Each capability card shows twelve numbered reference--execution pairs selected at clear corresponding action phases. Both crops in a pair come from the same composite-video timestamp; remaining pose differences reflect humanoid retargeting and dynamics rather than a timing shift introduced by the figure.}
\label{fig:appendix_real_robot_internet_video}
\end{figure*}

\begin{figure*}[p]
\centering
\includegraphics[width=\linewidth,height=0.92\textheight,keepaspectratio]{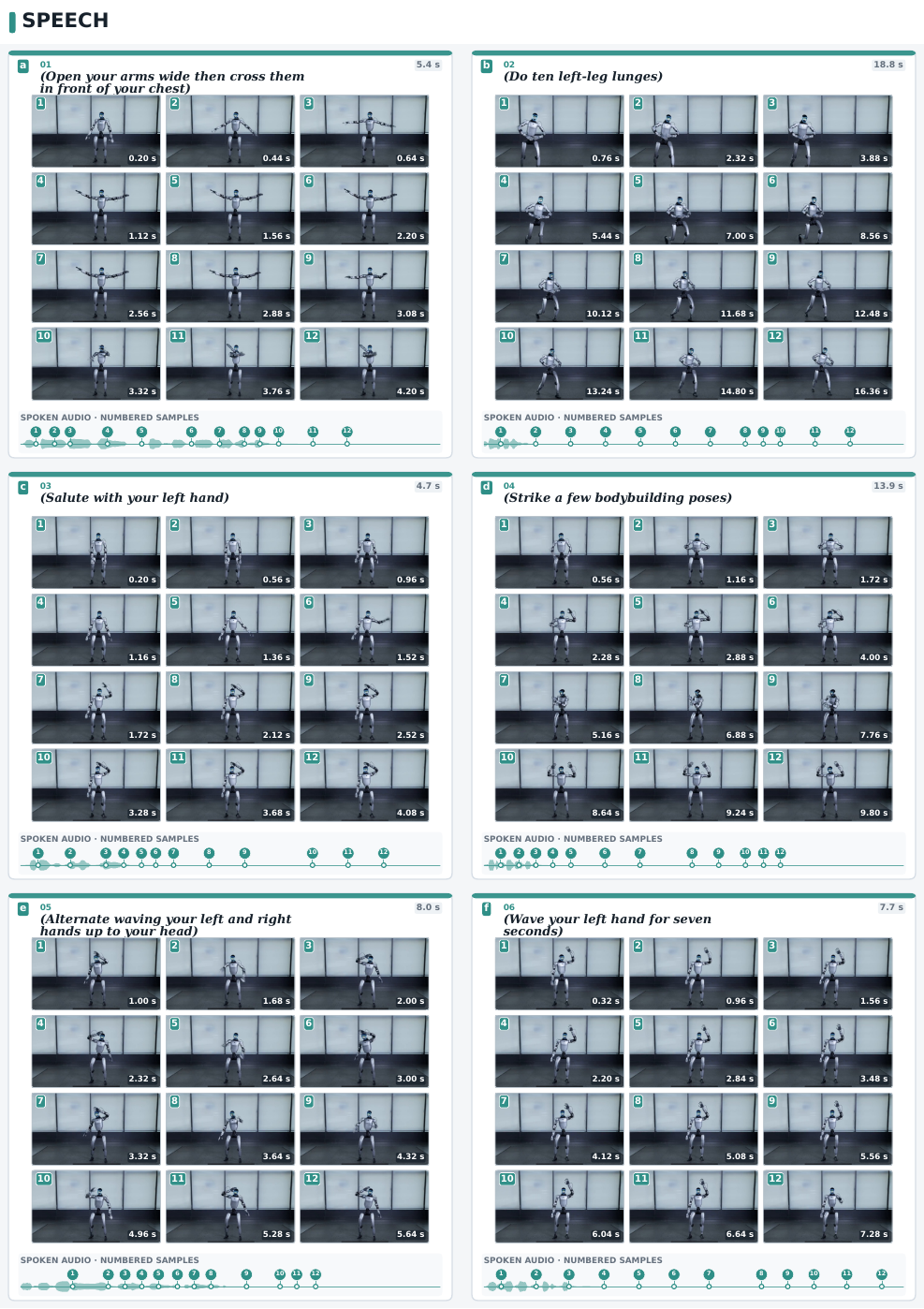}
\caption{Six speech-conditioned motions executed on Unitree G1. Each capability card sets the full spoken instruction in parentheses and italics, reports the clip duration, shows twelve numbered robot snapshots with timestamps, and maps those numbers to the complete input-audio waveform.}
\label{fig:appendix_real_robot_speech}
\end{figure*}

\begin{figure*}[p]
\centering
\includegraphics[width=\linewidth,height=0.92\textheight,keepaspectratio]{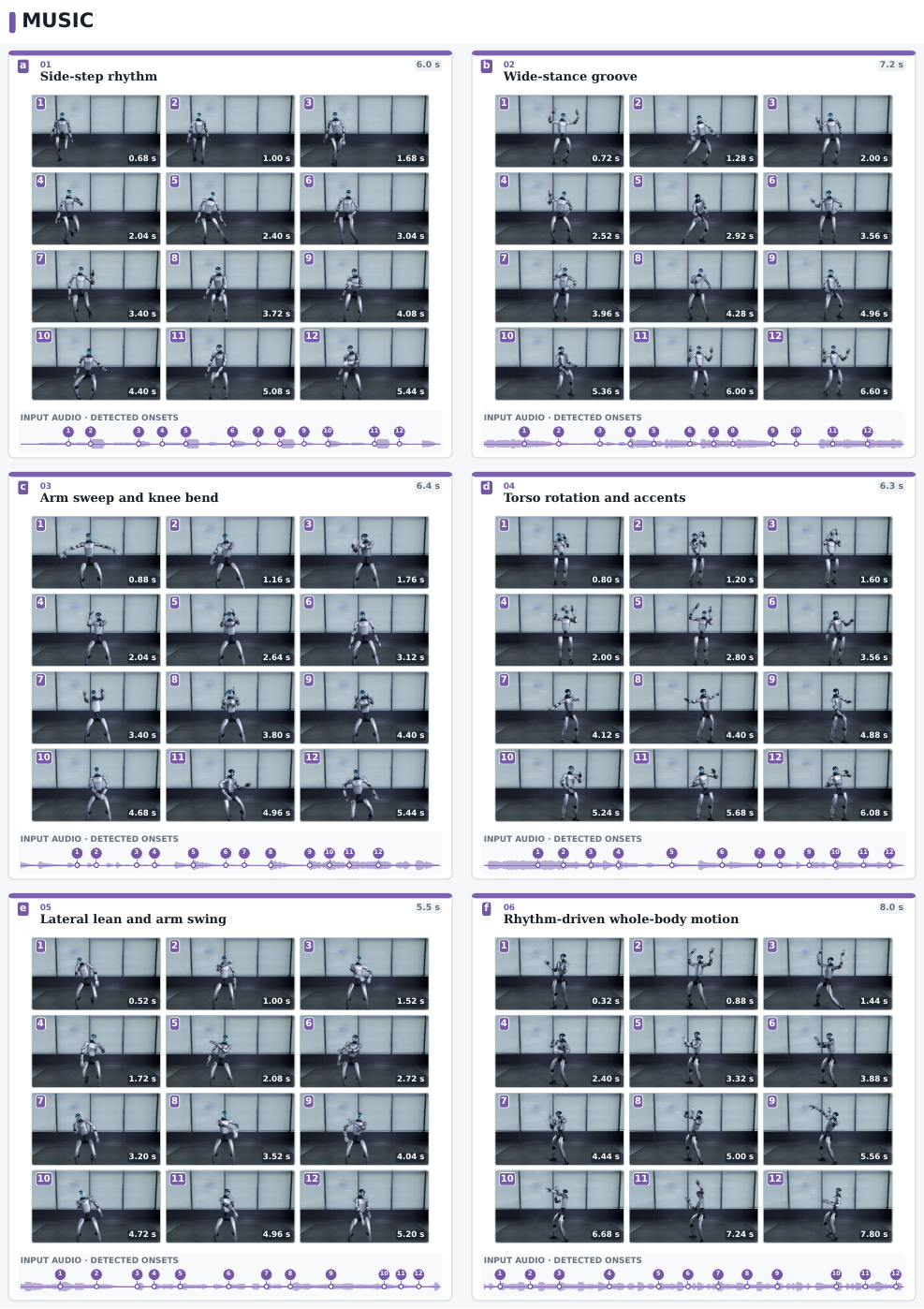}
\caption{Six music-conditioned motions executed on Unitree G1. Each capability card names the action, reports the clip duration, and shows twelve numbered robot snapshots above the complete input-audio waveform; matching numbers associate the snapshots with detected audio onsets.}
\label{fig:appendix_real_robot_music}
\end{figure*}

\begin{figure*}[p]
\centering
\includegraphics[width=\linewidth,height=0.92\textheight,keepaspectratio]{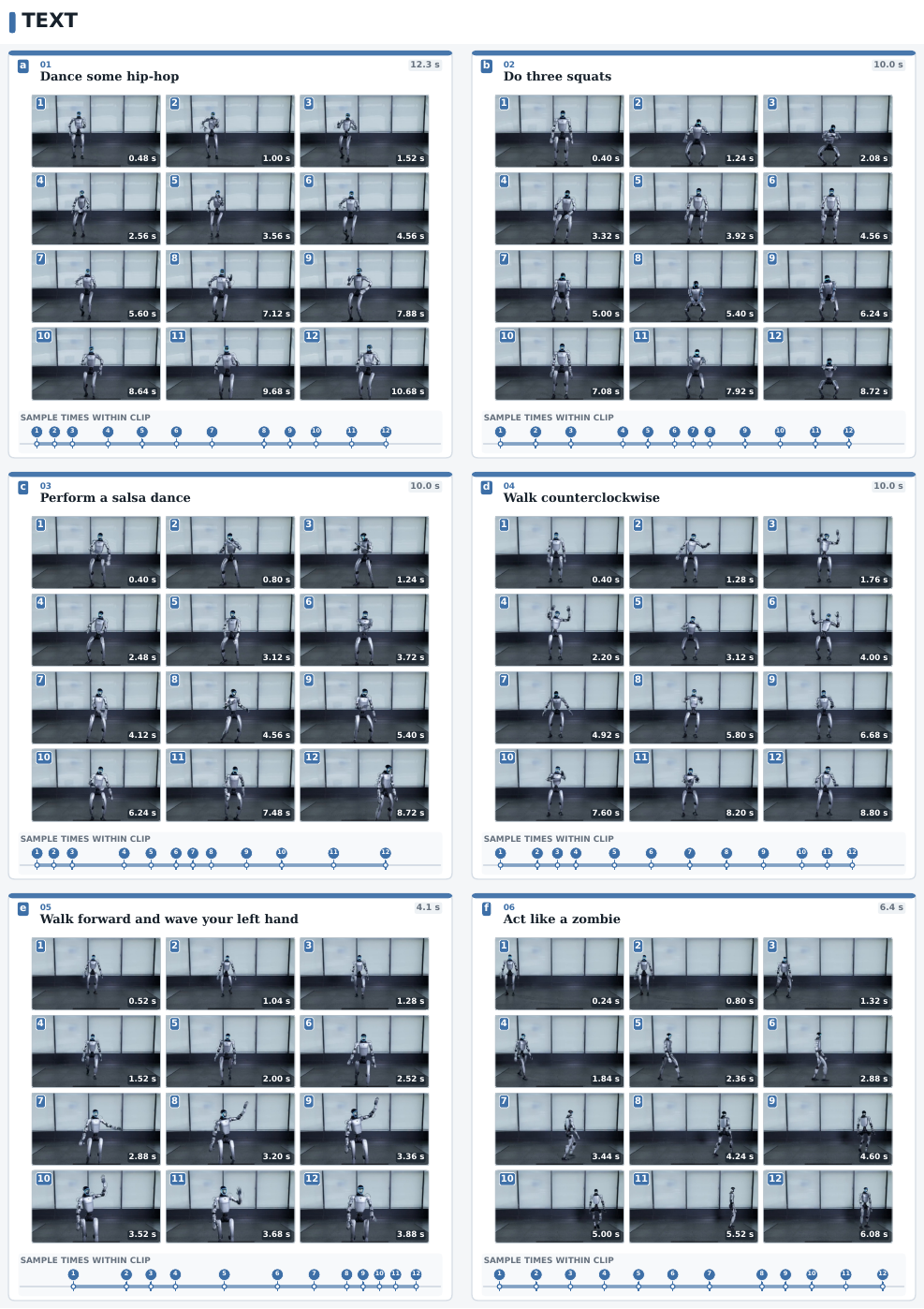}
\caption{Six text-conditioned motions executed on Unitree G1. Each capability card shows the complete input prompt, clip duration, twelve numbered robot snapshots with their timestamps, and the corresponding numbered locations on the execution timeline.}
\label{fig:appendix_real_robot_text}
\end{figure*}

\clearpage
\endgroup

\subsection{Failure Cases}
\label{app:failure_cases}

\paragraph{Highly dynamic actions.}
For an extreme action shown in the input video, \movenszero follows the overall motion but exhibits local discrepancies in the generated trajectory itself.
Because these discrepancies arise before humanoid tracking, they should not be attributed solely to low-level execution.
High-dynamic and contact-rich behaviors are relatively scarce in ordinary web videos, while physical verification retains only candidates that can be tracked reliably.
The targeted extreme-motion augmentation in Section~\ref{sec:data_augmentation} broadens this coverage but does not eliminate the remaining generation challenge.

\paragraph{Large camera motion.}
Video inputs with pronounced camera movement can yield generated trajectories that correspond less faithfully to the person's action.
Separating camera-induced apparent motion from articulated human motion is difficult in such footage.
Our data pipeline rejects temporal discontinuities and recovered sequences dominated by camera or root motion to prioritize reliable motion supervision (Section~\ref{sec:omnihm_pipeline}); it does not categorically remove every moving-camera video.
This curation may reduce coverage of the most challenging camera-motion conditions, but the qualitative case alone does not establish filtering as the cause of the observed error.

\clearpage

\section{Prompting Details}
\label{app:prompting_details}

\begingroup
\lstset{
    basicstyle=\rmfamily\fontsize{9.2}{11.2}\selectfont\color{black},
    columns=fullflexible,
    keepspaces=true,
    breaklines=true,
    breakatwhitespace=true,
    breakindent=1em,
    showstringspaces=false,
    frame=none,
    xleftmargin=0pt,
    xrightmargin=0pt,
    aboveskip=0pt,
    belowskip=0pt,
    morecomment=[l][\sffamily\bfseries\color{ZJUBlue}]{\#\#},
    morestring=[b]",
    stringstyle=\rmfamily,
    moredelim=[s][\ttfamily]{<}{>}
}
\lstdefinestyle{promptcode}{
    basicstyle=\ttfamily\fontsize{8.6}{10.8}\selectfont\color{black},
    stringstyle=\ttfamily
}
\lstdefinestyle{promptrepair}{stringstyle=\ttfamily}
\tcbset{
    promptpanel/.style={
        enhanced, breakable,
        colback=ZJUBlue!2!white, coltext=black,
        colframe=ZJUBlue!25!white,
        boxrule=0.4pt, arc=0.8mm,
        boxsep=0pt, left=3mm, right=3mm, top=2.5mm, bottom=2.5mm,
        before skip=8pt, after skip=10pt,
        colbacktitle=ZJUBlue!7!white, coltitle=ZJUBlue,
        fonttitle=\sffamily\bfseries\fontsize{9.5}{11.4}\selectfont,
        toptitle=1.8mm, bottomtitle=1.8mm,
        lines before break=5, pad at break*=2mm
    },
    promptnote/.style={
        enhanced, unbreakable,
        colback=black!2, colframe=black!2,
        boxrule=0pt, arc=0pt,
        borderline west={1.2pt}{0pt}{ZJUBlue!55!white},
        boxsep=0pt, left=3mm, right=3mm, top=2mm, bottom=2mm,
        before skip=6pt, after skip=8pt
    }
}

\subsection{Human Motion Video Filtering Prompts}
\label{app:data_filtering_prompts}
\label{app:human_motion_filtering_prompt}

The model answers yes or no, and the pipeline derives its filtering score from the corresponding token probabilities.

\begin{tcolorbox}[promptpanel, title={Prompt 1: Human motion video filtering}, title after break={Prompt 1: Human motion video filtering (continued)}, unbreakable]
\begin{lstlisting}
You are filtering videos for a human motion dataset.

Decide whether this video should be kept.

Answer yes ONLY if all of the following are true:
1. The main subject is a real live-action human, not a cartoon, anime character, 3D avatar, CGI person, game character, VTuber, or illustration.
2. The video is primarily about that person's body motion or action.
3. The human subject is clearly visible and large enough in the frame to observe their movement, rather than being tiny, far away, or hidden.
4. The video is not mainly a gameplay screen, livestream interface, animation clip, slideshow, static scene, or a crowd shot without a clear main human subject.

Real live-action cosplay, anime-style costumes, wigs, makeup, idol dance, stage dance, or dance-cover videos should still be kept if the visible subject is a real human and body motion is clear.

If there is any uncertainty, answer no.

Output exactly one word in lowercase:
yes
or
no
\end{lstlisting}
\end{tcolorbox}

\subsection{Audio-Visual Semantic Consistency Prompts}
\label{app:audio_visual_consistency_prompts}
\label{app:av_semantic_consistency_prompt}

This prompt assesses whether the sound is semantically compatible with the visible scene.
The complete system and user instructions are shown below.
The resulting score is used as a diagnostic rather than a hard acceptance gate.

\begin{tcolorbox}[promptpanel, title={Prompt 2: Audio-visual semantic consistency}, title after break={Prompt 2: Audio-visual semantic consistency (continued)}]
\begin{lstlisting}
System message:
You are an expert audio-video semantic analyst.

User message:
Analyze whether the audio semantically matches the video content. Focus on 'Semantic Consistency' (Does the sound object match the visual object?). Strict time synchronization is less important.

Here are some examples:
1. Video: A person playing piano. Audio: Piano music. -> Match (Yes)
2. Video: People playing basketball. Audio: Sound of bouncing ball and sneakers. -> Match (Yes)
3. Video: People playing basketball. Audio: Piano music. -> Mismatch (No)

Does the sound belong to the visual scene context? Answer with a single word: 'Yes' or 'No'.
\end{lstlisting}
\end{tcolorbox}

\clearpage
\subsection{Motion Classification Prompts}
\label{app:motion_classification_prompts}
\label{app:motion_type_classification_prompt}

Motion classification and scene classification (Section~\ref{app:scene_classification_prompts}) share one request in the implementation.
The two panels reproduce all task instructions, fixed label inventories, and decision rules, with shared guidelines repeated where relevant.
This panel contains all 306 motion labels and the complete motion rules.
The joint response format for both panels is shown at the end of Section~\ref{app:scene_classification_prompts}.

\begin{tcolorbox}[promptpanel, title={Prompt 3: Motion classification}, title after break={Prompt 3: Motion classification (continued)}]
\begin{lstlisting}
You are given a video clip of a person performing an action or motion.

Classify the clip directly from the visible video content. Do not rely on any external caption.

## Task: Motion Type
Choose ONE motion_type category from the following list:
[
"dance_hip_hop", "dance_street", "dance_breaking", "dance_popping", "dance_locking", "dance_house", "dance_waacking", "dance_voguing", "dance_kpop", "dance_ballet", "dance_contemporary", "dance_modern", "dance_jazz", "dance_tap", "dance_latin", "dance_salsa", "dance_bachata", "dance_cha_cha", "dance_ballroom", "dance_waltz", "dance_tango", "dance_foxtrot", "dance_traditional", "dance_folk", "dance_ethnic", "dance_social", "dance_party", "dance_freestyle", "dance_chinese_classical", "dance_chinese_folk",
"dance_trap", "dance_shuffle", "dance_square", "dance_line", "dance_belly", "dance_pole", "dance_reggaeton", "dance_afro", "dance_jazz_funk", "sports_basketball", "sports_american_football", "sports_volleyball", "sports_tennis", "sports_badminton", "sports_table_tennis", "sports_golf", "sports_baseball", "sports_softball", "sports_bowling", "sports_billiards", "sports_skateboarding", "sports_longboarding", "sports_roller_skating", "sports_inline_skating", "sports_bmx", "sports_scooter", "sports_surfing", "sports_windsurfing", "sports_kitesurfing", "sports_snowboarding",
"sports_skiing", "sports_ice_skating", "sports_speed_skating", "sports_swimming", "sports_diving", "sports_rowing", "sports_kayaking", "sports_canoeing", "sports_paddleboarding", "sports_cycling", "sports_mountain_biking", "sports_horse_riding", "sports_archery", "sports_climbing", "sports_bouldering", "sports_jumping", "sports_throwing", "sports_shooting", "sports_fishing", "sports_soccer", "sports_rugby", "sports_handball", "sports_cricket", "sports_hockey", "sports_lacrosse", "sports_frisbee", "sports_dodgeball", "sports_pickleball", "sports_track_and_field", "fitness_general_workout",
"fitness_aerobics", "fitness_hiit", "fitness_crossfit", "fitness_pilates", "fitness_bodyweight", "fitness_weightlifting", "fitness_powerlifting", "fitness_jump_training", "fitness_jumping_rope", "fitness_resistance_band", "fitness_rehabilitation", "fitness_push_up", "fitness_sit_up", "fitness_squat", "fitness_lunge", "fitness_plank", "fitness_burpee", "fitness_dumbbell", "fitness_kettlebell", "fitness_battle_rope", "fitness_medicine_ball", "fitness_box_jump", "fitness_balance", "fitness_stretching", "fitness_mobility", "fitness_warmup", "fitness_cooldown", "fitness_dance_workout", "fitness_hatha_yoga", "fitness_vinyasa_yoga",
"fitness_power_yoga", "fitness_ashtanga_yoga", "fitness_yin_yoga", "fitness_meditation", "fitness_tai_chi", "fitness_qigong", "sports_martial_arts", "sports_taekwondo", "sports_karate", "sports_judo", "sports_boxing", "sports_kickboxing", "sports_mma", "sports_muay_thai", "sports_wrestling", "sports_fencing", "sports_weapon", "sports_self_defense", "locomotion_walking", "locomotion_brisk_walking", "locomotion_jogging", "locomotion_running", "locomotion_sprinting", "locomotion_uphill", "locomotion_downhill", "locomotion_stairs", "locomotion_treadmill", "locomotion_hiking", "locomotion_crawling", "locomotion_shuffling",
"locomotion_limping", "locomotion_jumping", "locomotion_hopping", "locomotion_skipping", "locomotion_sidestepping", "locomotion_turning", "locomotion_spinning", "locomotion_backstepping", "locomotion_balancing", "sports_floor_exercise", "sports_apparatus_exercise", "sports_rhythmic_gymnastics", "sports_trampoline", "sports_bridge_pose", "sports_balance_beam", "extreme_tricking_combo", "extreme_spinning_kick", "extreme_jump_kick", "extreme_butterfly_kick", "extreme_twisting_kick", "extreme_frontflip", "extreme_backflip", "extreme_sideflip", "extreme_wall_flip", "extreme_aerial", "extreme_handspring", "extreme_cartwheel", "extreme_flip_combo", "extreme_parkour", "extreme_freerunning",
"extreme_vault", "extreme_wall_run", "extreme_wall_climb", "extreme_precision_jump", "extreme_gap_jump", "extreme_parkour_roll", "extreme_obstacle_traverse", "performance_speech", "performance_singing", "performance_conducting", "performance_acting", "performance_magic", "performance_mime", "performance_modeling", "performance_acrobatics", "performance_cheerleading", "performance_playing_guitar", "performance_playing_piano", "performance_playing_violin", "performance_playing_drums", "performance_dj", "performance_speaking_with_microphone", "performance_playing_instrument", "performance_posing", "performance_live_streaming", "gesture_sign_language", "gesture_communication", "gesture_greeting", "gesture_pointing", "gesture_waving",
"gesture_clapping", "gesture_thumbs_up", "gesture_salute", "gesture_bowing", "gesture_praying", "gesture_heart", "gesture_finger_heart", "gesture_victory_sign", "gesture_touching_face", "gesture_covering_face", "gesture_hair_adjusting", "gesture_head_nod", "gesture_head_shake", "gesture_shrugging", "daily_sitting", "daily_standing", "daily_sleeping", "daily_lying_down", "daily_cooking", "daily_cleaning", "daily_washing", "daily_dressing", "daily_undressing", "daily_eating", "daily_drinking", "daily_phone_use", "daily_typing", "daily_reading", "daily_writing", "daily_opening_door",
"daily_carrying", "daily_lifting", "daily_pushing", "daily_pulling", "daily_stretching", "daily_sitting_down", "daily_standing_up", "daily_getting_up", "daily_falling", "daily_kneeling", "daily_squatting", "daily_bending", "daily_leaning", "daily_reaching", "daily_holding", "daily_grabbing", "daily_picking_up", "daily_putting_down", "daily_placing", "daily_catching", "daily_throwing", "daily_pouring", "daily_cutting", "daily_brushing_teeth", "daily_brushing_hair", "daily_combing_hair", "daily_washing_hands", "daily_sweeping", "daily_mopping", "daily_vacuuming",
"daily_cleaning_window", "daily_shopping", "daily_driving", "daily_entering_vehicle", "daily_exiting_vehicle", "daily_using_tools", "daily_repairing", "daily_gardening", "daily_folding_clothes", "daily_ironing", "daily_making_bed", "daily_pet_interaction", "daily_childcare", "dance_general", "daily_organizing", "daily_sorting", "daily_assembling", "daily_installing", "daily_unboxing", "daily_packing", "daily_hanging", "daily_feeding", "daily_operating_machine", "daily_crafting", "daily_decorating", "daily_loading_unloading", "daily_material_handling", "dance_other", "sports_other", "fitness_other",
"locomotion_other", "extreme_other", "performance_other", "gesture_other", "daily_other", "other_other"
]

## Guidelines
- Focus on the main visible person and the dominant motion across the whole clip.
- Use the actual video motion as primary evidence for motion_type.
- If multiple actions appear, choose the category that best summarizes the whole clip.
- Be conservative with fine-grained labels. If the broad group is clear but the exact fine-grained style is not visually certain, choose that group's general/other label instead of guessing.
- Do not classify generic short-video dancing as hip_hop, kpop, popping, locking, jazz_funk, or other specific dance styles unless clear visual evidence supports it.
- Do not invent a new category for any field.

## Motion Type Rules
- If the broad motion group is clear but no specific child category matches, choose that group's other label, such as daily_other, sports_other, dance_other, fitness_other, performance_other, or gesture_other.
- Use motion_type="other_other" only when the clip has no clear human motion category, the visible subject is not a person-like actor, or the motion is not covered by any broad group.
- Use daily_* for everyday object interaction, household tasks, self-care, organizing, packing, installing, unboxing, feeding, and general manual work outside an explicit sport/fitness/performance context.
- Use fitness_* for exercise or workout actions, including yoga-style poses and flows such as fitness_hatha_yoga, fitness_vinyasa_yoga, fitness_power_yoga, fitness_tai_chi, fitness_qigong, fitness_stretching, fitness_balance, or fitness_meditation.
- Use sports_soccer for association football/soccer, and sports_american_football for American football.
- Use sports_* for named sports, formal gymnastics events, and combat or martial-arts sports. Use sports_weapon for sword/staff/weapon routines, and sports_martial_arts for broad martial-arts styles that are too fine-grained to label separately, such as capoeira.
- Use extreme_* only for no-equipment, high-difficulty body tricks such as tricking kicks, flips, parkour, freerunning, vaults, wall runs, wall climbs, precision jumps, gap jumps, and obstacle traversal.
- Use extreme_frontflip, extreme_backflip, extreme_sideflip, extreme_wall_flip, extreme_aerial, extreme_handspring, extreme_cartwheel, or extreme_flip_combo for free-form flips and tumbling that are not clearly a formal gymnastics event.
- Use sports_floor_exercise, sports_apparatus_exercise, sports_rhythmic_gymnastics, sports_trampoline, or sports_balance_beam only for formal gymnastics apparatus/event contexts.
- Use locomotion_* for movement through space such as walking, running, stairs, hiking, crawling, turning, or balancing when that is the dominant action.
- Use performance_* for staged speaking, singing, acting, posing, livestreaming, magic, conducting, or playing instruments.
- Use gesture_* for communicative hand, head, or face gestures when the main action is gesturing rather than full-body activity.
- Do not use a social interaction label. For clips with multiple people, classify the main tracked person's body motion itself: conversation as gesture_communication or performance_speech, fighting as sports_boxing, sports_martial_arts, sports_self_defense, or performance_acting, following/chasing as locomotion_*, and object handoff as daily_*.
\end{lstlisting}
\end{tcolorbox}

\clearpage
\subsection{Scene Classification Prompts}
\label{app:scene_classification_prompts}
\label{app:scene_classification_prompt}

This panel contains the scene components of the same classification request.
The complete taxonomy contains 17 parent groups and 208 scene labels, followed by five spatial settings, all scene decision rules, and the four-field JSON response format shared with Section~\ref{app:motion_classification_prompts}.
Only JSON whitespace is condensed for display.

\begin{tcolorbox}[promptpanel, title={Prompt 4: Scene classification}, title after break={Prompt 4: Scene classification (continued)}]
\begin{lstlisting}
You are given a video clip of a person performing an action or motion.

Classify the clip directly from the visible video content. Do not rely on any external caption.

## Task: Motion Scene Taxonomy
Choose ONE motion_scene category that best describes the environment/place where the motion happens.
Then choose motion_scene_group as the exact parent key that contains that motion_scene.
The parent-child relation is fixed by this taxonomy:
{
  "home": ["home_living_room", "home_bedroom", "home_kitchen", "home_dining_room", "home_bathroom", "home_hallway", "home_balcony", "home_yard", "home_backyard", "home_front_yard", "home_garage", "home_driveway", "home_porch", "home_patio", "home_terrace", "home_office", "home_laundry_room", "home_basement", "home_attic", "home_storage_room", "home_entryway", "home_staircase", "home_playroom", "home_utility_room", "home_plain_room", "home_other"],
  "training": ["training_dance_studio", "training_fitness_gym", "training_fitness_studio", "training_yoga_studio", "training_pilates_studio", "training_martial_arts_dojo", "training_boxing_gym", "training_climbing_gym", "training_other"],
  "sports": ["sports_indoor_sports_hall", "sports_school_gym", "sports_indoor_basketball_court", "sports_outdoor_basketball_court", "sports_indoor_tennis_court", "sports_outdoor_tennis_court", "sports_indoor_badminton_court", "sports_soccer_field", "sports_baseball_field", "sports_football_field", "sports_rugby_field", "sports_golf_course", "sports_running_track", "sports_stadium", "sports_skatepark", "sports_skating_rink", "sports_bowling_alley", "sports_billiards_room", "sports_swimming_pool_indoor", "sports_swimming_pool_outdoor", "sports_ski_slope", "sports_gymnastics_gym", "sports_other"],
  "studio_stage": ["studio_stage_stage", "studio_stage_concert_stage", "studio_stage_theater", "studio_stage_auditorium", "studio_stage_lecture_stage", "studio_stage_conference_stage", "studio_stage_presentation_stage", "studio_stage_tv_studio", "studio_stage_news_studio", "studio_stage_weather_studio", "studio_stage_radio_studio", "studio_stage_podcast_studio", "studio_stage_livestream_room", "studio_stage_broadcast_studio", "studio_stage_photo_studio", "studio_stage_rehearsal_room", "studio_stage_green_screen_studio", "studio_stage_plain_backdrop", "studio_stage_black_background", "studio_stage_other"],
  "education_office": ["education_office_classroom", "education_office_school_campus", "education_office_lecture_hall", "education_office_office", "education_office_conference_room", "education_office_corridor", "education_office_lobby", "education_office_library", "education_office_laboratory_classroom", "education_office_other"],
  "commercial_service": ["commercial_service_shopping_mall", "commercial_service_retail_store", "commercial_service_supermarket", "commercial_service_salon", "commercial_service_hotel_room", "commercial_service_hotel_lobby", "commercial_service_bank", "commercial_service_gas_station", "commercial_service_reception_area", "commercial_service_showroom", "commercial_service_market", "commercial_service_other"],
  "food_hospitality": ["food_hospitality_restaurant", "food_hospitality_cafe", "food_hospitality_bar", "food_hospitality_commercial_kitchen", "food_hospitality_restaurant_kitchen", "food_hospitality_food_stall", "food_hospitality_cafeteria", "food_hospitality_bakery", "food_hospitality_other"],
  "industrial_worksite": ["industrial_worksite_workshop", "industrial_worksite_woodworking_shop", "industrial_worksite_auto_repair_shop", "industrial_worksite_maker_space", "industrial_worksite_factory_floor", "industrial_worksite_warehouse", "industrial_worksite_laboratory", "industrial_worksite_construction_site", "industrial_worksite_garage_workshop", "industrial_worksite_tool_room", "industrial_worksite_farm_work_area", "industrial_worksite_other"],
  "transportation": ["transportation_parking_garage", "transportation_subway_station", "transportation_train_station", "transportation_airport_terminal", "transportation_bus_stop", "transportation_station_platform", "transportation_subway_platform", "transportation_train_platform", "transportation_other"],
  "vehicle_interior": ["vehicle_interior_car", "vehicle_interior_taxi", "vehicle_interior_bus", "vehicle_interior_truck_cabin", "vehicle_interior_van", "vehicle_interior_train_car", "vehicle_interior_subway_car", "vehicle_interior_airplane_cabin", "vehicle_interior_boat_cabin", "vehicle_interior_generic", "vehicle_interior_other"],
  "urban_public": ["urban_public_street", "urban_public_road", "urban_public_sidewalk", "urban_public_crosswalk", "urban_public_alley", "urban_public_plaza", "urban_public_parking_lot", "urban_public_rooftop", "urban_public_bridge", "urban_public_building_entrance", "urban_public_hotel_entrance", "urban_public_driveway", "urban_public_gas_station", "urban_public_outdoor_market", "urban_public_other"],
  "outdoor_recreation": ["outdoor_recreation_park", "outdoor_recreation_grass_lawn", "outdoor_recreation_garden", "outdoor_recreation_courtyard", "outdoor_recreation_playground", "outdoor_recreation_picnic_area", "outdoor_recreation_pavilion", "outdoor_recreation_covered_walkway", "outdoor_recreation_amusement_park", "outdoor_recreation_carnival", "outdoor_recreation_beach", "outdoor_recreation_waterfront", "outdoor_recreation_poolside", "outdoor_recreation_trail", "outdoor_recreation_other"],
  "nature": ["nature_forest_trail", "nature_mountain_trail", "nature_field", "nature_agricultural_field", "nature_grass_field", "nature_pasture", "nature_farm", "nature_tea_field", "nature_orchard", "nature_vineyard", "nature_snow_field", "nature_desert", "nature_lake", "nature_riverbank", "nature_other"],
  "medical_care": ["medical_care_hospital_room", "medical_care_clinic_room", "medical_care_rehabilitation_room", "medical_care_therapy_room", "medical_care_operating_room", "medical_care_other"],
  "digital_virtual": ["digital_virtual_game_interface", "digital_virtual_screen_recording", "digital_virtual_virtual_environment", "digital_virtual_fantasy_environment", "digital_virtual_space_environment", "digital_virtual_animated_scene", "digital_virtual_menu_screen", "digital_virtual_video_call", "digital_virtual_other"],
  "special_environment": ["special_environment_underwater", "special_environment_water_surface", "special_environment_snow_ice", "special_environment_cave", "special_environment_plain_background", "special_environment_other"],
  "other": ["other_other"]
}

Valid motion_scene_group values:
["home", "training", "sports", "studio_stage", "education_office", "commercial_service", "food_hospitality", "industrial_worksite", "transportation", "vehicle_interior", "urban_public", "outdoor_recreation", "nature", "medical_care", "digital_virtual", "special_environment", "other"]

## Task: Motion Scene Setting
Choose ONE motion_scene_setting category that best describes the spatial setting:
["indoor", "outdoor", "semi_outdoor", "vehicle", "other"]

## Guidelines
- Focus on the main visible person and the dominant motion across the whole clip.
- Use visible background/place as primary evidence for motion_scene, motion_scene_group, and motion_scene_setting.
- Be conservative with fine-grained labels. If the broad group is clear but the exact fine-grained style is not visually certain, choose that group's general/other label instead of guessing.
- motion_scene uses the pattern motion_scene_group + "_" + specific_place, for example home_kitchen, sports_soccer_field, vehicle_interior_bus, or digital_virtual_game_interface.
- motion_scene_group MUST be the parent key of motion_scene in the taxonomy above. Do not mix a child label with a different parent group.
- For vehicle interiors, prefer a specific motion_scene such as vehicle_interior_car, vehicle_interior_bus, vehicle_interior_train_car, vehicle_interior_subway_car, vehicle_interior_airplane_cabin, vehicle_interior_truck_cabin, vehicle_interior_van, vehicle_interior_taxi, or vehicle_interior_boat_cabin when clearly supported; use vehicle_interior_generic only as a fallback.
- For car/bus/train/subway/airplane/truck/van/taxi/boat interior scenes, choose motion_scene_group="vehicle_interior" and motion_scene_setting="vehicle".
- If the broad scene group is clear but no specific child category matches, choose that group's other label, such as home_other, sports_other, studio_stage_other, or urban_public_other.
- Use motion_scene="other_other" and motion_scene_group="other" only when even the broad scene group is unclear or not covered by the taxonomy.
- Do not invent a new category for any field.

## Scene Setting Rules
- indoor: enclosed building interiors such as homes, gyms, offices, classrooms, restaurants, shops, warehouses, hospitals, and studios.
- outdoor: open-air places such as streets, sidewalks, parks, fields, beaches, outdoor courts, mountains, forests, parking lots, and rooftops.
- semi_outdoor: partly covered or partly open spaces such as balconies, courtyards, covered walkways, open-air stations, patios, terraces, and building entrances.
- vehicle: inside a transportation vehicle or cabin such as a car, taxi, bus, truck, van, train car, subway car, airplane cabin, or boat cabin.
- other: use only when the spatial setting is unclear or does not fit the categories above.
\end{lstlisting}
\par\smallskip
\begin{lstlisting}[style=promptcode]
## OUTPUT FORMAT
You MUST respond with a valid JSON object in the following format (no other text before or after):
{
    "motion_type": "<one category from the motion_type list above>",
    "motion_scene": "<one child category from the scene taxonomy above>",
    "motion_scene_group": "<the exact parent key of motion_scene>",
    "motion_scene_setting": "<one category from the motion_scene_setting list above>"
}
\end{lstlisting}
\end{tcolorbox}

\clearpage
\subsection{Human Attribute Annotation Prompts}
\label{app:human_attribute_prompts}
\label{app:human_attribute_annotation_prompt}

The human-attribute, video-caption (Section~\ref{app:annotation_prompts}), and motion-timeline (Section~\ref{app:motion_caption_prompts}) panels contain the complete task components of one joint request to Qwen3-VL-32B-Instruct.
Shared viewing instructions are repeated for clarity.
Attributes are visual estimates.
The complete joint JSON format appears in Section~\ref{app:motion_caption_prompts}.

\begin{tcolorbox}[promptpanel, title={Prompt 5: Human attribute annotation}, title after break={Prompt 5: Human attribute annotation (continued)}]
\begin{lstlisting}
Analyze the supplied video and return one valid JSON object. The clip duration is <DURATION_SECONDS> seconds.

Before writing JSON, silently scan the clip from beginning to end, build an event timeline, and verify every claimed object and action against visible frames. Do not output this internal reasoning.

## Task: Human info
Estimate the primary person's attributes from visual observation only:
- age: integer
- gender: one of ["male", "female"]
- height: integer centimeters
- weight: integer kilograms
- mood: one of ["happy", "excited", "calm", "focused", "confident", "playful", "energetic", "relaxed", "serious", "tired", "sad", "angry", "surprised", "nervous", "bored", "neutral"]
\end{lstlisting}
\end{tcolorbox}

\subsection{Object Interaction Annotation Prompts}
\label{app:object_interaction_prompts}
\label{app:interaction_object_annotation_prompt}

Object interaction labels are obtained through a separate video-analysis request.
Both system and user instructions are shown below.
The prompt identifies objects contacted or manipulated by the hands or feet of visible people and returns a JSON array of object names, with an empty array when no qualifying interaction is visible.

\begin{tcolorbox}[promptpanel, title={Prompt 6: Object interaction annotation}, title after break={Prompt 6: Object interaction annotation (continued)}, unbreakable]
\begin{lstlisting}
System message:
You are an expert video analysis assistant. Pay close attention to temporal details and physical interactions.

User message:
Task: Identify all primary objects that people in the scene are interacting with.

Step 1: Locate all visible people.
Step 2: Only consider physical contact made with **hands or feet**.
Step 3: Ignore any contact with parts above the neck (head, face, mouth, nose, etc.) -- these do NOT count as object interaction.
Step 4: Identify objects being touched, held, manipulated, pressed, or used by hands or feet.
Step 5: Ignore background objects (floor, walls) unless essential.
Step 6: If absolutely no objects are touched by hands or feet in the entire scene (e.g., everyone is just standing without touching anything), output [].

Output: ONLY a valid JSON array of strings representing the identified objects. Do not write sentences or add explanations. Order does not matter.
Use double quotes, even if an object name contains apostrophes or brand names.

Example Output 1 (multiple objects): ["cellphone", "basketball", "cup"]
Example Output 2 (single object): ["laptop"]
Example Output 3 (no objects): []
\end{lstlisting}
\end{tcolorbox}

\clearpage
\subsection{Video Caption Annotation Prompts}
\label{app:annotation_prompts}
\label{app:video_caption_annotation_prompt}

This panel reproduces the complete video-caption task and the common viewing instruction from the joint request described in Section~\ref{app:human_attribute_prompts}.
It describes the primary person, scene, visible objects, and chronological action sequence, and requires consistency with the motion timeline in Section~\ref{app:motion_caption_prompts}.

\begin{tcolorbox}[promptpanel, title={Prompt 7: Video caption annotation}, title after break={Prompt 7: Video caption annotation (continued)}]
\begin{lstlisting}
Analyze the supplied video and return one valid JSON object. The clip duration is <DURATION_SECONDS> seconds.

Before writing JSON, silently scan the clip from beginning to end, build an event timeline, and verify every claimed object and action against visible frames. Do not output this internal reasoning.

## Task: Factual video caption
Write 2-4 concise English sentences describing only clearly visible evidence:
- the primary person's appearance and clothing;
- the physical environment and clearly visible objects;
- the person's action sequence in chronological order.

Accuracy rules:
- Do not invent a story, intention, occupation, relationship, location, or off-screen event.
- Do not identify an ambiguous small object more specifically than the pixels support. Use phrases such as "a small handheld object" when uncertain.
- Do not substitute a familiar activity name for the motion actually shown. In a fitness or exercise clip, describe the literal posture and body trajectory instead of relying on a named exercise. For example: "sitting with both feet raised and rotating the torso side to side while holding a dumbbell," never "performing sit-ups" for that motion.
- Describe visible contact rather than inferred purpose: "presses both hands to the cheeks," not "applies product," unless the product and application are visibly unambiguous; "looks at a card or strip," not "checks a result"; "raises an object to the ear," not "makes a call."
- Do not write hedges such as "appears to," "seems to," "as if," or "likely." When evidence is uncertain, replace the interpretation with literal body motion, object shape, and scene geometry.
- Name a room type, event, or activity only when physical visual evidence makes it unambiguous. On-screen text or promotional graphics alone do not establish the real-world setting.
- Make the action sequence in `video_caption` consistent with the detailed motion timeline. Re-check the beginning, middle, and end before finalizing it.
- Describe visible facial expressions, but do not infer hidden emotions or motivations.
- Distinguish camera motion or zoom from human motion.
- Use singular wording for one main subject: "the person", "the man", or "the woman". Never use singular "they/them/their"; repeating "the person" is preferable.
\end{lstlisting}
\end{tcolorbox}

\clearpage
\subsection{Motion Caption Annotation Prompts}
\label{app:motion_caption_prompts}
\label{app:motion_caption_annotation_prompt}

This panel reproduces the complete motion-timeline task and the common viewing instruction from the same joint request (Section~\ref{app:human_attribute_prompts}).
The factuality rules it references are included in the video-caption panel in Section~\ref{app:annotation_prompts}.
\texttt{<DURATION\_SECONDS>} is the clip-specific duration, and the segment limit is 24.
The final schema is the complete combined response for the video-caption, human-attribute, and motion-timeline tasks.

\begin{tcolorbox}[promptpanel, title={Prompt 8: Motion caption annotation}, title after break={Prompt 8: Motion caption annotation (continued)}]
\begin{lstlisting}
Analyze the supplied video and return one valid JSON object. The clip duration is <DURATION_SECONDS> seconds.

Before writing JSON, silently scan the clip from beginning to end, build an event timeline, and verify every claimed object and action against visible frames. Do not output this internal reasoning.

## Task: Atomic motion timeline
A motion segment is one temporally continuous atomic action or stable pose. Split at the frame where the primary action state changes.

You MUST start a new segment when any of these visibly changes:
1. body support or posture, such as standing, sitting, kneeling, lying, or moving onto hands and knees;
2. locomotion state or direction, such as idle -> walking, walking -> stopping, or turning around;
3. object interaction, such as playing with a controller -> reaching for a cup -> drinking -> returning to the controller;
4. a standalone semantic gesture starts or ends, such as standing neutrally -> raising both fists to cheer -> returning to neutral;
5. the main visible person changes after a shot cut.
6. a salient head or face action changes, such as eyes closed -> opening the eyes and looking at the camera, or looking down -> lifting the head and smiling.

Do NOT merge different sequential actions into one segment. If a segment description would need words such as "then", "before", "later", "followed by", "transitions", or "resumes", split it into multiple segments instead.

Do NOT over-segment continuous motion:
- Repeated cycles of the same semantic action remain one segment unless the activity clearly stops or changes.
- Repeated left/right or up/down phases of one exercise remain one repeated-routine segment. Conversational emphasis gestures also remain inside the ongoing speaking segment.
- When a deliberate gesture demonstration or choreographed hand sequence is itself the primary action, split its recognizable actions, such as waving -> forming a heart -> clapping -> lowering the hands. A later recurrence of the same gesture after another gesture is a new time interval.
- Continuous shoe tying, shaping clay, painting, walking, adjusting clothing/hair, speaking with similar gestures, or holding one pose remains one segment.
- Keep one uninterrupted semantic action together even when it has natural phases. For example, reaching for a cup, sipping, and placing it back is one drinking action.
- Do not divide a short clip into equal time bins merely because hand positions vary.
- Camera zoom, reframing, or a background slide change alone is not a human-motion boundary.

Timeline rules:
- Return between 1 and 24 segments, using as many as the visible action changes require.
- The first segment must start at 0.0 and the final segment must end at <DURATION_SECONDS>.
- Segments must be chronological, contiguous, non-overlapping, and cover the full clip.
- Use seconds with approximately 0.1-second precision.
- Each body-part description must describe only that segment, not summarize the whole video.
- The factuality rules for ambiguous objects and inferred intent apply to every motion-description field too.
- Give every segment an `action` field: a short gerund phrase naming its high-level semantic action, such as "shaping clay" or "raising clenched fists".
- In fitness/exercise footage, the `action` phrase MUST state the literal body movement and MUST NOT be only a named exercise. Prefer a literal movement phrase over any uncertain dance, sport, occupation, or intention.
- Give every segment a non-negative integer `subject_id`. Use the same ID for the same physical person and a different ID after a shot cut to a different person, even if both people perform the same action.
- If adjacent intervals would have the same `action` phrase for the same person, merge them before returning JSON.
- Before finalizing, compare face, hair, clothing, and body appearance across the clip to detect a person change.
- Use the person's own left/right when direction is clear; otherwise avoid guessing.
- If a body part is cropped, occluded, blurry, or off-screen, state that explicitly.

Segmentation examples:
- playing a game -> reaching for and drinking from a cup -> resuming the game: 3 segments.
- kneeling -> moving onto hands and knees -> sitting back down: 3 segments.
- standing neutrally -> raising both fists to cheer -> lowering the arms: 3 segments.
- continuously wrapping and tying one ballet shoe: 1 segment.
- a girl speaking -> shot cut to a boy speaking: 2 segments with subject_id 0 and 1.
\end{lstlisting}
\par\smallskip
\begin{lstlisting}[style=promptcode]
## Output format
Return ONLY a valid JSON object in this exact structure:
{
    "video_caption": "<2-4 sentence factual caption>",
    "human_info": {
        "age": <integer>,
        "gender": "<male or female>",
        "height": <integer>,
        "weight": <integer>,
        "mood": "<one mood option>"
    },
    "motion_caption": {
        "segments": [
            {
                "start_time": <float>,
                "end_time": <float>,
                "subject_id": <non-negative integer>,
                "action": "<short high-level gerund phrase>",
                "whole_body": "<atomic description or visibility explanation>",
                "upper_limb": "<atomic description or visibility explanation>",
                "lower_limb": "<atomic description or visibility explanation>",
                "hands": "<atomic description or visibility explanation>",
                "face": "<atomic description or visibility explanation>"
            }
        ]
    }
}
\end{lstlisting}
\end{tcolorbox}

\clearpage
\subsection{Motion Generation Task Prompts}
\label{app:motion_generation_prompts}

The following prompts specify the generation task for each input modality.
Angle-bracketed placeholders denote the text description, music waveform, speech waveform, or reference video.

The text prompt requests a motion sequence matching the supplied description.

\begin{tcolorbox}[promptpanel, title={Prompt 9: Text-conditioned motion generation}, title after break={Prompt 9: Text-conditioned motion generation (continued)}, unbreakable]
\begin{lstlisting}
Generate a motion sequence that matches the following description.
<MOTION_DESCRIPTION>
\end{lstlisting}
\end{tcolorbox}

The music prompt requests dance aligned with the supplied waveform's rhythm and energy.

\begin{tcolorbox}[promptpanel, title={Prompt 10: Music-conditioned motion generation}, unbreakable]
\begin{lstlisting}
Generate a dance motion that follows the rhythm and energy of this music.
<MUSIC_AUDIO>
\end{lstlisting}
\end{tcolorbox}

The speech prompt asks the model to realize the task spoken in the supplied audio as body motion.

\begin{tcolorbox}[promptpanel, title={Prompt 11: Speech-conditioned motion generation}, unbreakable]
\begin{lstlisting}
Listen for the spoken task in the audio, then carry out that task as body motion.
<SPEECH_AUDIO>
\end{lstlisting}
\end{tcolorbox}

The video prompt asks the model to imitate the body movement in the reference video.

\begin{tcolorbox}[promptpanel, title={Prompt 12: Video-conditioned motion generation}, unbreakable]
\begin{lstlisting}
The clip is the reference. Imitate the person's body movement as closely as possible.
<REFERENCE_VIDEO>
\end{lstlisting}
\end{tcolorbox}

\endgroup

\end{document}